\documentclass{article}

\PassOptionsToPackage{numbers, compress}{natbib}
\usepackage[eandd,final]{neurips_2026}

\usepackage[utf8]{inputenc} 
\usepackage[T1]{fontenc}    

\usepackage{xcolor}         
\definecolor{cvprblue}{rgb}{0.21,0.49,0.74}

\usepackage{float}
\usepackage[pagebackref,breaklinks,colorlinks,citecolor=cvprblue,linkcolor=cvprblue,urlcolor=cvprblue]{hyperref}
\usepackage{url}            
\usepackage{booktabs}       
\usepackage{amsfonts}       
\usepackage{nicefrac}       
\usepackage{microtype}      

\usepackage{graphicx}
\usepackage{amsmath}
\usepackage{amssymb}
\usepackage{wrapfig}

\usepackage{amsmath}
\usepackage{booktabs}
\usepackage{graphicx}
\usepackage{pifont}
\usepackage{todonotes}
\usepackage{multirow}
\usepackage{multicol}
\usepackage{makecell}
\usepackage{adjustbox}
\usepackage{tabularx}
\usepackage{ragged2e}
\usepackage{array}

\newcommand{\cmark}{\ding{51}}
\newcommand{\xmark}{\ding{55}}
\newcommand{\modelname}{\emph{Rescale}}
\newcommand{\benchmarkname}{\emph{GenScale}}

\usepackage[capitalize]{cleveref}
\crefname{section}{Sec.}{Secs.}
\Crefname{section}{Section}{Sections}
\Crefname{table}{Table}{Tables}
\crefname{table}{Tab.}{Tabs.}

\definecolor{amber}{rgb}{1.0, 0.4, 0.0}

\title{\benchmarkname{}: A Benchmark for Relative Object Scale in Image Generation and Editing}
\hypersetup{
  pdftitle={GenScale: A Benchmark for Relative Object Scale in Image Generation and Editing},
  pdfauthor={Lingxiao Li, Max Whitton, Ledell Wu, Boqing Gong}
}

\author{
Lingxiao Li$^{1}$ \quad Max Whitton$^{1}$ \quad Ledell Wu$^{2}$ \quad Boqing Gong$^{1}$ \\
\textsuperscript{1}Boston University\;
\textsuperscript{2}Creatify AI \; 
\and
\url{https://lingxiao-li.github.io/genscale/}
}

\begin{document}

\maketitle

\vspace{-5ex}
\begin{figure}[H]
    \centering
    \setlength{\abovecaptionskip}{0.2cm}   
    \setlength{\belowcaptionskip}{-0.2cm}   
    \includegraphics[width=0.96\textwidth]{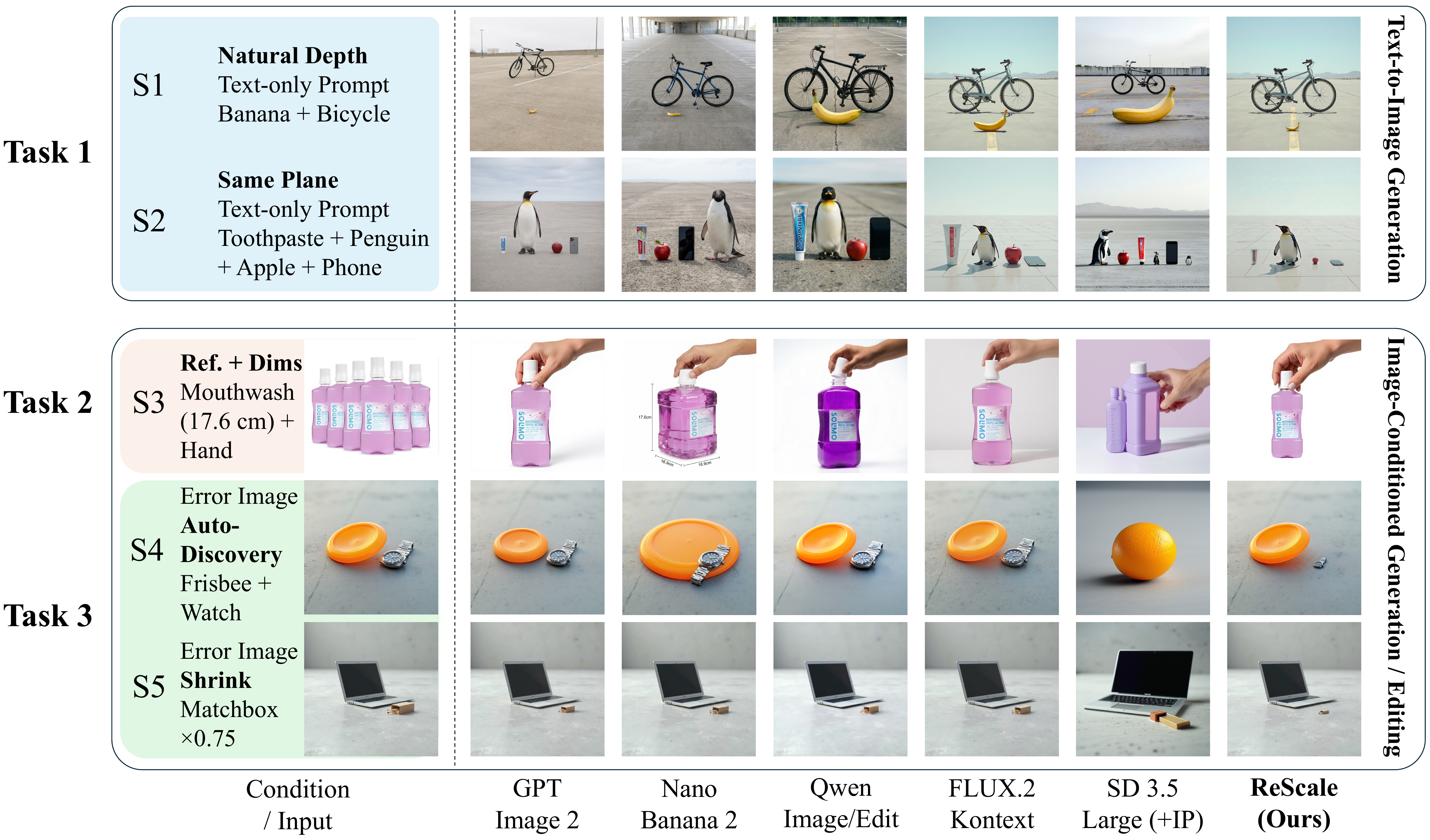}
    \caption{{\textbf{Relative object scale remains a challenge for modern image generators.} S1/S2: common-object generation under natural depth/same-plane layouts; S3: reference-conditioned human--product generation with metric dimensions; S4/S5: source cases for auto-discovery/precise resize instructions; \modelname{} uses evaluator feedback for correction.}
    The general-purpose image generators/editors achieve high visual realism and yet often violate real-world size relationships, whereas \textbf{\modelname{}} produces more plausible relative scale. Zoom in for the best viewing.}
        \label{fig:teaser}
\end{figure}

\begin{abstract}
\label{sec:abstract}
Modern image generation and editing systems can produce photorealistic, prompt-aligned images, but still often render familiar objects at implausible relative sizes. 
To measure this failure mode, we introduce \textbf{\benchmarkname{}}, a benchmark and evaluation protocol for real-world relative object scale in image generation and editing. 
\benchmarkname{} contains 900 image-level entries and 1,643 pairwise anchor-target scale relations across common-object generation, human-product generation with metric dimensions, and scale correction from failed generations. 
We further design a human-calibrated ordinal judge for scalable pairwise scale evaluation. 
Last but not least, we introduce \textbf{\modelname{}}, a model-agnostic post-processing agent for localized scale correction without modifying the source generator. 
Experiments reveal that state-of-the-art image generators and editors {do not reliably render plausible relative scale}, while \modelname{} consistently improves scale plausibility across generated and edited images. 
Together, \benchmarkname{} establishes relative object scale as a distinct, measurable, and actionable capability for image generation systems.
\end{abstract}
\vspace{-2ex}
\section{Introduction}
\label{sec:introduction}
\vspace{-1ex}
Recent text-to-image (T2I) generation has advanced rapidly, with diffusion-based models~\citep{rombach2022high}, autoregressive generators~\citep{yu2022parti}, and transformer-based systems~\citep{esser2024scaling} achieving strong visual fidelity, stylistic diversity, and prompt adherence. 
However, these advances do not guarantee a basic requirement of physical realism: rendering objects at plausible \emph{relative scale}. 
An image may look photorealistic and prompt-aligned while still placing familiar objects at mutually implausible sizes. 
This issue is increasingly important as generated images are used in advertising, product visualization, virtual character creation, and professional content production, where incorrect object proportions can immediately undermine visual credibility.

\begin{wrapfigure}{r}{0.55\textwidth}
    \vspace{-4ex}
    \centering
    \includegraphics[width=\linewidth]{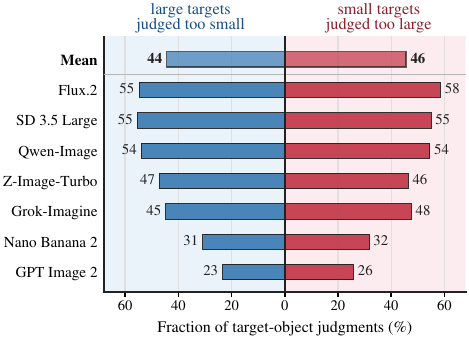}
    \vspace{-0.2in}
    \caption{\textbf{Mean-regression errors on \benchmarkname{} Task 1.}
    Bars show two directional scale failures (400 tests) across generators: large real-world target objects are rendered too small, while small target objects are rendered too large. 
    The consistent pattern indicates that current generators tend to compress object scale toward a typical visual size rather than preserving real-world size ratios.}
    \label{fig:mean_regression}
    \vspace{-3ex}
\end{wrapfigure}

As illustrated in~\cref{fig:teaser}, the failure appears across text-only generation, reference-conditioned human–product generation, and scale correction, where models often preserve identity and visual realism while violating real-world relative scale.
\cref{fig:mean_regression} further shows that these errors are systematic rather than anecdotal: small real-world objects are often rendered too large, while large objects are rendered too small. 
We refer to this recurring pattern as \emph{mean regression in generated object scale}, a failure mode that affects visual realism, product visualization, synthetic data construction, and scale-aware editing.

Existing benchmarks evaluate related capabilities, including object presence, counting, and attribute binding~\citep{ghosh2023geneval}, compositional prompt following~\citep{huang2025t2icompbenchpp}, commonsense plausibility~\citep{fu2024commonsense}, physical reasoning~\citep{meng2024phybench}, and spatial relation control~\citep{wang2025genspace}. 
Yet scale is usually subsumed under broader semantic, physical, or spatial reasoning rather than evaluated as a distinct object-level relation. 
The finer-grained question we study is \emph{relative object scale}: when several objects appear in a single image, \textbf{can a generative model render their size relationships in a way that faithfully reflects the real world?}
This question differs from general physical plausibility because a scene may satisfy coarse commonsense and layout constraints while still assigning an implausible size ratio to two objects. 

This motivates \textbf{\benchmarkname{}}, a benchmark for real-world relative object scale in image generation and editing.
\benchmarkname{} represents scale as a pairwise anchor-target relation grounded in external object-size metadata, and covers three regimes: implicit common-object size priors, human-product metric scale, and post-generation scale correction.
It contains 900 image-level entries and 1,643 anchor-target relations, enabling dense pairwise evaluation while retaining image-level grouping for model comparison.
\cref{tab:benchmark_comparison} summarizes its coverage relative to representative prior benchmarks.

Evaluating these anchor-target relations requires more than measuring pixel-space ratios, since scale depends jointly on object identity, physical size, perspective, depth, occlusion, and placement. 
\benchmarkname{} therefore uses a human-calibrated ordinal evaluation protocol rather than a raw pixel-ratio metric.
We calibrate a Gemini-based judge against annotations from nine human raters, achieving strong agreement with human consensus. 

We further introduce \textbf{\modelname{}}, a model-agnostic size-correction agent that uses structured scale information to execute localized, geometry-aware resizing and reinsertion. Ordinary users can use it to post-correct relative scale issues of any generated or edited images, and model developers may use it to enrich their training data (e.g., identifying hard examples using our calibrated judge, fixing their relative scale using \modelname{}, and mixing them back to the training set). 

In the experiments, we benchmark state-of-the-art image generators and editors, revealing persistent relative-scale failures and showing that \modelname{} consistently improves scale plausibility across generated and edited images.

\textbf{In summary, our contributions are four-fold:}
{
\setlength{\leftmargini}{1.6em}
\setlength{\itemsep}{1pt}
\setlength{\topsep}{2pt}
\begin{itemize}
    \item We introduce \textbf{\benchmarkname{}}, the first benchmark dedicated to relative object scale in image generation and editing, covering common-object scale priors, human-product metric scale, and post-generation scale correction.
    \item We develop a \textbf{human-calibrated ordinal evaluation protocol} for scalable pairwise scale judgment, enabling reliable assessment of whether generated object-object proportions match real-world size relationships.
    \item We propose \textbf{\modelname{}}, a model-agnostic scale-correction agent that uses structured scale metadata to perform localized, geometry-aware editing without modifying the source generator.
    \item We benchmark state-of-the-art image generators and editors, showing that they {do not reliably render plausible relative scale}, while \modelname{} consistently improves scale plausibility across the generated and edited images.
\end{itemize}
}

\begin{table}[t]
\centering
\caption{\textbf{Comparison with representative related benchmarks.}
Prior benchmarks cover individual axes such as composition, physical plausibility, or spatial reasoning, whereas \benchmarkname{} explicitly unifies object-object relative scale, human-product metric size, and post-generation scale correction.}
\label{tab:benchmark_comparison}
\small
\setlength{\tabcolsep}{4.5pt}
\resizebox{0.92\linewidth}{!}{
\begin{tabular}{lcccccc}
\toprule
& \multicolumn{3}{c}{Broad Capabilities} & \multicolumn{3}{c}{\bf Scale-focused Evaluation} \\
\cmidrule(lr){2-4}\cmidrule(lr){5-7}
\textbf{Work} & \textbf{Compos.} & \textbf{Physical} & \textbf{Spatial} & \textbf{Obj.-obj. scale} & \textbf{Human-prod. size} & \textbf{Correction} \\
\midrule
GenEval~\citep{ghosh2023geneval} & \cmark & \xmark & \cmark & \xmark & \xmark & \xmark \\
T2I-CompBench++~\citep{huang2025t2icompbenchpp} & \cmark & \xmark & \cmark & \xmark & \xmark & \xmark \\
GenAI-Bench~\citep{li2024genaibench} & \cmark & \xmark & \xmark & \xmark & \xmark & \xmark \\
PhyBench~\citep{meng2024phybench} & \xmark & \cmark & \xmark & \xmark & \xmark & \xmark \\
GenSpace~\citep{wang2025genspace} & \xmark & \xmark & \cmark & \cmark & \xmark & \xmark \\
\midrule
\textbf{\benchmarkname{}} & \cmark & \cmark & \cmark & \cmark & \cmark & \cmark \\
\bottomrule
\end{tabular}}
\end{table}

\vspace{-1ex}
\section{Related Work}
\label{sec:related_work}
\vspace{-1ex}
\subsection{Image Generative Models}
\vspace{-1ex}
{Image generation has progressed from diffusion, autoregressive, and latent-space models~\citep{Ho20DDPM,Song21Scorebased,ramesh2022hierarchical,saharia2022imagen,rombach2022high,yu2022parti,li2023hae,li2025hypdae} to transformer-based diffusion and rectified-flow systems~\citep{peebles2023scalable,esser2024scaling}, substantially improving visual fidelity, controllability, and scalability.}
Recent open and closed models, including FLUX.2~\citep{blackforestlabs2026flux2}, Qwen-Image~\citep{wu2025qwenimagetechnicalreport,qwen2025qwenimage2512}, Z-Image~\citep{tongyi2025zimage}, Seedream 4.5~\citep{bytedance2025seedream45}, GPT-Image-2~\citep{openai2026gptimage2}, and Nano Banana 2~\citep{google2026gemini3image}, further extend these capabilities to high-fidelity generation, reference conditioning, and image editing.
{Masked-token and autoregressive generators improve visual generation through token prediction, generation order, and patch representation~\citep{chang2022maskgit,sun2024llamagen,pang2025randar,yu2025rar,srivastava2026dpar}. \benchmarkname{} complements them by measuring real-world relative object scale.}
\vspace{-1ex}
\subsection{Evaluation Benchmarks for Text-to-image Generation}
\vspace{-1ex}
Early evaluation relied on distributional quality metrics such as FID and IS~\citep{heusel2017gans,salimans2016improved}, image-text similarity metrics such as CLIPScore~\citep{hessel2021clipscore}, and curated prompt suites such as DrawBench and PartiPrompts~\citep{saharia2022imagen,yu2022parti}.
More targeted benchmarks evaluate compositional generalization~\citep{park2021benchmark}, object presence, counting, color, position, and attribute binding~\citep{ghosh2023geneval}, open-world composition and numeracy~\citep{huang2023t2icompbench,huang2025t2icompbenchpp}, complex prompt following~\citep{li2024genaibench}, reasoning~\citep{chen2025r2ibench}, long-prompt control~\citep{jiao2025detailmaster}, factuality~\citep{huang2025t2ifactualbench}, and world-knowledge-informed semantic evaluation~\citep{niu2025wise}.
Automatic evaluation increasingly uses VLM-based protocols~\citep{hu2023tifa,lin2024evaluating,li2024genaibench}, but recent analyses show that off-the-shelf LMM judges require task-specific validation for nuanced generated-image judgments~\citep{zhang2025abench}.
In contrast, \benchmarkname{} uses a human-calibrated ordinal protocol for scale-specific pairwise judgment.
\vspace{-1ex}
\subsection{Spatial, Physical, and Commonsense Evaluation}
\vspace{-1ex}
{Relative object scale is related to spatial reasoning and physical consistency~\citep{li2024bifrost,li2026visreason,qing2026moire}, but existing benchmarks usually treat it as part of broader capabilities.}
Commonsense-T2I~\citep{fu2024commonsense} evaluates everyday commonsense consistency, PhyBench~\citep{meng2024phybench} targets physical commonsense errors, and Generate Any Scene~\citep{gao2024generate} evaluates structured object and relation control through scene graphs.
Most closely related to our work, GenSpace~\citep{wang2025genspace} benchmarks spatially aware image generation through pose, spatial-relation, and metric-measurement tasks, while recent spatial-intelligence benchmarks emphasize object arrangement and layout consistency~\citep{wang2026everything}.
\benchmarkname{} complements these benchmarks by isolating real-world relative object scale as a pairwise, physically grounded evaluation target, with additional coverage of human-product metric scale and post-generation scale correction.
\vspace{-1ex}
\section{GenScale: A Benchmark for Physically Grounded Relative Scale}
\label{sec:genscale}
\vspace{-1ex}
\benchmarkname{} evaluates physically grounded relative scale as a pairwise anchor-target relation with structured metadata, including object identities, physical reference lengths, expected 3D scale ratios, scenario labels, and product reference images when applicable. 
It contains 900 image-level entries and 1,643 pairwise scale relations across three tasks and five scenarios: common-object generation (Task 1 or {\bf T1}), human-product scale realization ({\bf T2}), and post-generation scale correction ({\bf T3}). 
{Each anchor-target pair is the atomic evaluation unit, with physical-size metadata defining the intended relative scale.} 
\cref{tab:genscale_overview} summarizes the benchmark taxonomy, and \cref{fig:genscale_examples} illustrates the construction pipeline and input conditions detailed in the following.

\begin{table}[t]
\centering
\caption{\textbf{Overview of \benchmarkname{}.} \benchmarkname{} evaluates physically grounded relative scale across implicit common-object priors (T1), explicit metric product scale (T2), and scale correction (T3). Image-level entries may contain multiple anchor-target pairs.}
\label{tab:genscale_overview}
\renewcommand\arraystretch{1.1}
\small
\resizebox{0.92\linewidth}{!}{
\begin{tabular}{llccl}
\toprule
Task & Scenario & Images & Pairs & Capability tested \\
\midrule
T1 & S1: Natural Depth & 200 & 570 & Implicit scale priors under perspective \\
T1 & S2: Same Plane & 200 & 573 & Implicit scale priors with depth controlled \\
T2 & S3: Human-Product & 300 & 300 & Metric product scale with human anchors \\
T3 & S4: Auto-Discovery & 100 & 100 & Diagnose scale error and choose resize \\
T3 & S5: Precise Instruction & 100 & 100 & Execute exact numeric resize factor \\
\midrule
\multicolumn{2}{l}{Total} & 900 & 1,643 & Generation, customization, and correction \\
\bottomrule
\end{tabular}
}
\end{table}

\vspace{-1ex}
\subsection{Scale Metadata}
\vspace{-1ex}
\benchmarkname{} is grounded in two types of physical-size metadata. 
For common-object scale reasoning (T1), we construct a category-level knowledge base from COCO~\citep{Lin14COCO} and LVIS~\citep{Gupta19lvis}, retaining visually identifiable categories with relatively stable physical extent and removing classes with large intra-class variation, ambiguous semantics, or non-rigid size, such as \emph{person}, \emph{dog}, \emph{bird}, \emph{tree}, \emph{bag}, \emph{chair}, and \emph{boat}. 
This pool contains 97 non-human common objects and is used for Task~1 and the common-object correction cases in Task~3. 
For human-product scale (T2), we use product-level dimensions and reference images from Amazon Berkeley Objects (ABO)~\citep{collins2022abo}, together with human anchors including hand, head/face, foot/leg, and full body. 
Object and anchor dimensions are grounded in cross-domain sources, including anthropometric surveys~\citep{gordon2014ansur}, biological and food references~\citep{myers2026adw,usda2026fdc}, standardized manufactured-object specifications~\citep{iso7810_2019,iso216_2007,ansi2021c181m}, and traffic and sport-object specifications~\citep{fhwa2023mutcd,mlb2026officialrules}. 
For each entry, we store a characteristic length, an acceptable physical range when available, and the source used to ground the measurement. Check~\cref{app:benchmark_details} for more details.

\vspace{-1ex}
\subsection{Benchmark Construction}
\vspace{-1ex}

\begin{figure}
    \centering
    \includegraphics[width=0.98\linewidth]{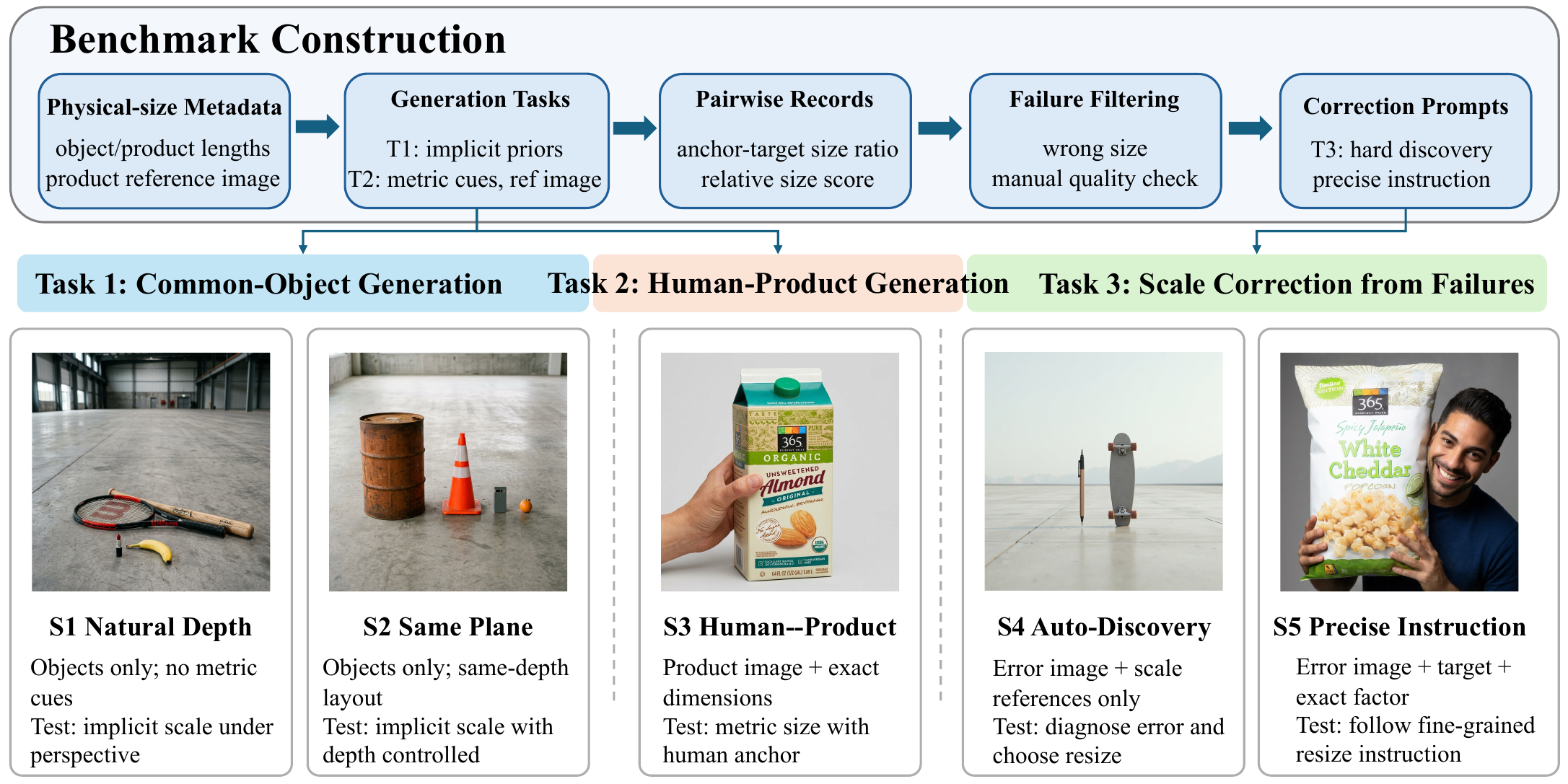}
    \vspace{-1ex}
    \caption{\textbf{\benchmarkname{} benchmark construction.}
    \benchmarkname{} is built from physical-size metadata and product references, then instantiated as three tasks covering common-object generation, human-product metric scale, and scale correction. 
    \textbf{Task~3} converts non-plausible but editable outputs from \textbf{Tasks~1} and~\textbf{2} into S4 Auto-Discovery and S5 Precise Instruction, which separately test automatic scale-error diagnosis and exact numeric resize following.}
    \label{fig:genscale_examples}
\end{figure}

As shown in \cref{fig:genscale_examples}, the input to image generators differs by tasks: An input in Task~1 consists of only a text prompt, Task~2 uses a product reference image plus a text prompt, and Task~3 pairs an erroneous source image with one of two edit prompts.

\paragraph{Task 1: Implicit Common-Object Scale.}

Task~1 is text-only image generation that tests whether models can infer relative sizes of familiar objects without explicit metric cues. 
Each image-level entry contains a  prompt with two to four non-human objects sampled from the common-object pool; no object-specific numeric dimensions or reference images are given to the model. 
We retain object sets whose largest-to-smallest characteristic-length ratio lies in $[4,20]$, avoiding near-equal comparisons that are hard to judge and extreme ratios where layout constraints dominate. 
The generated image is then evaluated through all valid anchor-target pairs induced by the objects in the prompt.

Task~1 has two scenarios. 
In \textbf{S1: Natural Depth}, objects may appear with realistic perspective, mild occlusion, and depth ordering, reflecting ordinary generation settings where scale errors can be hidden by near-far placement. 
In \textbf{S2: Same Plane}, objects are constrained to approximately the same depth plane, reducing perspective ambiguity and making image-space proportions more directly reflect real-world scale ratios.

\vspace{-2ex}
\paragraph{Task 2: Explicit Metric Scale in Human-Product Interaction.}

Task~2 is an image-conditioned human-product generation task. 
Each entry contains a product reference image and a prompt specifying the product's metric dimensions and a size-conditioned human anchor, requiring the model to preserve product identity while rendering plausible scale relative to the human body. 
This tests product-level scale realization, where category priors are insufficient because visually similar commercial products can have different dimensions. 
Products shorter than 30 cm are paired with a hand, products between 30 and 60 cm with a head/face anchor, products between 60 and 100 cm with a foot/leg anchor, and products larger than 100 cm with a full-body anchor. 
Prompts require natural, use-consistent interaction and catalog-style visibility, so generators must translate metric dimensions into plausible human-object proportions rather than merely copy the reference appearance.

\vspace{-1ex}
\paragraph{Task 3: Scale Correction from Failed Generations.}

Task~3 converts failed generations from Tasks~1 and~2 into scale-aware image editing tasks. 
Each source case contains an erroneous image and one pairwise relation whose relative scale is judged implausible under the pairwise criterion introduced in~\cref{sec:evaluation}. 
We use FLUX.2~\citep{blackforestlabs2026flux2} generations for common-object scenes and Qwen-Image~\citep{wu2025qwenimagetechnicalreport} generations for human-product scenes, retaining only pairwise cases: two-object Task~1 images and one product-human relation in Task~2. 
The corresponding pairwise record provides the rendered object-size ratio and target physical ratio, from which we derive the fixed reference object, editable object, resize direction, and multiplicative scale factor. 
We remove images with missing or unidentifiable objects, duplicated or merged objects, severe artifacts, heavy occlusion or blur, and extreme correction factors that make localized editing ill-defined; this filtering only ensures that each retained source image contains a scorable and editable scale error.

Each of the 100 retained source images is expanded into two benchmark entries with the same erroneous image but different edit prompts. 
In \textbf{S4: Hard Auto-Discovery}, the model receives the source image and scale references without a prescribed resize direction or factor and must decide whether and how to correct the scale error.
For common-object sources, the model also chooses the editable object. For human--product sources, the prompt identifies the product as editable and keeps the human anchor fixed.
In \textbf{S5: Precise Scale Instruction}, the prompt directly gives the editable object, reference object, resize direction, and exact scale factor, isolating fine-grained numeric resize following from visual error diagnosis.

\vspace{-1ex}
\subsection{Human-Calibrated Scale Evaluation}
\label{sec:evaluation}
\vspace{-1ex}

\benchmarkname{} evaluates relative scale at the object-pair level. 
A raw object-space ratio is insufficient because the same apparent ratio may be plausible or implausible depending on object identity, depth ordering, foreshortening, occlusion, and camera perspective. 
We therefore develop a human-calibrated ordinal protocol: Each generated image is decomposed into anchor--target pairs from the \benchmarkname{} metadata; the anchor is treated as the reference, and the evaluator judges whether the target has a plausible real-world size relative to it.

\begin{wrapfigure}{r}{0.5\textwidth}
    \vspace{-2ex}
    \centering
    \includegraphics[width=\linewidth]{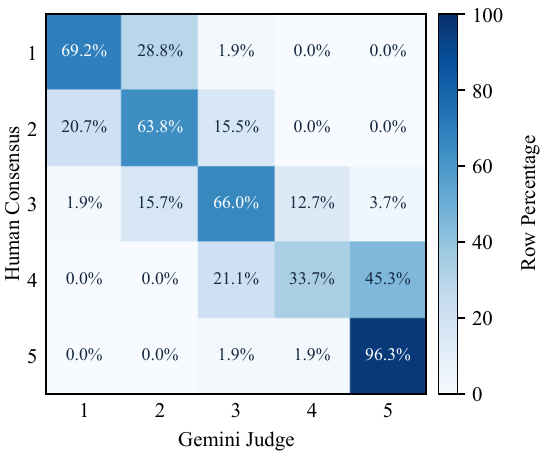}
    \vspace{-0.2in}
    \caption{\textbf{Disagreement structure of the calibrated Gemini judge.}
    The row-normalized confusion matrix shows that Gemini-human disagreements are concentrated near the diagonal, with {$3.23\%$} of pairs differing by two or more score levels and no coarse under-to-over or over-to-under sign-flip errors.}
    \label{fig:gemini_human_heatmap}
    \vspace{-1.0ex}
\end{wrapfigure}

\vspace{-2ex}
\paragraph{Pairwise Ordinal Rubric.}
For each anchor-target pair, evaluators are shown the image, object names, and reference physical lengths. 
They assign a five-point ordinal score: 1/2 indicate that the target is severely/slightly undersized, 3 indicates physically plausible scale, and 4/5 indicate that it is slightly/severely oversized. 
Operationally, score 3 corresponds to an estimated scale error within approximately $\pm20\%$, scores 2/4 to errors between $20\%$ and $60\%$, and scores 1/5 to errors larger than $60\%$. 
Evaluators are instructed to account for perspective, occlusion, partial visibility, and foreshortening. 
A pair is marked invalid only when reliable scale judgment is impossible, e.g., because an object is missing, merged, ambiguously duplicated, or too degraded to identify.

\begin{table}[t]
\centering
\caption{\textbf{Human Reliability and Gemini-Human Alignment.}
Exact and $\leq 1$ denote agreement on the five-point ordinal scale defined as exact match and difference by at most 1, respectively.
MAE is the mean absolute ordinal deviation. $r$ is Pearson correlation on 1-5 scores.
QWK denotes quadratic-weighted kappa. Human rows report mean$\pm$std across nine annotators.}
\label{tab:human_gemini_alignment}
\vspace{-1ex}
\renewcommand\arraystretch{1.1}
\small
\resizebox{\linewidth}{!}{
\begin{tabular}{llccccc}
\toprule
Evaluator / Reference 
& Split 
& Exact $\uparrow$ 
& $\leq 1$ $\uparrow$ 
& MAE $\downarrow$ 
& $r$ $\uparrow$ 
& QWK $\uparrow$ \\
\midrule
Human vs. Aggregate Consensus 
& T1+T2 
& $65.1{\pm}7.0$ 
& $94.0{\pm}3.2$ 
& $0.413{\pm}0.099$ 
& $0.764{\pm}0.076$ 
& $0.754{\pm}0.077$ \\
Human vs. LOO Consensus 
& T1+T2 
& $58.0{\pm}6.6$ 
& $93.3{\pm}3.0$ 
& $0.491{\pm}0.093$ 
& $0.719{\pm}0.069$ 
& $0.706{\pm}0.070$ \\
\midrule
Gemini vs. Human Consensus 
& T1+T2 
& {$63.38$} 
& {$96.77$} 
& {$0.398$} 
& {$0.832$} 
& {$0.820$} \\
Gemini vs. Human Consensus 
& T1 
& {$61.12$} 
& {$96.02$} 
& {$0.429$} 
& {$0.846$} 
& {$0.834$} \\
Gemini vs. Human Consensus 
& T2 
& $73.00$ 
& $100.00$ 
& $0.270$ 
& $0.470$ 
& $0.468$ \\
\bottomrule
\end{tabular}
}
\end{table}

\vspace{-2ex}
\paragraph{Human Consensus.}
We collect annotations from nine human raters on a calibration split sampled from Tasks~1 and~2, covering natural-depth common-object scenes, same-plane common-object scenes, and human-product scenes. 
After aligning completed annotations and retaining visible, scorable pairs, the calibration set contains 285 images and 527 anchor-target pairs. {The 4,523 individual ratings yield ordinal Krippendorff's $\alpha=0.617$ (95\% CI [0.580, 0.651]) and split-half consensus correlation of 0.805.} 
For each pair, we define human consensus as the modal score, with ties broken by the median. 
{\cref{tab:human_gemini_alignment} shows strong agreement with human consensus under both aggregate and leave-one-rater-out comparisons.}

\vspace{-2ex}
\paragraph{Automatic Judge Calibration.}
\leavevmode{We calibrate Gemini 3.1 Pro Preview~\citep{google2026gemini31pro} against human consensus for scalable pairwise evaluation.}
The judge receives the same core evidence as human raters---the generated image, anchor and target names, physical reference lengths, the expected 3D length ratio, and the five-point rubric.
The generation prompt, when provided, is used only to disambiguate intended objects or layout, not as evidence that the rendered scale is correct.
We use scenario-specific judge prompts, with Task~3 inheriting the judge type of its source case, and provide the full prompts in~\cref{app:human_vlm_eval}.
Before scoring scale, we first check whether the required objects are visible, identifiable, and unambiguous, following the invalid-pair rule above; failed pairs are omitted from valid-pair counts and scale metrics, whereas visible but incorrectly scaled pairs are still scored.
For each remaining pair, we query the judge five times and aggregate scores by majority vote, using the median when no unique mode exists, which reduces sensitivity to isolated outlier judgments.
\cref{tab:human_gemini_alignment} shows that the calibrated judge reaches inter-human-level alignment with the consensus, with especially strong within-one agreement and QWK on the full calibration split.
\cref{fig:gemini_human_heatmap} further shows that remaining disagreements are ordinally local rather than directionally reversed.

We therefore use the calibrated Gemini judge as the default evaluator for large-scale \benchmarkname{} comparisons and report valid image and pair counts for all systems.

\begin{table}[t]
\centering\small
\caption{{\textbf{Cross-vendor judge validation.} Agreement with human consensus on the same 527 relations. All judges use the rubric calibrated for Gemini 3.1 Pro Preview, without model-specific tuning for alternatives. QWK is quadratic-weighted kappa; the human reference pools leave-one-rater-out comparisons.}}
\label{tab:new_judges}
\begingroup
\begin{tabular}{@{}lcc@{}}
\toprule Judge & Pearson $r\uparrow$ & QWK $\uparrow$ \\
\midrule
Gemini 3.1 Pro Preview & 0.832 & 0.820 \\
Claude Opus 5 & 0.798 & 0.775 \\
GPT-5.6 Sol & 0.784 & 0.764 \\
Gemini 3.6 Flash & 0.731 & 0.638 \\
Pooled Human Leave-One-Out & 0.728 & 0.727 \\
\bottomrule\end{tabular}
\endgroup
\end{table}
{Agreement across vendors supports transfer of the scale rubric beyond the default judge. Ratio elicitation also enables open-weight scale evaluation; full analyses appear in~\cref{app:new_judges,app:new_open}.}

\vspace{-1ex}
\section{\modelname{}: Agentic Relative-Scale Correction}
\label{sec:rescale}

\begin{figure}
    \centering
    \includegraphics[width=\linewidth]{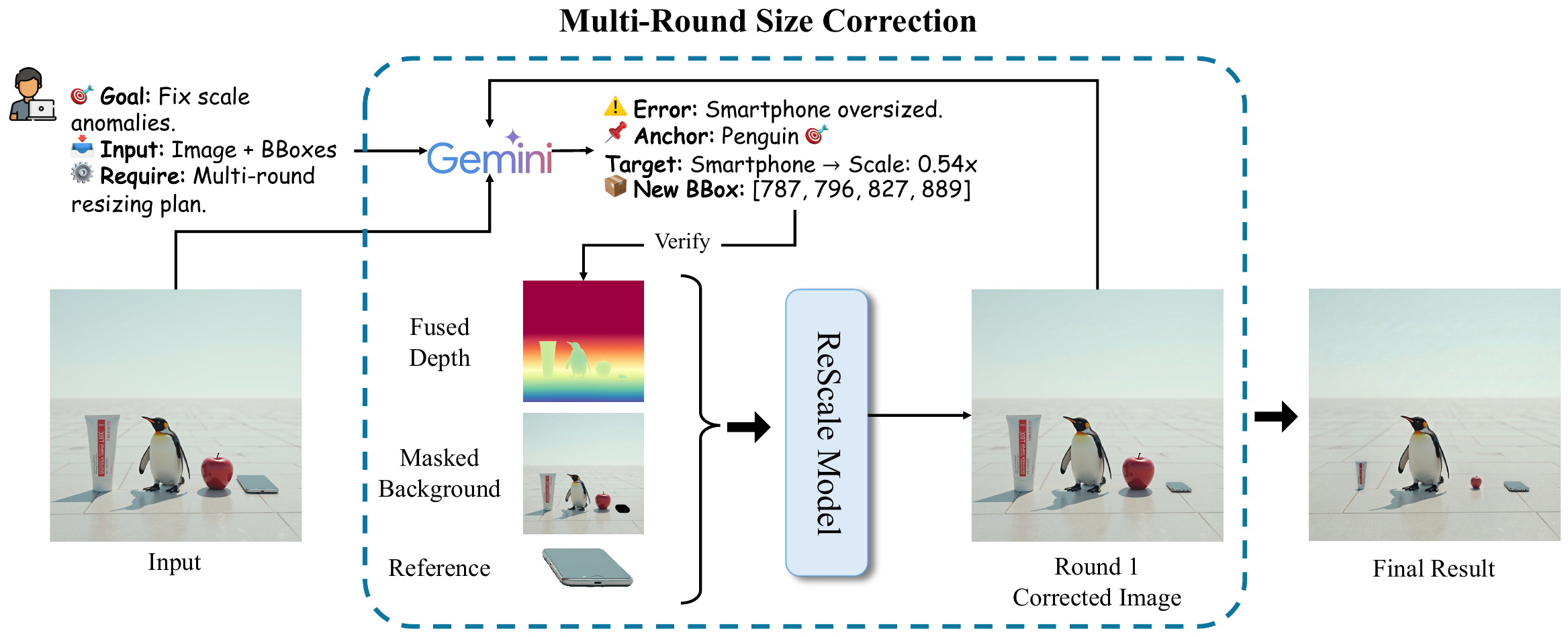}
    \vspace{-0.5ex}
    \caption{\textbf{Agentic multi-round inference in \modelname{}.}
    Given a generated image, object/scale metadata, and {evaluator feedback in the main experiments}, the agent identifies a scale anomaly, selects the target and anchor, predicts a resize factor and contact-preserving anchor point, and executes localized correction through extraction, background completion, mask/depth adjustment, and modular insertion. 
    The edited result is verified for additional correction rounds when needed.}
    \label{fig:rescale_inference}
\end{figure}

\vspace{-1ex}
The \benchmarkname{} formulation makes relative-scale errors actionable beyond evaluation: 
We introduce \textbf{\modelname{}}, a model-agnostic post-processing pipeline that repairs scale inconsistencies in generated images without modifying the source generator. 
{Given an image and physical-size metadata, \modelname{} corrects localized scale errors while preserving surrounding content. The main experiments additionally provide evaluator feedback containing object identities, relation scores, and explanations to guide edit selection and planning. Fully autonomous diagnosis is evaluated in~\cref{app:new_ablation}.}

At inference time, a multimodal agent grounds relevant objects and converts pairwise scale evidence into an edit plan. 
For common-object scenes (T1), it aggregates inconsistent anchor-target relations into a conservative object-level plan and edits over multiple rounds; for human-product scenes (T2), the product is the only editable target while the human body part is fixed. 
For each edit, the agent selects a target $o_t$ and anchor $o_a$, refines the target box $b_t$, and estimates a resize factor $\alpha$ by comparing the apparent target-anchor ratio with the plausible real-world ratio under perspective, depth, visibility, contact, boundary, and collision constraints. 
The plan $\pi=\{o_t,o_a,b_t,\alpha,p\}$ specifies the target, anchor, box, resize factor, and contact-preserving anchor point $p$; if the multimodal agent detects no reliable scale inconsistency or finds the edit unsafe, \modelname{} returns no edit.

The planned edit is executed by a modularized, local insertion pipeline shown in Figure~\ref{fig:rescale_inference}. 
We segment the target with SAM~2~\citep{ravi2024sam2}, extract it as an identity reference, remove the original instance to obtain a completed background, construct an enlarged edit mask around the resized box, and estimate monocular depth with DepthAnythingV2~\citep{yang2024depthanythingv2}. 
These steps produce backend-agnostic conditions, including the reference crop, completed background, resized mask, text instruction, and, when supported, fused depth. 
Our default backend is InsertAnything~\citep{song2025insertanything}, but it can be replaced by other insertion or inpainting editors. 
After each edit, the agent verifies whether an obvious scale error remains and applies another round of editing only when needed. \cref{app:rescale_details} describes \modelname{} and some ablation studies in detail. 
\begin{table}
\centering
\begingroup
\caption{\textbf{Task 1: Common-Object Relative Scale.}
Results are reported for S1, S2, and both. 
Err. is scale error, mean $|s-3|$; Plaus.\ is the percentage of plausible relations (score $s=3$); MR is mean-regression error, where small targets are judged too large, or vice versa.}
\label{tab:task1_model_benchmark}
\renewcommand\arraystretch{0.96}
\setlength{\tabcolsep}{3.2pt}
\small
\resizebox{0.92\linewidth}{!}{
\begin{tabular}{@{}lccc ccc ccc@{}}
\toprule
\multirow{2}{*}{Model}
& \multicolumn{3}{c}{S1}
& \multicolumn{3}{c}{S2}
& \multicolumn{3}{c}{S1 + S2} \\
\cmidrule(lr){2-4}\cmidrule(lr){5-7}\cmidrule(l){8-10}
& Err. $\downarrow$ & Plaus. $\uparrow$ & MR $\downarrow$
& Err. $\downarrow$ & Plaus. $\uparrow$ & MR $\downarrow$
& Err. $\downarrow$ & Plaus. $\uparrow$ & MR $\downarrow$ \\
\midrule
Nano Banana 2~\citep{google2026gemini3image}
& 0.51 & 65.1 & 15.8 & 0.69 & 47.8 & 47.9 & \textbf{0.60} & \textbf{56.5} & 31.8 \\
GPT-Image-2~\citep{openai2026gptimage2}
& 0.78 & 51.1 & 12.1 & 0.63 & 52.8 & 35.8 & 0.71 & 51.9 & \textbf{24.0} \\
Z-Image-Turbo~\citep{tongyi2025zimage}
& 0.64 & 53.4 & 35.7 & 0.88 & 38.1 & 55.7 & 0.75 & 46.2 & 45.1 \\
Grok Imagine~\citep{xai2026grokimage}
& 0.60 & 58.7 & 26.9 & 0.98 & 34.7 & 61.1 & 0.79 & 46.6 & 44.1 \\
Qwen-Image 2512~\citep{qwen2025qwenimage2512}
& 0.70 & 53.1 & 37.5 & 1.10 & 27.0 & 67.6 & 0.91 & 39.8 & 52.9 \\
FLUX.2~\citep{blackforestlabs2026flux2}
& 0.83 & 45.5 & 40.4 & 1.21 & 23.3 & 69.4 & 1.03 & 33.9 & 55.5 \\
SD3.5-Large~\citep{stabilityai2024sd35}
& 0.96 & 36.7 & 46.7 & 1.19 & 24.5 & 70.0 & 1.07 & 31.0 & 57.7 \\
\bottomrule
\end{tabular}}\par
\endgroup
\end{table}

\begin{figure}[t]
\centering
\resizebox{0.98\linewidth}{!}{%
\begin{minipage}{\linewidth}
\centering
\begin{adjustbox}{valign=c}
\includegraphics[width=0.62\linewidth]{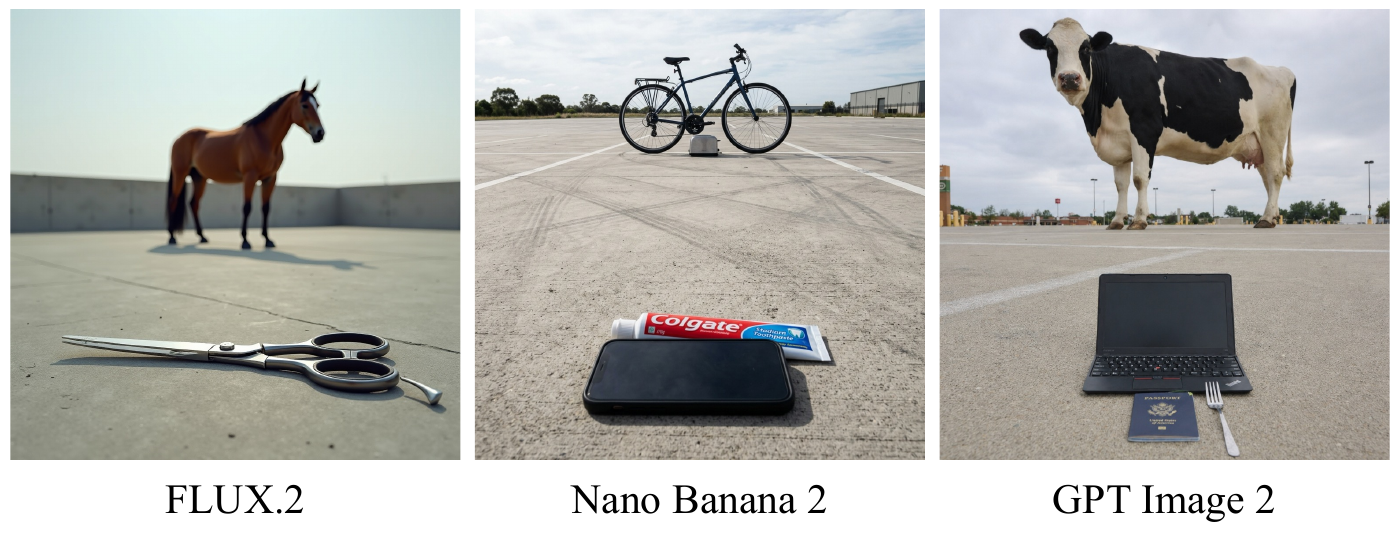}
\end{adjustbox}
\hfill
\begin{adjustbox}{valign=c}
\resizebox{0.37\linewidth}{!}{
\begin{tabular}{@{}lcc@{}}
\toprule
Size bucket & Objects & Depth rank $\uparrow$ \\
\midrule
Smallest & 1,367 & 0.175 \\
Middle   & 1,039 & 0.451 \\
Largest  & 1,362 & 0.863 \\
\midrule
Smaller closer & \multicolumn{2}{c}{79.2\%} \\
$\rho(\log \ell,\mathrm{depth})$ & \multicolumn{2}{c}{0.558} \\
\bottomrule
\end{tabular}
}
\end{adjustbox}
\end{minipage}%
}
\caption{\textbf{Depth-mediated scale compression in Task 1 S1.}
S1 examples show that image generators often place smaller objects closer and larger objects farther away. In the table, we group objects according to real-world size $l$ and then compute their depth ranks in images averaged per group.}
\label{fig:s1_depth_size}
\end{figure}

\section{Evaluation Results}
\label{sec:evaluation_results}
\vspace{-1ex}
\subsection{Benchmarking State-of-the-Art Models on \benchmarkname{}}
\label{sec:model_benchmarking}
\vspace{-1ex}

\begin{table}
\centering
\caption{\textbf{Task 2: Human-Anchored Product Scale.}
Each prompt contains one product-human scale relation. 
Scale error is mean $|s-3|$; Plausible denotes score $s=3$; Severe denotes scores of 1 or 5. 
Full error breakdown, including too-small and too-large rates, is reported in~\cref{app:full_results}}
\label{tab:task2_model_benchmark}
\renewcommand\arraystretch{0.96}
\setlength{\tabcolsep}{6.0pt}
\small
\resizebox{0.98\linewidth}{!}{
\begin{tabular}{@{}lcccc@{}}
\toprule
Model 
& Valid img. / pairs 
& Scale error $\downarrow$
& Plausible (\%) $\uparrow$
& Severe (\%) $\downarrow$ \\
\midrule
GPT-Image-2~\citep{openai2026gptimage2}
& 294 / 294
& \textbf{0.231}
& \textbf{78.9}
& \textbf{2.0} \\

Nano Banana 2~\citep{google2026gemini3image}
& 295 / 295
& 0.268
& 75.6
& 2.4 \\

Seedream 4.5~\citep{bytedance2025seedream45}
& 295 / 295
& 0.302
& 74.9
& 5.1 \\

Qwen-Image-Edit-2511~\citep{qwen2025qwenimageedit2511}
& 297 / 297
& 0.327
& 71.0
& 3.7 \\

FLUX.1 Kontext-dev~\citep{blackforestlabs2025fluxkontext}
& 291 / 291
& 0.423
& 63.2
& 5.5 \\

SD3.5-Large + IP-Adapter~\citep{stabilityai2024sd35,ye2023ipadapter}
& {269 / 269}
& {0.595}
& {52.8}
& {12.3} \\
\bottomrule
\end{tabular}}
\end{table}

We benchmark representative open- and closed-source generative/editing models using the calibrated judge from~\cref{sec:evaluation}. 
All metrics are computed on valid pair-level judgments after the visual-quality filter (see~\cref{sec:evaluation}), and the scale metrics measure performance on those valid judgments: For score $s\in\{1,\ldots,5\}$, with 3 denoting the plausible scale, \emph{Scale error} is mean $|s-3|$.

\vspace{-2ex}
\paragraph{Task 1: Common-Object Relative Scale.}
Nano Banana 2~\citep{google2026gemini3image} leads the common-object benchmark with a 56.5\% plausible rate (\cref{tab:task1_model_benchmark}).
Mean regression, in which small objects appear too large and large objects too small, is more frequent in S2 than S1 for all seven models.
Six of seven models achieve lower scale error and higher plausible rates in S1 (natural depth) than in S2 (same plane), consistent with perspective and depth ordering masking some relative-scale errors.
\cref{fig:s1_depth_size} supports this interpretation, showing that generators tend to place smaller objects closer and larger objects farther away.
Thus, current generators often absorb unrealistic scale ratios through layout choices instead of preserving real-world object-size relations. {Consistent with this pattern, smaller objects are placed closer to the camera in 79.2\% of S1 comparisons.}

\vspace{-2ex}
\paragraph{Task 2: Human-Anchored Product Scale.}
\cref{tab:task2_model_benchmark} indicates that explicit metric cues and human anchors make scale realization easier, but not solved. 
Most models achieve higher plausible rates than in Task~1, but the errors are strongly asymmetric: Products are more often enlarged than shrunk. 
This product-magnification bias can practically hinder the image generators' applications in advertising, virtual reality, and others.

\vspace{-2ex}
\paragraph{Task 3: Scale-Error Correction.}
\cref{tab:task3_editing_results} shows that general-purpose image editors can reduce scale errors on failed generations to very limited degrees. 
The S4-S5 split reveals a clear diagnosis-execution gap: models perform much better when given the target object, reference object, resize direction, and scale factor, but they are substantially weaker when they must discover the scale error and choose the correction autonomously. 
Hence, these general-purpose image editors can follow explicit localized resize instructions more reliably than they can infer physically implausible scale.

\begin{table}
\begingroup
\centering
\caption{\textbf{Task 3: Scale-Error Correction.}
Results are reported for S4, S5, and both. 
S4 requires automatic error discovery; S5 provides the target object, reference object, correction direction, and scale factor. 
Gain is the matched-pair reduction in scale error relative to the before-edit image. 
B / W denotes the number of pairs with better / worse scale score after editing. Full results in~\cref{app:task3_full_results}.}
\label{tab:task3_editing_results}
\renewcommand\arraystretch{0.96}
\setlength{\tabcolsep}{3.2pt}
\small
\resizebox{0.98\linewidth}{!}{
\begin{tabular}{@{}lccc ccc ccccc@{}}
\toprule
\multirow{2}{*}{Model}
& \multicolumn{3}{c}{S4}
& \multicolumn{3}{c}{S5}
& \multicolumn{5}{c}{S4 + S5} \\
\cmidrule(lr){2-4}\cmidrule(lr){5-7}\cmidrule(l){8-12}
& Err. $\downarrow$ & Plaus. $\uparrow$ & Gain $\uparrow$
& Err. $\downarrow$ & Plaus. $\uparrow$ & Gain $\uparrow$
& Valid
& Err. $\downarrow$ & Plaus. $\uparrow$ & Gain $\uparrow$ & B / W \\
\midrule
Before edit & 1.27 & 20.4 & -- & 1.27 & 20.4 & -- & 196 / 196 & 1.27 & 20.4 & -- & -- \\
GPT-Image-2 & 0.96 & 40.2 & +0.33 & 0.41 & 66.3 & +0.87 & 195 / 195 & \textbf{0.68} & \textbf{53.3} & \textbf{+0.60} & 91 / 9 \\
Nano Banana 2 & 1.01 & 32.0 & +0.24 & 0.93 & 33.7 & +0.34 & 195 / 195 & 0.97 & 32.8 & +0.29 & 55 / 11 \\
FLUX.1 Kontext-dev & 1.29 & 21.2 & -0.02 & 0.96 & 36.2 & +0.28 & 193 / 193 & 1.13 & 28.5 & +0.13 & 35 / 16 \\
Qwen-Image-Edit-2511 & 1.24 & 23.5 & +0.02 & 1.00 & 34.1 & +0.25 & 189 / 189 & 1.13 & 28.6 & +0.13 & 32 / 9 \\
Seedream 4.5 & 1.17 & 28.0 & +0.10 & 1.10 & 35.4 & +0.20 & 196 / 196 & 1.14 & 31.6 & +0.15 & 43 / 17 \\
\makecell[l]{SD3.5-Large +\\IP-Adapter} & 0.86 & 42.9 & -0.07 & 1.31 & 22.0 & -0.16 & 73 / 73 & 1.22 & 26.0 & -0.14 & 10 / 18 \\
\bottomrule
\end{tabular}}
\vspace{0.2ex}
\begin{flushleft}
\footnotesize
{SD3.5-Large results cover 14 S4 and 59 S5 outputs; valid coverage is reported alongside scale accuracy.}
\end{flushleft}
\endgroup
\end{table}

Overall, \benchmarkname{} exposes three distinct failure modes: mean regression in common-object generation, product magnification in human-product generation, and weak autonomous diagnosis in scale correction. 
These results support the central claim that relative scale is not subsumed by visual fidelity or prompt adherence, but remains a separate unsolved capability.

\vspace{-1ex}
\subsection{\modelname{} Correction Results}
\label{sec:rescale_results}

\vspace{-1ex}
We evaluate \modelname{} on image pairs before and after applying \modelname{} using the calibrated \benchmarkname{} judge. 
\cref{tab:rescale_task1,tab:rescale_task23} shows that \modelname{} reduces scale errors for every Task~1 and Task~2 source model, suggesting that common-object mean-regression and human-product metric-scale errors are often locally correctable. 
{On Task~3 source images, feedback-guided \modelname{} yields the largest correction gain, demonstrating how evaluator information can guide structured local editing.} 
Overall, \benchmarkname{}'s physical-size structure is not only evaluative but actionable.

\begin{table}[t]
\centering\normalsize
\caption{{\textbf{Matched correction results.} Each block uses relations scorable before and after editing. T1/T2 apply \modelname{} to each source model; T3 reports general-purpose editors under S4/S5 prompts and feedback-guided \modelname{} once on 93 unique source relations. B/A/G = error before/after/gain. Qwen and FLUX denote Qwen-Image 2512 and FLUX.2 in T1, and Qwen-Image-Edit-2511 and FLUX.1 Kontext-dev in T2/T3. Dashes denote untested settings.}}
\label{tab:rescale_task1}\label{tab:rescale_task23}
\begingroup
\renewcommand{\arraystretch}{1.08}
\setlength{\tabcolsep}{4.5pt}
\begin{tabular}{@{}lccc ccc ccc@{}}
\toprule
&\multicolumn{3}{c}{T1: common objects}&\multicolumn{3}{c}{T2: human--product}&\multicolumn{3}{c}{T3: correction}\\
\cmidrule(lr){2-4}\cmidrule(lr){5-7}\cmidrule(l){8-10}
Source/editor & B$\downarrow$&A$\downarrow$&G$\uparrow$&B$\downarrow$&A$\downarrow$&G$\uparrow$&B$\downarrow$&A$\downarrow$&G$\uparrow$\\
\midrule
Nano Banana 2 & 0.592&0.384&+0.208&.261&.086&+.175&1.255&0.964&+0.292\\
GPT-Image-2 & 0.688&0.443&+0.245&.231&.090&+.141&1.275&0.674&+0.601\\
Z-Image & 0.759&0.431&+0.328&--&--&--&--&--&--\\
Grok Imagine & 0.784&0.447&+0.337&--&--&--&--&--&--\\
Qwen & 0.918&0.422&+0.497&.306&.144&+.162&1.255&1.122&+0.133\\
FLUX & 1.020&0.524&+0.496&.406&.228&+.178&1.247&1.121&+0.126\\
SD3.5 & 1.045&0.630&+0.415&.565&.256&+.309&1.056&1.197&-0.141\\
Seedream 4.5 & --&--&--&.295&.168&+.126&1.280&1.130&+0.150\\
\modelname{} &--&--&--&--&--&--&1.258&0.548&+0.710\\
\bottomrule\end{tabular}
\endgroup
\end{table}

\begin{table}[t]
\centering\normalsize
\caption{{\textbf{Human evaluation of scale correction.} Four raters; 112 source/edited pairs with valid consensus in both images. Scale error is mean $|s-3|$, and gain is its paired reduction. Intervals are 95\% bootstrap confidence intervals.}}
\label{tab:new_humans}
\begingroup
\renewcommand{\arraystretch}{1.08}
\setlength{\tabcolsep}{5pt}
\begin{tabular}{@{}lccccc@{}}
\toprule
Split & $n$ & Error B/A$\downarrow$ & Gain$\uparrow$ & 95\% CI & Plausible B/A (\%)$\uparrow$\\
\midrule
Representative &77&1.104/.130&.974&[.805,1.130]&11.7/88.3\\
Challenge &35&.829/.371&.457&[.143,.771]&37.1/65.7\\
\bottomrule\end{tabular}
\endgroup
\end{table}

\paragraph{Scale Improvement and Visual-quality Preservation.}
{The human evaluation in~\cref{tab:new_humans} confirms scale improvement in both representative and challenging cases. High image similarity and modest changes in automatic quality scores provide complementary evidence of visual preservation (\cref{tab:app_rescale_quality_full}).}

{The Task~1 correction analysis further shows that erroneous relations are repaired much more frequently than plausible relations are disturbed. The resulting net increase in plausible relations supports feedback-guided local correction (\cref{app:new_overshoot}).}

\begin{samepage}
\paragraph{Components and Difficult Conditions.}
{Depth conditioning strengthens structural reliability, and evaluator feedback increases correction coverage (\cref{tab:new_structure,tab:new_autonomy}). Positive gains in cluttered scenes and with high contrast, elongated objects, and transparent or specular materials show that localized correction extends across varied scene conditions (\cref{app:new_hard}).}
\par
\end{samepage}

\vspace{-1ex}
\section{Conclusion}
\label{sec:conclusion}
\vspace{-1ex}
In summary, we introduced \benchmarkname{}, a benchmark for relative object scale in image generation and editing, covering common-object scale priors, human-product metric scale, and scale correction from failed generations.
We developed a human-calibrated ordinal evaluator that grounds pairwise scale judgments in object-size metadata, enabling scalable assessment beyond raw pixel ratios.
We further introduced \modelname{}, a model-agnostic post-generation correction agent that uses structured scale information for localized geometry-aware editing.
Experiments on contemporary open and closed models show that relative scale remains unreliable, while \modelname{} consistently improves scale plausibility.
Together, these results establish relative object scale as a distinct, measurable, and actionable dimension of physical realism.

\vspace{-2ex}
\paragraph{Limitations and Future Work.}
\benchmarkname{} targets relative object scale rather than general physical or spatial realism, and its knowledge base covers visually identifiable categories with stable physical dimensions. 
It therefore excludes deformable, fine-grained, or context-dependent objects, and needs broader validation across viewpoints, occlusions, domains, and future model families. 
\modelname{} assumes scale-relevant objects are visible and locally editable; future work could incorporate metric size priors or scale-aware preference learning directly into generative models.
\clearpage
\section*{Acknowledgements and Disclosures}
The work was supported in part by NSF \#2540851.
We thank Hanna, Lawrence, Manaswi, Saniya, Tuan, Visista, and Puchen for their annotation work.
We thank Creatify AI for providing computing resources.

{\small
\bibliographystyle{natbib}
\bibliography{reference}
}

\newpage
\appendix
\newpage
\section*{\huge Supplementary Material}
\section*{Overview}
This appendix is organized as follows:

\cref{app:benchmark_details} provides additional details of \benchmarkname{} construction, including physical-size metadata, object and product filtering, prompt construction, task-specific sampling rules, Task~3 failure-case selection, and the released metadata format. This section supports Sec.~3.1 and Sec.~3.2 of the main paper.

\cref{app:human_vlm_eval} gives full details of the human-calibrated scale evaluation protocol, including the annotation interface, pairwise ordinal rubric, exact scene-prefilter and task-specific Gemini judge prompts, repeated-query aggregation, validity criteria, calibration diagnostics, and bootstrap confidence intervals. This section supports Sec.~3.3 of the main paper.

\cref{app:rescale_details} describes \modelname{} in more detail, including agentic scale diagnosis, edit-plan construction, object localization, segmentation, background completion, depth conditioning, local insertion, insertion-backend training, and ablation studies. This section supports Sec.~4 of the main paper.

\cref{app:full_results} reports the complete quantitative results for GenScale Tasks~1--3 and \modelname{} correction, including full score distributions, directional error breakdowns, valid-pair coverage, matched before/after metrics, statistical confidence intervals, and quality-preservation metrics. This section supports Sec.~5.1 and Sec.~5.2 of the main paper.

\cref{app:qualitative} presents additional qualitative examples across all GenScale tasks, including generations from evaluated models, correction results from general-purpose editors, \modelname{} before/after examples, and representative failure cases.

\cref{app:genspace_comparison} provides a direct comparison between \benchmarkname{} and GenSpace, clarifying how \benchmarkname{} differs in its focus on real-world relative object scale, human-product metric scale, and post-generation scale correction.

\section{\benchmarkname{} Benchmark Details}
\label{app:benchmark_details}

\subsection{Physical-Size Knowledge Base}
\label{app:physical_size_kb}

\benchmarkname{} is grounded in a physical-size knowledge base that stores characteristic object lengths and, when available, plausible length ranges. 
For common-object scale reasoning, we start from visually identifiable categories in COCO~\citep{Lin14COCO} and LVIS~\citep{Gupta19lvis}, and remove categories whose physical extent is highly variable, visually ambiguous, or non-rigid. 
For product-scale reasoning, we use product-level dimensions and reference images from Amazon Berkeley Objects (ABO)~\citep{collins2022abo}. 
Human-anchor dimensions are grounded in ANSUR II anthropometric measurements~\citep{gordon2014ansur}. 
Additional object dimensions are derived from domain-specific references, including Animal Diversity Web~\citep{myers2026adw}, USDA FoodData Central~\citep{usda2026fdc}, ISO and ANSI standards~\citep{iso7810_2019,iso216_2007,ansi2021c181m}, traffic specifications~\citep{fhwa2023mutcd}, and official sport-object specifications~\citep{mlb2026officialrules}. 
The goal is not maximal category coverage, but stable and externally grounded physical extent for reliable relative-scale evaluation.

\cref{tab:kb_coverage} summarizes the semantic and source coverage of the resulting knowledge base. 
The 100 entries include 97 non-human common objects used for Task~1 and common-object correction cases in Task~3, together with three human body-part anchors used for Task~2.

\begin{table}[t]
\centering
\small
\caption{\textbf{Coverage of the 100-entry physical scale knowledge base.}
Objects are grouped by semantic domain and dimensional source type. 
The knowledge base contains 42 COCO-derived and 58 LVIS-derived categories, covering both everyday objects and physically standardized objects.}
\label{tab:kb_coverage}
\renewcommand{\arraystretch}{1.22}
\setlength{\extrarowheight}{1.5pt}
\setlength{\tabcolsep}{5pt}
\begin{tabularx}{\linewidth}{
@{}
>{\RaggedRight\arraybackslash}p{0.25\linewidth}
>{\centering\arraybackslash}p{0.055\linewidth}
>{\centering\arraybackslash}p{0.055\linewidth}
>{\RaggedRight\arraybackslash}p{0.29\linewidth}
>{\RaggedRight\arraybackslash}X
@{}}
\toprule
Group & \#Obj. & \% & Example categories & Dimensional grounding \\
\midrule

Human body-part anchors
& 3 & 3.0
& hand, face, foot
& ANSUR II anthropometric measurements~\citep{gordon2014ansur}. \\

Animals
& 13 & 13.0
& elephant, giraffe, horse, cat, turtle
& Animal Diversity Web and species-level references~\citep{myers2026adw}. \\

Sports / recreation
& 14 & 14.0
& baseball, basketball, tennis racket, surfboard
& Official sport rules and standardized equipment dimensions. \\

Electronics / appliances / media
& 10 & 10.0
& keyboard, laptop, TV, microwave, CD
& ANSI, ISO, and typical product specifications~\citep{iso216_2007,ansi2021c181m}. \\

Office / personal small objects
& 16 & 16.0
& battery, credit card, passport, pencil, toothbrush
& ISO, ANSI, and common product standards~\citep{iso7810_2019,iso216_2007,ansi2021c181m}. \\

Tableware / containers
& 11 & 11.0
& bottle, wine glass, mug, fork, bowl, can
& Standard tableware and container dimensions. \\

Food items
& 7 & 7.0
& apple, orange, banana, carrot, egg
& USDA FoodData Central and common food-size standards~\citep{usda2026fdc}. \\

Vehicles / traffic / public infrastructure
& 12 & 12.0
& bicycle, car, bus, stop sign, fire hydrant
& Vehicle, traffic, and infrastructure specifications~\citep{fhwa2023mutcd}. \\

Household / tools / accessories / instruments
& 14 & 14.0
& umbrella, suitcase, hammer, watch, guitar
& Common product and instrument dimensions. \\

\midrule
Total & 100 & 100.0 & -- & -- \\
\bottomrule
\end{tabularx}
\end{table}

\subsection{Task 1: Common-Object Sampling and Prompt Templates}
\label{app:task1_details}

Task~1 evaluates implicit common-object scale priors in text-to-image generation. 
Each entry contains only a text prompt with two to four non-human objects; no object-specific metric size or reference image is provided to the generator. 
Objects are sampled from the 97 non-human common-object entries in the physical-size knowledge base. 
We retain sampled sets whose largest-to-smallest characteristic-length ratio lies in $[4,20]$, which removes near-equal comparisons that are difficult to judge and extremely disparate combinations that often become layout-dominated rather than scale-diagnostic.

\cref{tab:task1_sampling} summarizes the sampling protocol, and \cref{tab:t1_object_count} reports the final object-count distribution. 
For an $n$-object prompt, we evaluate all $\binom{n}{2}$ unordered object pairs, so multi-object prompts increase pair-level evaluation density without increasing the number of generated images. 
Task~1 contains 400 prompts and 1{,}143 evaluated relations, split nearly evenly between S1 Natural Depth and S2 Same Plane.

To extend the analysis from~\cref{fig:s1_depth_size}, we estimate object depth for every Task~1 generated image using Depth Anything V2.
Because bounding boxes are not segmentation masks, we avoid full-box averaging: for each object, we use the central 60\% of its bounding box, discard the top and bottom 5\% depth values, and take the median of the remaining values as the object depth.
We then convert object depths into within-image depth ranks, where 0 denotes the nearest object and 1 denotes the farthest.
Objects are grouped by their real-world size rank within each image, and we compute both the average depth rank of each size group and the Spearman correlation between the benchmark physical length $\log \ell$ and the estimated depth rank.

The full diagnostic in~\cref{tab:appendix_task1_depth_size_full} confirms that depth placement mainly affects S1.
In S1, all seven generators place smaller objects closer and larger objects farther away: on average, the smallest objects have depth rank 0.176, while the largest objects have depth rank 0.863.
Equivalently, 79.2\% of unequal-size object pairs place the smaller object closer to the camera, and the mean size--depth Spearman correlation is 0.558.
In contrast, S2 largely removes this pattern: the average depth ranks of small, middle, and large objects become much closer (0.423/0.480/0.592), the smaller-closer rate drops to 52.2\%, and the correlation falls to 0.129.
This supports the interpretation that natural depth in S1 can partially hide relative-scale errors through depth-mediated scale compression, whereas S2 exposes scale errors more directly by constraining objects to a similar depth plane.

\begin{table}[t]
\centering
\small
\caption{\textbf{Task 1 object-count distribution.}
Each $n$-object prompt induces all $\binom{n}{2}$ unordered pairwise scale relations. 
Thus, three- and four-object prompts substantially increase the number of evaluated relations without increasing the number of generated images.}
\label{tab:t1_object_count}
\renewcommand{\arraystretch}{1.15}
\setlength{\tabcolsep}{7pt}
\begin{tabularx}{\linewidth}{
@{}
>{\RaggedRight\arraybackslash}p{0.21\linewidth}
>{\centering\arraybackslash}X
>{\centering\arraybackslash}X
>{\centering\arraybackslash}X
>{\centering\arraybackslash}X
@{}}
\toprule
Scenario & 2 objects & 3 objects & 4 objects & Total \\
\midrule
S1 Natural Depth 
& \makecell[c]{93 images \\ 93 pairs}
& \makecell[c]{55 images \\ 165 pairs}
& \makecell[c]{52 images \\ 312 pairs}
& \makecell[c]{200 images \\ 570 pairs} \\

\midrule
S2 Same Plane 
& \makecell[c]{81 images \\ 81 pairs}
& \makecell[c]{74 images \\ 222 pairs}
& \makecell[c]{45 images \\ 270 pairs}
& \makecell[c]{200 images \\ 573 pairs} \\
\midrule
Task 1 total 
& \makecell[c]{174 images \\ 174 pairs}
& \makecell[c]{129 images \\ 387 pairs}
& \makecell[c]{97 images \\ 582 pairs}
& \makecell[c]{400 images \\ 1,143 pairs} \\
\bottomrule
\end{tabularx}
\end{table}

\begin{table}[t]
\centering
\small
\caption{\textbf{Task 1 sampling protocol.}
Task~1 samples 2--4 common objects and filters combinations by physical disparity.}
\label{tab:task1_sampling}
\renewcommand{\arraystretch}{1.12}
\setlength{\tabcolsep}{6pt}
\begin{tabularx}{\linewidth}{
@{}
>{\RaggedRight\arraybackslash}p{0.20\linewidth}
>{\RaggedRight\arraybackslash}X
@{}}
\toprule
Choice & Implementation \\
\midrule
Object pool 
& 97 non-human common objects from the physical-size knowledge base; human body-part anchors are excluded. \\

Prompt size 
& Each prompt contains 2--4 objects. The final split contains 174 two-object, 129 three-object, and 97 four-object prompts. \\

Disparity filter 
& With $D=\max_i l_i / \min_i l_i$, we keep only combinations satisfying $4 \leq D \leq 20$. \\

Rationale 
& Smaller disparities are often hard to judge reliably, whereas larger disparities tend to produce layout-dominated scenes. \\

Scenario split 
& S1 allows natural depth and perspective; S2 places objects on a similar depth plane to reduce perspective shortcuts. \\

Metric cues 
& No numeric size is shown to the model; the task tests implicit common-object scale priors. \\
\bottomrule
\end{tabularx}
\end{table}

\begin{table}[t]
\centering
\small
\caption{\textbf{Task 1 prompt templates.}
S1 permits realistic perspective and depth variation, while S2 constrains the objects to a similar depth plane.}
\label{tab:t1_prompt_templates}
\resizebox{\linewidth}{!}{
\begin{tabular}{l p{0.74\linewidth}}
\toprule
Scenario & Prompt template \\
\midrule
S1: Natural Depth 
& Create a photorealistic scene containing \texttt{[object list]}. The objects should appear naturally in the scene with physically plausible real-world relative sizes. \\
S2: Same Plane 
& Create a photorealistic scene containing \texttt{[object list]}. Place all objects on approximately the same depth plane, such as on the same tabletop or floor surface, with physically plausible real-world relative sizes. \\
\bottomrule
\end{tabular}}
\end{table}

\begin{table*}[t]
\centering
\small
\caption{\textbf{Full Task 1 depth-size diagnostic.}
Objects are grouped by real-world size within each generated image. 
Depth rank is averaged within each size group, where 0 denotes the nearest object and 1 denotes the farthest.}
\label{tab:appendix_task1_depth_size_full}
\vspace{0.6ex}
\renewcommand\arraystretch{0.96}
\setlength{\tabcolsep}{4.0pt}
\resizebox{0.96\textwidth}{!}{%
\begin{tabular}{@{}llrrrrrrrr@{}}
\toprule
Model 
& Split 
& Img. 
& Obj. 
& Pairs 
& \multicolumn{3}{c}{Mean depth rank by size bucket} 
& Smaller closer (\%) 
& Size--depth $\rho$ \\
\cmidrule(lr){6-8}
& & & & 
& Small 
& Middle 
& Large 
& & \\
\midrule
\multirow{3}{*}{Nano Banana 2}
& S1  & 199 & 555  & 562  & 0.136 & 0.452 & 0.903 & 80.6 & 0.620 \\
& S2  & 196 & 551  & 556  & 0.545 & 0.493 & 0.460 & 41.5 & -0.085 \\
& All & 395 & 1106 & 1118 & 0.339 & 0.473 & 0.683 & 61.2 & 0.269 \\
\midrule
\multirow{3}{*}{GPT-Image-2}
& S1  & 199 & 555  & 562  & 0.148 & 0.457 & 0.888 & 81.5 & 0.595 \\
& S2  & 200 & 564  & 572  & 0.501 & 0.503 & 0.497 & 44.1 & -0.006 \\
& All & 399 & 1119 & 1134 & 0.325 & 0.480 & 0.692 & 62.6 & 0.294 \\
\midrule
\multirow{3}{*}{Qwen-Image 2512}
& S1  & 196 & 542  & 540  & 0.164 & 0.450 & 0.876 & 78.1 & 0.582 \\
& S2  & 198 & 557  & 562  & 0.342 & 0.504 & 0.656 & 58.0 & 0.246 \\
& All & 394 & 1099 & 1102 & 0.253 & 0.478 & 0.765 & 67.9 & 0.412 \\
\midrule
\multirow{3}{*}{FLUX.2}
& S1  & 198 & 537  & 519  & 0.145 & 0.476 & 0.872 & 82.3 & 0.591 \\
& S2  & 199 & 551  & 543  & 0.352 & 0.462 & 0.678 & 58.7 & 0.258 \\
& All & 397 & 1088 & 1062 & 0.249 & 0.469 & 0.775 & 70.2 & 0.422 \\
\midrule
\multirow{3}{*}{SD3.5-Large}
& S1  & 179 & 495 & 495 & 0.233 & 0.419 & 0.830 & 74.1 & 0.502 \\
& S2  & 166 & 455 & 442 & 0.326 & 0.477 & 0.692 & 61.3 & 0.290 \\
& All & 345 & 950 & 937 & 0.278 & 0.447 & 0.763 & 68.1 & 0.398 \\
\midrule
\multirow{3}{*}{Grok Imagine}
& S1  & 198 & 553  & 562  & 0.183 & 0.452 & 0.855 & 79.7 & 0.536 \\
& S2  & 200 & 564  & 572  & 0.494 & 0.473 & 0.528 & 47.6 & 0.021 \\
& All & 398 & 1117 & 1134 & 0.340 & 0.463 & 0.691 & 63.5 & 0.274 \\
\midrule
\multirow{3}{*}{Z-Image-Turbo}
& S1  & 193 & 531  & 526 & 0.224 & 0.449 & 0.816 & 77.8 & 0.483 \\
& S2  & 181 & 491  & 468 & 0.404 & 0.445 & 0.635 & 54.1 & 0.178 \\
& All & 374 & 1022 & 994 & 0.311 & 0.447 & 0.728 & 66.6 & 0.333 \\
\midrule
\multirow{3}{*}{Average}
& S1  & 1362 & 3768 & 3766 & 0.176 & 0.451 & 0.863 & 79.2 & 0.558 \\
& S2  & 1340 & 3733 & 3715 & 0.423 & 0.480 & 0.592 & 52.2 & 0.129 \\
& All & 2702 & 7501 & 7481 & 0.299 & 0.465 & 0.728 & 65.7 & 0.343 \\
\bottomrule
\end{tabular}%
}
\end{table*}

\subsection{Task 2: ABO Product Filtering and Human-Anchor Assignment}
\label{app:task2_details}

Task~2 evaluates explicit metric scale realization in human-product image generation. 
Each entry contains one ABO product~\citep{collins2022abo}, one product reference image retrievable from ABO metadata, a text prompt specifying the product dimensions, and one size-conditioned human anchor. 
We use ABO because it provides product-level dimensions and product images, enabling evaluation of whether a model can translate metric product size into plausible human-product proportions.

We filter ABO products using four criteria. 
First, we retain products with an English listing title. 
Second, we require valid item dimensions. 
Third, we require a valid main image identifier that can be mapped to an available ABO image. 
Fourth, after converting all available dimensions to centimeters, we use the maximum of length, width, and height as the characteristic product length and retain products whose characteristic length lies in $[5,250]$ cm. 
This range removes tiny objects that are difficult to resolve visually and very large products that are unlikely to form a well-controlled human-anchor interaction. 
For each retained product, we store length, width, height, characteristic length, and a narrow product-specific tolerance interval.

Human anchors are assigned deterministically from the product characteristic length. 
Products shorter than 30 cm are paired with a human hand; products between 30 and 60 cm are paired with a human face/head; products between 60 and 100 cm are paired with a human foot/leg; products longer than 100 cm are paired with a full human body. 
The corresponding canonical anchor lengths are derived from anthropometric references~\citep{gordon2014ansur}. 
\cref{tab:task2_anchor_policy} gives the anchor-assignment policy, and~\cref{tab:t2_anchor_distribution} reports the final anchor distribution in the 300-entry Task~2 split.

The Task~2 prompt contains both semantic and metric constraints. 
The product title is cleaned into a concise noun phrase, while the metric block provides the characteristic length and all available axis-aligned dimensions. 
The prompt asks the model to depict a natural, purpose-consistent interaction between the product and the assigned human anchor, with catalog-style visibility and mild perspective so that product-to-anchor scale remains interpretable.

\begin{table}[t]
\centering
\small
\caption{\textbf{Human anchor selection in Task 2.}
Products are paired with a human anchor according to their characteristic length, enabling interpretable product-to-human scale evaluation across small, medium, large, and oversized products.}
\label{tab:task2_anchor_policy}
\resizebox{0.7\linewidth}{!}{
\begin{tabular}{c c c c}
\toprule
Product characteristic length & Human anchor & Anchor length & \#Entries \\
\midrule
$<30$ cm & human hand & 19.3 cm & 151 \\
$30$--$60$ cm & human head/face & 24.0 cm & 77 \\
$60$--$100$ cm & human foot/leg & 26.5 cm & 31 \\
$\geq100$ cm & full human body & 170.0 cm & 41 \\
\bottomrule
\end{tabular}}
\end{table}

\begin{table}[t]
\centering
\small
\caption{\textbf{Human-anchor distribution in Task 2.}
Each Task~2 prompt contains one product and one human anchor, resulting in one evaluated product-to-anchor scale relation per image.
Product length is computed from \texttt{product\_scale.typical\_len\_cm}. 
Ratio denotes the mean ground-truth target-to-anchor length ratio.}
\label{tab:t2_anchor_distribution}
\renewcommand{\arraystretch}{1.12}
\setlength{\tabcolsep}{6pt}
\begin{tabularx}{\linewidth}{
@{}
>{\RaggedRight\arraybackslash}p{0.22\linewidth}
>{\centering\arraybackslash}p{0.08\linewidth}
>{\centering\arraybackslash}p{0.06\linewidth}
>{\centering\arraybackslash}p{0.18\linewidth}
>{\centering\arraybackslash}p{0.22\linewidth}
>{\centering\arraybackslash}X
@{}}
\toprule
Human anchor & Entries & \% & Median length & Length range & Ratio \\
\midrule
Human hand        & 151 & 50.3 & 18.29 cm  & 5.08--30.23 cm    & 0.94 \\
Human head / face & 77  & 25.7 & 39.62 cm  & 30.00--58.42 cm   & 1.72 \\
Human foot / leg  & 31  & 10.3 & 70.61 cm  & 31.75--96.52 cm   & 2.81 \\
Full human body   & 41  & 13.7 & 150.00 cm & 100.00--243.84 cm & 0.95 \\
\midrule
Total             & 300 & 100.0 & -- & -- & -- \\
\bottomrule
\end{tabularx}
\end{table}

\begin{table}[t]
\centering
\small
\caption{\textbf{Task 2 prompt-construction constraints.}
Task~2 combines an ABO product reference, explicit metric product dimensions, and a size-conditioned human anchor.}
\label{tab:t2_prompt_template}
\renewcommand{\arraystretch}{1.12}
\setlength{\tabcolsep}{5pt}
\begin{tabularx}{\linewidth}{
@{}
>{\RaggedRight\arraybackslash}p{0.22\linewidth}
>{\RaggedRight\arraybackslash}X
@{}}
\toprule
Component & Construction rule \\
\midrule
Product scale block
& Provide the characteristic product length and all available length/width/height dimensions in centimeters. The prompt emphasizes that the generated product must match this order of magnitude rather than infer size from the title alone. \\

Human anchor
& Select hand, face/head, foot/leg, or full body according to the product characteristic length. The anchor is fixed for metric consistency. \\

Interaction constraint
& Ask for a natural, purpose-consistent interaction between the product and the human anchor, avoiding arbitrary or decorative placements that ignore product use. \\

Visibility constraint
& Require catalog-style framing where the product remains recognizable and largely unobstructed; mild contact or partial occlusion by the human anchor is allowed. \\

Camera constraint
& Favor front or three-quarter mild-perspective views rather than extreme wide-angle compositions, so apparent product-to-anchor ratios remain interpretable. \\
\bottomrule
\end{tabularx}
\end{table}

\subsection{Task 3: Failure-Case Selection and Correction Prompt Construction}
\label{app:task3_details}

Task~3 evaluates scale-error correction rather than average-case image editing. 
We construct Task~3 from failed but editable outputs in Tasks~1 and~2. 
A source case is retained only when the main objects are recognizable, the scene is visually usable, and at least one pairwise scale relation is not perfectly plausible under the GenScale ordinal rubric. 
For Task~1-derived cases, we restrict source images to two-object prompts so that the correction target is unambiguous. 
For both Task~1 and Task~2 sources, we remove images with missing or unidentifiable objects, duplicate primary objects, severe synthesis artifacts, heavy occlusion, blur, or extreme correction factors that make localized editing ill-defined. 
This filtering isolates correction of object scale rather than recovery from invalid image generation.

The final Task~3 set contains 100 source images, with 61 from Task~1 and 39 from Task~2. 
Each source image is converted into two edit settings. 
In S4 Auto-Discovery, the editor receives the source image and scale references and decides whether and how to resize, while preserving identity, composition, background, and lighting.
For Task~1 sources, the editor also selects the editable object. For Task~2 sources, the prompt includes product and human-anchor reference lengths and limits edits to the product.
In S5 Precise Scale Instruction, the prompt directly specifies the fixed reference object, editable target object, resize direction, numeric scale factor, and target size ratio. 
This separates fine-grained resize-instruction following from autonomous error diagnosis.

\cref{tab:task3_construction_protocol} summarizes the construction protocol, and~\cref{tab:t3_prompt_templates} summarizes the S4/S5 edit-instruction format.

\cref{tab:task3_qc_stats} summarizes the visual quality check (\textbf{QC}) applied when constructing Task~3 from failed two-object examples in Task~1 and Task~2. Most candidates passed QC and were retained, yielding 61 Task~1 sources and 39 Task~2 sources. The rejected cases were mainly due to missing or visually unclear target objects, while duplicate-object ambiguity and severe synthesis failures were less frequent.

\begin{table}[t]
\centering
\small
\caption{\textbf{Task 3 construction protocol.}
Task 3 is built from failed generations in Task 1 and Task 2, and each selected source image is converted into two correction prompts.}
\label{tab:task3_construction_protocol}
\resizebox{\linewidth}{!}{
\begin{tabular}{l p{0.66\linewidth}}
\toprule
Component & Description \\
\midrule
Source images & Failed or imperfect generations from Task 1 and Task 2. In the final benchmark, Task 3 uses 100 source images: 61 from T1 and 39 from T2. \\
Source models & T1 source images are generated by FLUX.2; T2 source images are generated by Qwen-Image. \\
Quality control & We remove images with severe synthesis failures, missing/unidentifiable target objects, duplicate main objects, or heavy occlusion/blur that would make scale correction ill-defined. Selection is not based on downstream editing success. \\
S4: Hard auto-discovery & The model diagnoses and plans the correction without a prescribed direction or factor. For human--product cases, the product is editable and the human anchor is fixed. \\
S5: Precise scale instruction & The model is directly given the object to resize and the exact scale factor, isolating scale-instruction following from error diagnosis. \\
Expansion & Each source image yields two prompts, one S4 and one S5, producing 200 Task 3 entries. \\
\bottomrule
\end{tabular}}
\end{table}

\begin{table}[t]
\centering
\small
\caption{\textbf{Task 3 correction-prompt construction.}
Each retained erroneous source image is converted into both an auto-discovery correction prompt and a precise scale-instruction prompt.}
\label{tab:t3_prompt_templates}
\renewcommand{\arraystretch}{1.12}
\setlength{\tabcolsep}{5pt}
\begin{tabularx}{\linewidth}{
@{}
>{\RaggedRight\arraybackslash}p{0.20\linewidth}
>{\RaggedRight\arraybackslash}X
@{}}
\toprule
Scenario & Construction rule \\
\midrule
S4: Auto-Discovery
& Ask the editor to inspect the image for unrealistic object-size proportions. If a scale error exists, the editor must automatically identify the necessary object-size correction and preserve object identity, pose, viewpoint, composition, background, and lighting. \\

S4 from Task~2
& Use the same auto-discovery instruction, but also provide compact product and human-anchor reference lengths, because Task~2 errors depend on explicit product-human metric scale. \\

S5: Precise Scale Instruction
& Provide a strict edit instruction specifying the object to keep unchanged, the object to edit, the resize direction, the exact multiplicative scale factor, and the target pairwise ratio. The prompt forbids identity, position, background, lighting, or viewpoint changes. \\
\bottomrule
\end{tabularx}
\end{table}

\begin{table}[t]
\centering
\small
\caption{\textbf{Task 3 visual quality control statistics.}
We report the number of source candidates before and after QC, together with rejection reasons.}
\label{tab:task3_qc_stats}
\begin{tabular}{l c c c c c}
\toprule
Source task & Candidates & Kept & Missing object & Duplicate object & Severe synthesis failure \\
\midrule
T1 & 75 & 61 & 8 & 4 & 2 \\
T2 & 42 & 39 & 2 & 0 & 1 \\
\bottomrule
\end{tabular}
\end{table}

\subsection{Pairwise Ratio Definition and Metadata Schema}
\label{app:metadata_schema}

Each GenScale entry is an image-level prompt or editing instance, but evaluation is performed at the pair level. 
For an anchor-target pair $(a,t)$, let $l_a$ and $l_t$ denote their characteristic physical lengths. 
We define the expected target-to-anchor physical ratio as
\[
    r_{t/a} = \frac{l_t}{l_a}.
\]
When lower and upper physical ranges are available, we also store a tolerance interval
\[
    \left[
    \frac{l_t^{\min}}{l_a^{\max}},
    \frac{l_t^{\max}}{l_a^{\min}}
    \right],
\]
which accounts for intra-class variation and measurement uncertainty. 
For Task~2, product dimensions are taken from ABO metadata~\citep{collins2022abo}, while human-anchor lengths are fixed canonical measurements derived from anthropometric references~\citep{gordon2014ansur}. 
For Task~3, the pairwise ratio is inherited from the corresponding Task~1 or Task~2 source case.

\cref{tab:prompt_gt_construction} summarizes how prompts, scale information, and ground-truth ratios differ across the three tasks.

\begin{table}[t]
\centering
\small
\caption{\textbf{Prompt and ground-truth construction.}
\benchmarkname{} separates implicit common-object scale priors from explicit metric scale control and scale-error correction.}
\label{tab:prompt_gt_construction}
\renewcommand{\arraystretch}{1.13}
\setlength{\tabcolsep}{6pt}
\begin{tabularx}{\linewidth}{
@{}
>{\RaggedRight\arraybackslash}p{0.22\linewidth}
>{\RaggedRight\arraybackslash}X
@{}}
\toprule
Aspect & Description \\
\midrule

\multicolumn{2}{@{}l}{\textbf{T1: Common objects}} \\
Input
& Text-only generation with two to four common objects. \\
Scale prompt
& No object-specific numeric size is provided; prompts only request physically accurate real-world proportions. \\
Ground truth
& For objects $i,j$, the expected ratio is $r_{ij}=l_i/l_j$, with acceptable interval
$[l_i^{\min}/l_j^{\max},\, l_i^{\max}/l_j^{\min}]$. \\
Evaluation target
& Tests whether the model's implicit real-world object-size prior is accurate. \\

\midrule

\multicolumn{2}{@{}l}{\textbf{T2: Human--product}} \\
Input
& Product reference image plus text prompt. \\
Scale prompt
& Product dimensions from ABO metadata are explicitly included; the human anchor has a fixed canonical size. \\
Ground truth
& The product-to-anchor ratio is computed from the ABO characteristic product length and the assigned human-anchor length, with tolerance intervals. \\
Evaluation target
& Tests whether the model can realize explicit metric product scale while preserving product identity and natural human interaction. \\

\midrule

\multicolumn{2}{@{}l}{\textbf{T3: Scale correction}} \\
Input
& Erroneous generated image plus edit instruction. \\
Scale prompt
& S4 asks the model to diagnose whether and how to resize; S5 specifies the target object, reference object, resize direction, and scale factor. \\
Ground truth
& The pairwise ratio is inherited from the corresponding T1 or T2 source case. \\
Evaluation target
& Tests whether an editing model can diagnose and correct scale errors without changing identity, pose, or scene composition. \\

\bottomrule
\end{tabularx}
\end{table}

\begin{table}[t]
\centering
\small
\caption{\textbf{Benchmark metadata schema.}
Each image-level entry contains task-level information and one or more pairwise scale records used for evaluation.}
\label{tab:metadata_schema}
\resizebox{\linewidth}{!}{
\begin{tabular}{l p{0.70\linewidth}}
\toprule
Field & Description \\
\midrule
\texttt{entry\_id} & Unique image-level benchmark identifier. \\
\texttt{task} / \texttt{scenario} & Task label and scenario label, e.g., T1/S1, T1/S2, T2/S3, T3/S4, or T3/S5. \\
\texttt{prompt} & Generation or editing prompt shown to the model. \\
\texttt{objects} & Object names appearing in the prompt or source image. \\
\texttt{product\_image} & Product reference image path for Task~2 entries; absent for Task~1. \\
\texttt{source\_image} & Erroneous source image path for Task~3 entries. \\
\texttt{physical\_lengths} & Characteristic physical lengths and optional lower/upper ranges for each object or anchor. \\
\texttt{pairs} & List of anchor-target records used as the atomic evaluation units. \\
\texttt{pairs.anchor} / \texttt{pairs.target} & Reference object and evaluated target object. \\
\texttt{pairs.expected\_ratio} & Expected target-to-anchor physical ratio $r_{t/a}$. \\
\texttt{pairs.ratio\_interval} & Optional plausible physical-ratio interval induced by object length ranges. \\
\texttt{edit\_metadata} & For Task~3, editable object, fixed reference object, resize direction, and scale factor when available. \\
\bottomrule
\end{tabular}}
\end{table}

\subsection{Benchmark Release Format}
\label{app:release_format}

The public release of \benchmarkname{} includes \texttt{GenScale\_Benchmark.json}, with 900 entries across Tasks~1--3, and CSV and Parquet tables for benchmark entries, pairwise scale relations, and editing plans. It also includes the physical-size knowledge base and the sampled ABO candidate metadata.
The image files comprise 294 Task~2 product references, 39 Task~3 product references, and 100 generated Task~3 inputs. The product references are selected from Amazon Berkeley Objects (ABO)~\citep{collins2022abo}; the full ABO image pool is not bundled.

\begin{table}[t]
\centering
\small
\caption{\textbf{Released benchmark artifacts.}
Paths to packaged images are relative to the dataset root.}
\label{tab:release_artifacts}
\renewcommand{\arraystretch}{1.12}
\setlength{\tabcolsep}{5pt}
\begin{tabularx}{\linewidth}{@{}l l X@{}}
\toprule
Artifact & Format & Content \\
\midrule
Benchmark records & JSON, CSV, Parquet & 900 benchmark entries, pairwise scale relations, and editing plans. \\
Supporting metadata & CSV & Physical-size knowledge base and sampled ABO candidate metadata. \\
Product references & Images & 294 Task~2 and 39 Task~3 product-reference files selected from ABO. \\
Task~3 inputs & Images & 100 generated source images shared by S4 and S5. \\
Evaluation code & Python & Task-specific scripts for scene filtering, judge sampling, and score aggregation. \\
Documentation & Markdown, JSON & Dataset and code documentation, license information, and Croissant metadata. \\
\bottomrule
\end{tabularx}
\end{table}

The JSON schema is task-dependent but follows a common structure.
Task~1 entries contain a text prompt, a scenario label, the included common objects, and their pairwise scale ratios.
Task~2 entries include product dimensions, the human anchor, and the path to the packaged product reference image.
Task~3 entries include the source-image path, correction prompt, source-task provenance, and the pairwise ratio inherited from the source case.
For S5, each entry also specifies the editable object, fixed reference object, resize direction, and scale factor.

\begin{table}[t]
\centering
\small
\caption{\textbf{Core fields in the released benchmark JSON.}
All paths are relative to the release root. 
Task-specific fields are present only when applicable.}
\label{tab:release_schema}
\renewcommand{\arraystretch}{1.12}
\setlength{\tabcolsep}{5pt}
\begin{tabularx}{\linewidth}{@{}l l X@{}}
\toprule
Field & Tasks & Description \\
\midrule
\texttt{task\_id}
& T1--T3
& Stable entry identifier, e.g., \texttt{T1\_0000}, \texttt{T2\_0000}, or \texttt{T3\_0000}. \\

\texttt{scenario}
& T1--T3
& Scenario label: S1/S2 for Task~1, S3 for Task~2, and S4/S5 for Task~3. \\

\texttt{prompt}
& T1--T3
& Text prompt given to the generator or editor. \\

\texttt{objects\_included}
& T1--T3
& Canonical object names expected in the image. \\

\texttt{gt\_ratios}
& T1--T3
& Dictionary of anchor--target pair records. Each record stores the expected length ratio and an acceptable physical-size range. \\

\texttt{reference\_image\_path}
& T2
& Relative path to the product reference image used as the image condition. \\

\texttt{product\_scale}
& T2
& Product-level physical dimensions, including typical, minimum, and maximum characteristic lengths in centimeters. \\

\texttt{raw\_listing\_title}
& T2
& Original or lightly normalized product title used to identify the product category and dimensions. \\

\texttt{source\_task\_id}
& T3
& Identifier of the failed Task~1 or Task~2 entry from which the correction case is derived. \\

\texttt{source\_task\_type}
& T3
& Source task type, either \texttt{T1} or \texttt{T2}; used to dispatch Task~3 outputs to the corresponding evaluator. \\

\texttt{image\_path}
& T3
& Relative path to the erroneous source image used as the editing input. \\

\texttt{prompt\_type}
& T3
& Correction setting: \texttt{hard\_auto\_discovery} for S4 or \texttt{precise\_scale\_instruction} for S5. \\

\texttt{edit\_plan}
& T3-S5
& Explicit correction metadata containing target object, reference object, resize direction, and scale factor. \\
\bottomrule
\end{tabularx}
\end{table}

To evaluate a new model, users generate images using the released prompts and reference images, save outputs under the expected task identifiers, and run the provided task-specific evaluation scripts.
The evaluator reads the corresponding pairwise records from \texttt{GenScale\_Benchmark.json}, applies the scene-level validity prefilter, queries the calibrated VLM judge for each scorable pair, and aggregates pairwise ordinal scores by task, scenario, and model.
Task~3 evaluation reuses the Task~1 or Task~2 scoring logic according to each entry's \texttt{source\_task\_type}, ensuring that edited images are judged with the same criterion as their original source case.

\section{Human and VLM Evaluation Details}
\label{app:human_vlm_eval}

\subsection{Human Annotation Interface and Instructions}
\label{app:human_interface}

\begin{figure}[t]
    \centering
    \includegraphics[width=0.98\linewidth]{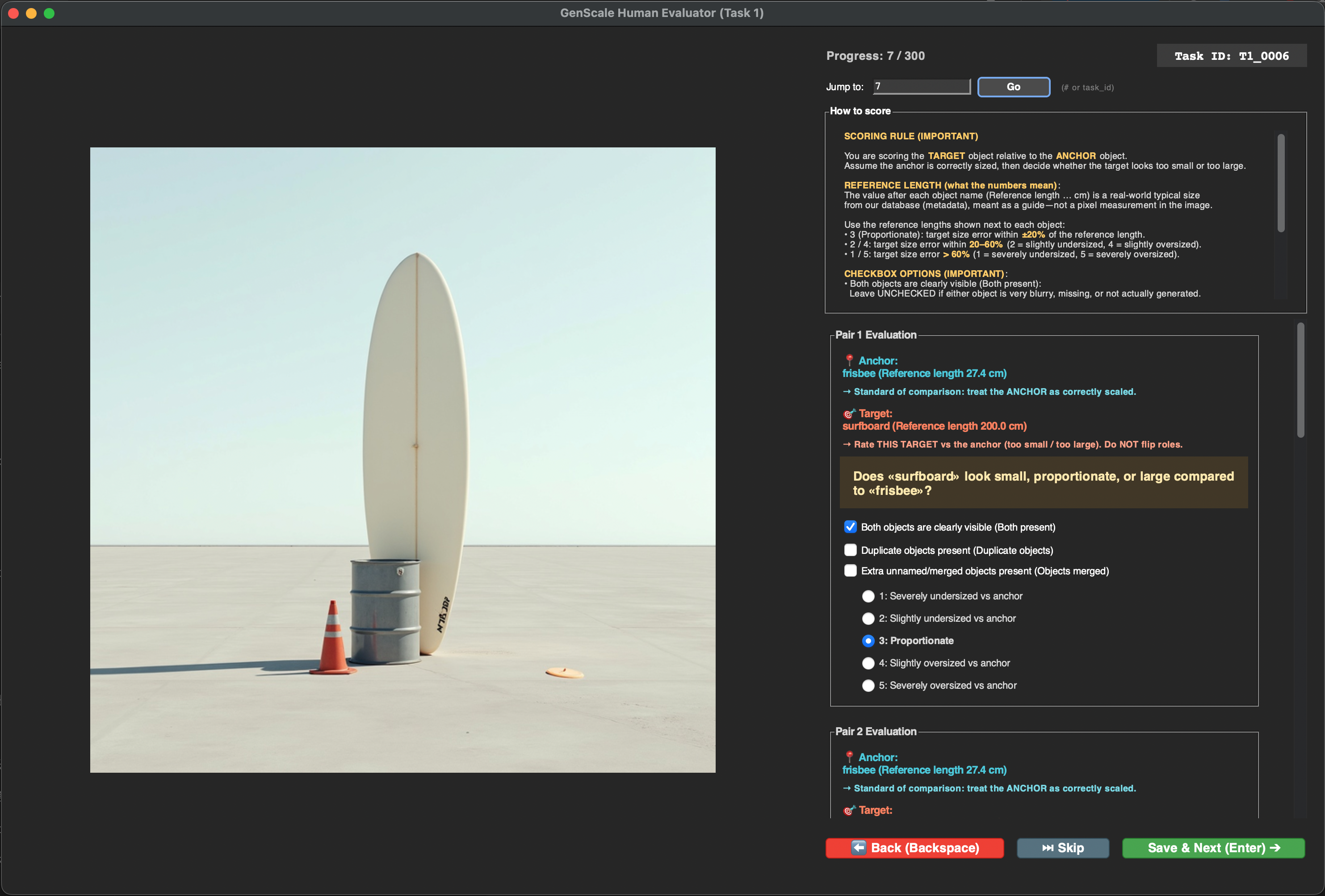}
    \caption{\textbf{Human annotation interface.}
    The interface shows one generated image together with one or more anchor--target pair cards.
    For each pair, annotators are shown the anchor and target names, their reference physical lengths, pair-specific validity checkboxes, and a five-point ordinal scale for judging whether the target is too small, proportionate, or too large relative to the anchor.}
    \label{fig:human_annotation_interface}
\end{figure}

We built a custom annotation GUI for pairwise relative-scale evaluation, shown in~\cref{fig:human_annotation_interface}. 
Each screen contains a generated image on the left and pairwise evaluation cards on the right. 
A single image may induce multiple anchor--target pairs, especially in Task~1 multi-object prompts; annotators score each pair independently while looking at the same image.

For each pair, the interface displays the \emph{anchor} object, the \emph{target} object, and their reference physical lengths in centimeters. 
Annotators are instructed to treat the anchor as correctly sized and judge only whether the target appears too small, proportionate, or too large relative to that anchor. 
This role assignment is fixed by the benchmark metadata and must not be reversed. 
The reference lengths shown in the interface are real-world typical lengths from the physical-size metadata or product catalog; they are intended as semantic scale references rather than pixel measurements in the rendered image.

The interface also records pair-level validity information. 
Annotators mark whether both objects are clearly visible, whether duplicate instances make the pair ambiguous, and whether merged objects or severe artifacts prevent reliable scale judgment. 
Navigation controls allow annotators to save the current image-level ratings, move backward, skip unusable examples, or jump to a specific index or task identifier. 
The calibration interface supports Task~1 scenarios S1 and S2 and Task~2 scenario S3, with configurable scenario ranges and per-scenario quotas.

\subsection{Pairwise Ordinal Rubric and Invalid-Pair Criteria}
\label{app:ordinal_rubric}

GenScale uses a five-point ordinal rubric rather than a raw pixel-ratio metric because apparent image size depends on object identity, real-world extent, camera perspective, depth ordering, foreshortening, occlusion, and partial visibility. 
For each valid anchor--target pair, annotators assign a score $s\in\{1,2,3,4,5\}$:
\begin{itemize}
    \item $s=1$: the target is severely undersized relative to the anchor.
    \item $s=2$: the target is slightly undersized relative to the anchor.
    \item $s=3$: the target has a physically plausible size relative to the anchor.
    \item $s=4$: the target is slightly oversized relative to the anchor.
    \item $s=5$: the target is severely oversized relative to the anchor.
\end{itemize}
Operationally, score 3 corresponds to an inferred target-size error within approximately $\pm 20\%$ of the reference scale, scores 2 and 4 correspond to errors between $20\%$ and $60\%$, and scores 1 and 5 correspond to errors larger than $60\%$. 
These thresholds are used as perceptual guidance rather than exact pixel-caliper rules.

Annotators are explicitly instructed to account for perspective and depth before assigning a score. 
For example, a target farther from the camera may appear smaller in image space without being physically undersized, and a foreground target may appear larger without being physically oversized. 
A penalty is assigned only when the target still appears implausibly small or large after considering the likely 3D layout, object contact, foreshortening, and partial visibility. 
In S2 same-plane cases, the default assumption is that objects lie at approximately similar depth, so visible size differences provide stronger evidence of real-world scale errors unless the image clearly contradicts the prompt.

A pair is treated as invalid only when reliable scale judgment is not possible. 
Invalid cases include missing or unidentifiable anchor/target objects, very blurry objects, fused or merged objects, severe generation artifacts, or duplicate object instances that make it ambiguous which instance should be scored. 
Visible but incorrectly scaled objects are not invalidated; they remain scorable and contribute to scale metrics.

\subsection{Human Consensus Construction}
\label{app:human_consensus}

We collect human annotations on a calibration split sampled from Task~1 and Task~2, covering S1 natural-depth common-object scenes, S2 same-plane common-object scenes, and S3 human--product scenes. 
{The original judge-calibration set covers Tasks~1 and~2; Task~3 inherits their source-task criteria. A separate blinded human study now evaluates source and edited images across tasks (\cref{app:new_humans}).}

For each pair, we aggregate all available valid human scores into a consensus label. 
Let $\mathcal{R}_{p}$ be the multiset of valid ordinal scores assigned to pair $p$. 
The consensus score $c_p$ is the modal score in $\mathcal{R}_{p}$. 
When multiple scores are tied for the mode, we break the tie by taking the median of the tied modal scores and rounding to the nearest ordinal label. 
This preserves the ordinal nature of the rubric and avoids imposing an artificial continuous scale.

We report two forms of human reliability. 
First, in~\cref{tab:human_aggregate_reliability}, each annotator is compared with the aggregate consensus constructed from all available raters. 
This measures agreement with the final calibration target but includes the evaluated annotator in the reference. 
Second, in~\cref{tab:loo_human_reliability}, we rebuild the consensus from the remaining eight annotators and compare the held-out annotator against this leave-one-rater-out reference. 
This removes self-inclusion bias and gives a more conservative estimate of human reliability.

\begin{table}[t]
\centering
\small
\caption{\textbf{Per-annotator agreement with the aggregate human consensus.}
For each annotator, the reference is the aggregate human consensus constructed from all available raters using modal score aggregation with median-rounded tie breaking. 
Annotator identities are anonymized.}
\label{tab:human_aggregate_reliability}
\resizebox{\linewidth}{!}{
\begin{tabular}{lccccccc}
\toprule
Annotator & Pairs & Exact (\%) $\uparrow$ & $\leq 1$ (\%) $\uparrow$ & MAE $\downarrow$ & QWK $\uparrow$ & $r$ $\uparrow$ & Mean diff. \\
\midrule
Annotator A & 506 & 70.95 & 95.26 & 0.3478 & 0.8087 & 0.8181 & +0.0040 \\
Annotator B & 472 & 64.19 & 94.28 & 0.4280 & 0.7125 & 0.7183 & -0.0593 \\
Annotator C & 519 & 65.51 & 94.99 & 0.4027 & 0.7853 & 0.7931 & -0.0366 \\
Annotator D & 517 & 70.41 & 99.03 & 0.3056 & 0.8204 & 0.8404 & -0.0387 \\
Annotator E & 522 & 61.88 & 90.42 & 0.4828 & 0.6006 & 0.6193 & -0.0460 \\
Annotator F & 513 & 70.18 & 96.10 & 0.3411 & 0.7947 & 0.7955 & +0.0331 \\
Annotator G & 511 & 55.19 & 90.02 & 0.5519 & 0.7433 & 0.7700 & -0.0039 \\
Annotator H & 516 & 54.26 & 90.12 & 0.5562 & 0.6796 & 0.6813 & +0.0252 \\
Annotator I & 526 & 73.76 & 96.01 & 0.3023 & 0.8371 & 0.8396 & -0.0133 \\
\midrule
Mean $\pm$ std 
& $511{\pm}16$ 
& $65.15{\pm}6.98$ 
& $94.03{\pm}3.16$ 
& $0.4132{\pm}0.0989$ 
& $0.7536{\pm}0.0773$ 
& $0.7640{\pm}0.0760$ 
& -- \\
Min--Max 
& 472--526 
& 54.26--73.76 
& 90.02--99.03 
& 0.3023--0.5562 
& 0.6006--0.8371 
& 0.6193--0.8404 
& -- \\
\bottomrule
\end{tabular}}
\end{table}

\begin{table}[t]
\centering
\small
\caption{\textbf{Leave-one-rater-out human reliability.}
For each annotator, the reference consensus is rebuilt from the remaining eight annotators only, using modal score aggregation with median-rounded tie breaking. 
This removes self-inclusion bias from the human reliability estimate. Annotator identities are anonymized and match~\cref{tab:human_aggregate_reliability}.}
\label{tab:loo_human_reliability}
\resizebox{\linewidth}{!}{
\begin{tabular}{lccccccc}
\toprule
Annotator & Ref. raters & Pairs & Exact (\%) $\uparrow$ & $\leq 1$ (\%) $\uparrow$ & MAE $\downarrow$ & QWK $\uparrow$ & $r$ $\uparrow$ \\
\midrule
Annotator A & 8 & 504 & 63.69 & 94.84 & 0.4246 & 0.7721 & 0.7862 \\
Annotator B & 8 & 515 & 59.61 & 93.79 & 0.4757 & 0.6783 & 0.6841 \\
Annotator C & 8 & 568 & 57.04 & 93.66 & 0.5000 & 0.7268 & 0.7374 \\
Annotator D & 8 & 558 & 64.70 & 98.39 & 0.3692 & 0.7737 & 0.7996 \\
Annotator E & 8 & 576 & 56.25 & 90.80 & 0.5347 & 0.5663 & 0.5885 \\
Annotator F & 8 & 553 & 60.76 & 95.48 & 0.4412 & 0.7430 & 0.7434 \\
Annotator G & 8 & 564 & 48.23 & 88.48 & 0.6365 & 0.6915 & 0.7271 \\
Annotator H & 8 & 565 & 47.26 & 90.27 & 0.6248 & 0.6397 & 0.6411 \\
Annotator I & 8 & 574 & 64.46 & 94.43 & 0.4146 & 0.7608 & 0.7667 \\
\midrule
Mean $\pm$ std 
& 8 
& $553{\pm}26$ 
& $58.00{\pm}6.56$ 
& $93.35{\pm}3.03$ 
& $0.4913{\pm}0.0928$ 
& $0.7058{\pm}0.0695$ 
& $0.7193{\pm}0.0695$ \\
Min--Max 
& 8 
& 504--576 
& 47.26--64.70 
& 88.48--98.39 
& 0.3692--0.6365 
& 0.5663--0.7737 
& 0.5885--0.7996 \\
\bottomrule
\end{tabular}}
\end{table}

\subsection{Gemini Judge Model, Prompt, and Output Schema}
\label{app:gemini_judge_prompt}

For scalable evaluation, we use Gemini 3.1 Pro Preview (\texttt{gemini-3.1-pro-preview})~\citep{google2026gemini31pro}, calibrated against human consensus. 
The judge receives the same core evidence as human annotators: the generated image, anchor and target names, reference physical lengths, expected 3D target-to-anchor length ratio, and the five-point ordinal rubric. 
When the original generation prompt is provided, it is explicitly treated as low-priority disambiguation evidence only. 
The prompt may help identify intended objects or layout, but it is not allowed to justify a visible scale error simply because the prompt requested accurate scale.

Before pairwise scale scoring, we apply a scene-level validity prefilter to remove images whose visual defects make pairwise scale judgment unreliable. 
The prefilter checks for duplicate objects, extra unnamed clutter, severe generation artifacts, and whether the intended objects are individually clear. 
Images failing this prefilter are excluded from scored-pair metrics, while ordinary scale errors remain scored. 
{Pairwise scoring continues if the scene-prefilter API call fails.}

The full prompt templates used by the released evaluation scripts are provided below. 
Line wrapping is added only for readability. 
Placeholders such as \texttt{[object\_a]}, \texttt{[len\_a\_cm]}, and \texttt{[benchmark\_prompt]} are filled from the benchmark metadata at evaluation time.

\paragraph{Scene-level prefilter prompt.}
The same prefilter template is used before Task~1 and Task~2 pairwise scoring.
\begingroup\small
\begin{verbatim}
You audit a synthetic image BEFORE an automated object size-evaluation pipeline.

Prominent labels from our benchmark (reference only): [primary_object_names].

The list may include evaluator synonyms for the same physical object
(e.g. short vs parenthesized names). Treat those as ONE intended label set --
do not count them as separate extra clutter.

From the image alone, output ONE JSON object with:
- "duplicate_objects" (bool): two+ clearly separate instances of the SAME
  category so it is ambiguous which to judge (e.g. two identical eggs,
  two rulers).
- "extra_unnamed_objects" (bool): major extra props/clutter/repeated shapes
  beyond the intended label set so relative-scale reasoning is unreliable
  (not mere synonyms in the list above).
- "severe_generation_artifacts" (bool): obvious AI flaws (fused objects,
  melted geometry, incoherent boundaries, extra limbs) that would break
  size reasoning.
- "objects_individually_clear" (bool): each listed label could be matched
  to a distinct instance with usable boundaries; false if blur/heavy
  overlap/crop blocks that.

Set "skip_size_correction" (bool) true if ANY of: duplicate_objects,
extra_unnamed_objects, severe_generation_artifacts, OR
objects_individually_clear is false. Add "brief_reason" (string,
<= 35 words, English).

JSON only, no markdown.
\end{verbatim}
\endgroup

\paragraph{Task~1 pairwise judge prompt.}
Task~1 evaluates common-object relative scale. 
For each unordered object pair, the first object is treated as the anchor and the second as the target. 
The template below is instantiated once per pair; in multi-object images, all unordered pairs are scored independently.

\begingroup\small
\begin{verbatim}
You are an expert physical spatial reasoning engine and a professional
photography adjudicator. Your task is to evaluate the physical size accuracy
of objects in the provided generated image. Your scoring should closely mirror
the aggregate judgement of human annotators, not a purely literal physics
calculator.

### GROUND TRUTH REFERENCE:
- Object A (Anchor): [object_a], Typical physical longest edge: [len_a_cm] cm.
- Object B (Target): [object_b], Typical physical longest edge: [len_b_cm] cm.
- Expected 3D longest-edge ratio: Object B is about [ratio]x Object A
  (Object A is about [inverse_ratio]x Object B).

### OPTIONAL GENERATION PROMPT (LOW PRIORITY):
Human annotators did NOT see this text; they only saw the image, object names,
and reference lengths. Use this prompt only to identify intended objects or
ambiguous layout. Do NOT use words like "strictly accurate" as evidence that
the rendered sizes are correct, and do NOT forgive a visible size error because
the prompt intended correctness.
[benchmark_prompt]

### SCENARIO CONTEXT (natural depth and perspective; used for S1):
Objects may intentionally sit at different depths. First infer the 3D layout,
then compare real-world scale. Do not punish an object simply because it looks
visually large/small in 2D if foreground/background placement plausibly
explains it. Conversely, if the TARGET still violates the expected 3D ratio
after this perspective correction, penalize it.

### SCENARIO CONTEXT (same depth plane; used for S2):
This benchmark row is a coplanar layout: objects are meant on the same ground
plane with little depth separation. Use the stated reference lengths and the
visible 3D/2D size ratio as the primary evidence. Do not use depth or
perspective as an excuse unless the image clearly shows large depth separation.
Occlusion, foreshortening, and flexible-object pose still apply, but the
default assumption is direct same-plane comparison.

### EVALUATION PROTOCOL (MATCH THE HUMAN GUI):
Please analyze the image step-by-step:
1. Detection: Are BOTH Object A and Object B clearly identifiable? If either
   is very blurry, missing, fused, or not actually generated, set
   both_objects_present=false.
2. Roles: Object A is the ANCHOR. Treat it as correctly scaled. Score ONLY
   Object B, the TARGET. Do NOT flip roles.
3. Perspective adjustment: Account for near-vs-far perspective before scoring.
   A far target may look smaller; a near target may look larger. Only penalize
   if the TARGET still looks implausibly small/large after that adjustment.
4. Estimate the TARGET's inferred real 3D longest edge relative to the anchor,
   using the expected ratio above. For same-plane scenes, this is close to the
   visible ratio. For natural-depth scenes, first mentally correct for depth.
5. Use the target's full intended extent, not a misleading subpart: e.g. use a
   frisbee/CD/plate diameter rather than rim thickness; use a folded towel's
   visible folded extent, not the unfolded towel length; account for occlusion
   and foreshortening.

### FINAL JUDGMENT:
Assume Object A is its real-world physical size in 3D space. Accounting for
 depth, perspective, occlusion, and realistic configuration
(folding/rolling/foreshortening), how accurate is the size of Object B compared
 to its stated typical length of [len_b_cm] cm?
Use the same quantitative rubric shown to human annotators:
- Score 3 (Proportionate): inferred TARGET size error is within +/-20% of its
  reference length.
- Score 2 (Slightly undersized): TARGET is about 20--60% too small.
- Score 4 (Slightly oversized): TARGET is about 20--60% too large.
- Score 1 (Severely undersized): TARGET is more than 60% too small.
- Score 5 (Severely oversized): TARGET is more than 60% too large.
Important calibration: do NOT default ambiguous cases to 3 if the visible
same-plane ratio clearly crosses the 20% or 60% threshold. At the same time,
do NOT choose 1/5 unless the inferred TARGET/reference error is beyond 60%
after perspective and pose correction.
Select exactly one category from the 1-5 scale below:
1: Severely Undersized
2: Slightly Undersized
3: Proportionate
4: Slightly Oversized
5: Severely Oversized

### OUTPUT FORMAT:
You MUST output your response in valid JSON format. Do not include markdown
code blocks. CRITICAL: Keep BOTH reasoning fields extremely short
(<= 25 words each). Do NOT use ellipses (...).
{
  "reasoning_detection": "...",
  "reasoning_depth_and_perspective": "...",
  "both_objects_present": true,
  "size_score": 3
}
\end{verbatim}
\endgroup

The S1 and S2 scenario-context blocks are mutually exclusive in the implementation: the S1 natural-depth block is inserted only for \texttt{S1\_Natural\_Depth}, and the S2 same-plane block is inserted only for \texttt{S2\_Extreme\_Contrast}. 
The optional generation-prompt block is included only when the benchmark row stores the original generation prompt.

\paragraph{Task~2 pairwise judge prompt.}
Task~2 uses one human--product pair per image. 
The human body part is always the fixed anchor and the product is always the target.
\begingroup\small
\begin{verbatim}
You are an expert physical spatial reasoning engine and a professional
photography adjudicator. Your task is to evaluate the physical size accuracy
of objects in the provided generated image. Your scoring should closely mirror
the aggregate judgement of human annotators using a quick visual GUI, not a
purely literal pixel-measurement or product-spec calculator.

### GROUND TRUTH REFERENCE:
- Object A (Human Anchor): [human_anchor], Typical physical longest edge:
  [len_a_cm] cm.
- Object B (Target Product): [product], Typical physical longest edge:
  [len_b_cm] cm.
- Expected 3D longest-edge ratio: Product B is about [ratio]x the human anchor
  (the anchor is about [inverse_ratio]x Product B).

### OPTIONAL GENERATION PROMPT (LOW PRIORITY):
Human annotators did NOT see this text; they only saw the image, object names,
and reference lengths. Use this prompt only to identify the intended product,
packaging/bundle extent, or ambiguous interaction. Do NOT use exact centimeter
claims in the prompt as a pixel ruler, and do NOT forgive or penalize a size
relationship solely because the prompt intended it.
[benchmark_prompt]

### EVALUATION PROTOCOL (MATCH THE HUMAN GUI):
This task focuses on the direct interaction between a human body (or body part)
and a product. Since the human is interacting with the product, they are usually
roughly at the SAME depth plane, but catalog photos often use close-up framing,
partial hands/faces/feet, foreshortening, and product-forward composition.

Please analyze the image step-by-step:
1. Detection: Are BOTH Object A (Human/part) and Object B (Product) clearly
   identifiable? If either is very blurry, missing, fused, or not actually
   generated, set both_objects_present=false.
2. Roles: Object A is the HUMAN ANCHOR. Treat it as correctly scaled. Score
   ONLY Object B, the TARGET PRODUCT. Do NOT flip roles.
3. Human-anchor caution: hands, heads/faces, feet/legs, and full bodies may be
   cropped, angled, closer to the camera, or only partially visible. Do not
   infer exact centimeters from a cropped palm, a close-up face, or a partial
   foot/leg. Use them as approximate scale references.
4. Product extent: Judge the intended product as presented, not a misleading
   subcomponent. For packs/bundles/stacks, use the full visible pack/bundle
   footprint; for folded bedding/clothing, use the folded visible package; for
   jewelry in a display box, judge the visible retail presentation as plausible
   rather than treating the ring diameter alone as the whole target; for paired
   products (shoes, gloves, etc.), judge the displayed pair/item as a normal
   product presentation and do not double-penalize because two units appear.
5. Size relationship: Ask whether a typical human annotator would immediately
   feel the product is implausibly small/large in this interaction. Do not
   score a catalog-style close-up as oversized merely because the product
   occupies many pixels or is foregrounded.

### FINAL JUDGMENT:
Assume Object A is its real-world physical size in 3D space. How accurate is
 the size of Object B compared to its stated typical length of [len_b_cm] cm?
Use the same quantitative rubric shown to human annotators, but apply it
perceptually rather than with exact pixel calipers:
- Score 3 (Proportionate): inferred product size error is within about +/-20%,
  OR the catalog interaction looks plausible after crop/pose/packaging
  correction.
- Score 2 (Slightly undersized): product is clearly 20--60% too small.
- Score 4 (Slightly oversized): product is clearly 20--60% too large.
- Score 1 (Severely undersized): product is more than 60% too small and looks
  comically/impossibly tiny.
- Score 5 (Severely oversized): product is more than 60% too large and looks
  comically/impossibly huge.
Important Task2 calibration: human annotators usually give Score 3 for
plausible product catalog interactions. Use 4/2 only for obvious size errors,
and use 5/1 very rarely. If the only evidence for 5/1 is an exact ratio
estimate from a cropped hand/head/foot or a close-up product-forward
composition, choose 4/2 or 3 instead.
Select exactly one category from the 1-5 scale below:
1: Severely Undersized
2: Slightly Undersized
3: Proportionate
4: Slightly Oversized
5: Severely Oversized

### OUTPUT FORMAT:
You MUST output your response in valid JSON format. Do not include markdown
code blocks. CRITICAL: Keep BOTH reasoning fields extremely short
(<= 25 words each). Do NOT use ellipses (...).
{
  "reasoning_detection": "...",
  "reasoning_scale_and_interaction": "...",
  "both_objects_present": true,
  "size_score": 3
}
\end{verbatim}
\endgroup

\paragraph{Task~3 judge routing.}
Task~3 does not use a separate VLM judge prompt. 
Each Task~3 entry stores \texttt{source\_task\_type} and \texttt{source\_task\_id}. 
The evaluator splits Task~3 outputs by provenance, constructs temporary Task~1 or Task~2 benchmark rows using the referenced source entries, scores the edited images with the corresponding Task~1 or Task~2 judge above, and then remaps the synthetic identifiers back to the original Task~3 identifiers. 
This ensures that edited images are judged under the same criterion as their original failed generation.

\paragraph{Response-format retry suffix.}
For each pairwise judge call, the evaluator appends the following suffix to the prompt to reduce invalid JSON responses:
\begingroup\small
\begin{verbatim}
CRITICAL: Output MUST be minified JSON in a SINGLE LINE. The JSON MUST include
all required keys and end with a closing brace '}'. Keep BOTH reasoning fields
<= 25 words each.
\end{verbatim}
\endgroup

\subsection{Repeated Sampling and Score Aggregation}
\label{app:repeated_sampling}

For each scorable anchor--target pair, we query the Gemini judge five times. 
Repeated sampling reduces sensitivity to isolated parsing errors, unstable visual interpretations, or overly literal ratio estimates. 
Each call returns a numeric score in $\{1,2,3,4,5\}$ and a pair-validity flag. 
A pair is included in the final scale metrics only when the required objects are judged present and the scene passes the validity criteria.

Given the five sampled scores for a pair, we first use majority vote. 
If there is no unique mode, we take the median of the five ordinal scores. 
This aggregation preserves the discrete ordinal scale while reducing the influence of a single outlier sample. 
All benchmark models, general-purpose editors, and \modelname{} outputs are evaluated with the same fixed judge, prompt templates, repeated-sampling protocol, and aggregation rule.

\subsection{Calibration Diagnostics}
\label{app:calibration_diagnostics}

We evaluate calibration using exact agreement, within-one agreement, mean absolute error (MAE), Pearson correlation $r$, and quadratic-weighted kappa (QWK) on the five-point ordinal scale. 
Exact agreement measures strict label equality, while within-one agreement measures whether two labels differ by at most one ordinal level. 
MAE is computed as the mean absolute difference between predicted and consensus ordinal scores. 
QWK is useful because it penalizes large ordinal disagreements more heavily than adjacent disagreements.

{\cref{tab:human_aggregate_reliability,tab:loo_human_reliability} show consistent human agreement with both aggregate and leave-one-rater-out consensus.}

{Gemini matches or exceeds the leave-one-rater-out human reference on all five agreement metrics (\cref{tab:human_gemini_alignment,tab:gemini_human_bootstrap}), supporting scalable automatic evaluation.}

\subsection{Bootstrap Confidence Intervals}
\label{app:bootstrap_ci}

We report uncertainty for Gemini--human alignment using image-level bootstrap confidence intervals in~\cref{tab:gemini_human_bootstrap}. 
The bootstrap resamples images rather than individual object pairs, and all pairwise judgments associated with a sampled image are included together. 
{This preserves the grouping of pairwise judgments within each image.} 
For each bootstrap replicate, we recompute exact agreement, within-one agreement, MAE, Pearson correlation, and QWK. 
The reported intervals are percentile confidence intervals over the resulting bootstrap distribution.

\begin{table}[t]
\begingroup
\centering
\small
\caption{\textbf{Gemini--human alignment with image-level bootstrap confidence intervals.}
Gemini 3.1 Pro Preview is evaluated with five repeated samples per pair and majority/median aggregation. 
Confidence intervals are computed by resampling images rather than individual object pairs.}
\label{tab:gemini_human_bootstrap}
\resizebox{\linewidth}{!}{
\begin{tabular}{lcccccc}
\toprule
Split & Images / Pairs & Exact (\%) $\uparrow$ & $\leq 1$ (\%) $\uparrow$ & MAE $\downarrow$ & $r$ $\uparrow$ & QWK $\uparrow$ \\
\midrule
Task 1: S1--S2 & 185 / 427 & \makecell[c]{61.12\\[-0.25ex]{\scriptsize [56.29,66.11]}} & \makecell[c]{96.02\\[-0.25ex]{\scriptsize [93.59,98.10]}} & \makecell[c]{0.4286\\[-0.25ex]{\scriptsize [0.3695,0.4870]}} & \makecell[c]{0.8460\\[-0.25ex]{\scriptsize [0.8117,0.8774]}} & \makecell[c]{0.8337\\[-0.25ex]{\scriptsize [0.7950,0.8671]}} \\
Task 2: S3 & 100 / 100 & \makecell[c]{73.00\\[-0.25ex]{\scriptsize [65.00,81.00]}} & \makecell[c]{100.00\\[-0.25ex]{\scriptsize [100.00,100.00]}} & \makecell[c]{0.2700\\[-0.25ex]{\scriptsize [0.1900,0.3500]}} & \makecell[c]{0.4701\\[-0.25ex]{\scriptsize [0.2346,0.6544]}} & \makecell[c]{0.4677\\[-0.25ex]{\scriptsize [0.2263,0.6487]}} \\
Task 1+2 & 285 / 527 & \makecell[c]{63.38\\[-0.25ex]{\scriptsize [58.58,68.27]}} & \makecell[c]{96.77\\[-0.25ex]{\scriptsize [94.87,98.46]}} & \makecell[c]{0.3985\\[-0.25ex]{\scriptsize [0.3439,0.4521]}} & \makecell[c]{0.8317\\[-0.25ex]{\scriptsize [0.7959,0.8648]}} & \makecell[c]{0.8203\\[-0.25ex]{\scriptsize [0.7800,0.8558]}} \\
\bottomrule
\end{tabular}}
\endgroup
\end{table}

\section{\modelname{} Implementation Details}
\label{app:rescale_details}

This appendix expands the implementation of \modelname{}, the model-agnostic scale-correction pipeline used in Sec.~\ref{sec:rescale}. We focus on the agentic edit plan, the local insertion interface, the training design of our depth-aware correction backend, and the backend/training ablations.

\subsection{Agent Inputs and Edit-Plan Format}
\label{app:rescale_agent_plan}

{\modelname{} takes a generated image $I$, object identities, and physical-size references. For human--product scenes it also receives the product reference image and dimensions. The main experiments provide evaluator feedback with relation scores, object-presence information, and scale explanations to guide object selection, resize direction, and factor estimation. Bounding boxes and masks are obtained by grounding and segmentation. Task~3 uses this same feedback-guided protocol on the 100 unique source images; its correction output is shared across the corresponding S4/S5 benchmark entries. Fully autonomous operation is evaluated separately in~\cref{app:new_ablation}.}

The edit planner in our released code uses Gemini 3.1 Pro Preview (\texttt{gemini-3.1-pro-preview})~\citep{google2026gemini31pro} for Task~1 and defaults to Gemini 3 Flash Preview (\texttt{gemini-3-flash-preview})~\citep{google2025gemini3flashpreview} for Tasks~2 and~3.
The agent converts these inputs into a structured local edit plan. For each edit round $k$, the plan is
\begin{equation}
\pi^{(k)} = \{o_t, o_a, \mathbf{b}_t, \tilde{\mathbf{b}}_t, \alpha, p, q\},
\end{equation}
where $o_t$ is the editable target object, $o_a$ is the fixed anchor object, $\mathbf{b}_t$ is the current target bounding box, $\tilde{\mathbf{b}}_t$ is the resized target box, $\alpha$ is the multiplicative resize factor, $p$ is a contact-preserving anchor point, and $q$ is a short natural-language rationale used for verification. The resized box $\tilde{\mathbf{b}}_t$ is obtained by scaling $\mathbf{b}_t$ by $\alpha$ around $p$, so that contact points such as object bases, hand-contact regions, or support surfaces remain approximately fixed.

For common-object images, multiple pairwise judgments can implicate the same object. The agent therefore aggregates pairwise evidence into a conservative object-level plan: it edits the object that most consistently explains the observed scale errors and avoids large changes when the pairwise evidence is contradictory. For human--product images, the product is treated as editable and the human body part is fixed. {Evaluator scores constrain the resize direction and plausible scale interval, while the agent combines this evidence with scene geometry to plan the edit.} If the agent cannot localize the objects reliably, finds no clear scale inconsistency, or predicts that the edit would create severe collisions, truncation, or boundary artifacts, \modelname{} returns a no-edit decision.

\subsection{Localization, Segmentation, and Local Editing Interface}
\label{app:rescale_local_interface}

Given an edit plan, \modelname{} standardizes all downstream backends to the same local insertion interface. We first refine the target box with multimodal localization and segment the target instance using SAM~2~\citep{ravi2024sam2}. The segmented target is extracted before removal and used as the appearance reference $R$. We then remove the original target from $I$ to obtain a completed background $B$. {An off-the-shelf inpainting/editing model completes the background after object removal.}

The edit mask $M$ is constructed around the resized box $\tilde{\mathbf{b}}_t$ and is deliberately enlarged beyond the expected object support. This gives the insertion backend enough local context for contact shadows, occlusion boundaries, and small background corrections. We also estimate a monocular depth map with DepthAnythingV2~\citep{yang2024depthanythingv2}. For depth-aware backends, we fuse the completed-background depth with the resized foreground support to form a correction-aware depth condition $D$. The resulting backend inputs are therefore
\begin{equation}
(R, B, M, D, \tau),
\end{equation}
where $\tau$ is a compact text instruction describing the target object and the intended resize operation. Backends that do not support depth receive the same reference, background, mask, and text instruction, but ignore $D$. This interface lets us substitute only the final local generation model while keeping agent planning, localization, segmentation, background completion, and mask construction fixed.

\subsection{Depth-Aware Correction Backend}
\label{app:rescale_training_details}

\begin{figure}[t]
    \centering
    \includegraphics[width=0.9\linewidth]{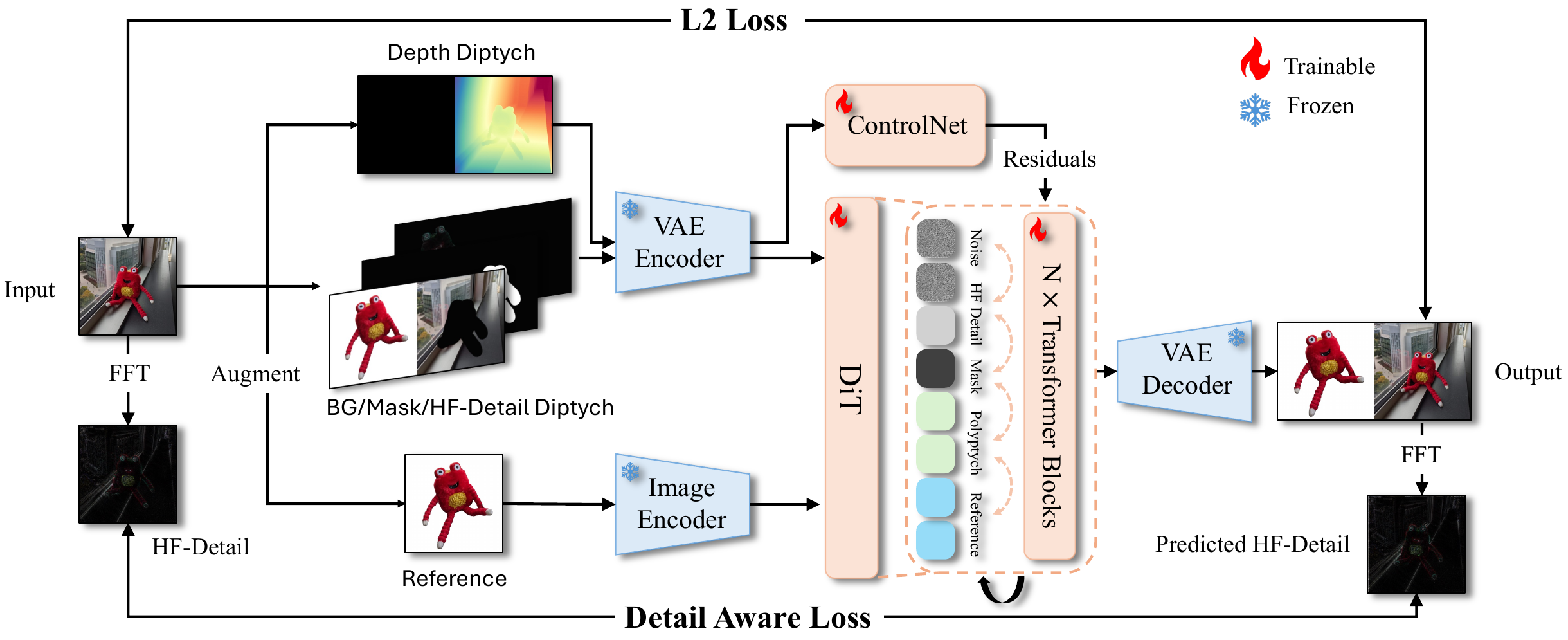}
    \vspace{-0.5ex}
    \caption{\textbf{Training pipeline of our depth-aware correction backend.}
    The model adopts a diptych-style formulation, with a reference half that provides object identity and appearance, and a target half that contains the completed background and the local region to be corrected. 
    A DiT-based inpainting backbone performs the local reinsertion, a depth ControlNet injects geometry-aware signals from the fused depth condition, and an optional high-frequency branch provides detail cues derived from the reference crop via FFT. 
    Training combines the backbone objective with an optional detail-aware reconstruction loss to improve fidelity after resizing.}
    \label{fig:rescale_training_app}
    \vspace{-1.5ex}
\end{figure}

{The depth-aware local insertion backend is designed for controlled resizing and reinsertion.} As shown in \cref{fig:rescale_training_app}, the model follows a diptych-style in-context formulation: the left half provides a clean reference crop of the object to preserve identity and appearance, while the right half contains the target-side completed background together with the local region to be edited. This design is more suitable for scale correction than a standard inpainting setup, because the model must simultaneously preserve object identity and synthesize a resized insertion that matches the surrounding scene.

Let $R$ denote the clean reference crop, $B$ the completed target background after removing the original instance, $I$ the desired corrected image, $M$ the editable target-side mask, and $D$ the fused depth condition. We construct the training inputs as
\begin{equation}
\mathcal{S} = [R \mid B], \qquad
\mathcal{Y} = [R \mid I], \qquad
\mathcal{M} = [0 \mid M], \qquad
\mathcal{D} = [0 \mid D],
\end{equation}
where $[\cdot \mid \cdot]$ denotes horizontal concatenation. The left half serves as visual reference, and the right half specifies the local correction problem. During training, $B$ is obtained by removing the ground-truth foreground object from the original image; at inference time, it is replaced by the completed background produced by the upstream removal stage of \modelname{}.

The DiT inpainting backbone is initialized from InsertAnything~\citep{song2025insertanything}, which builds on FLUX.1 Fill [dev]~\citep{bfl2024fluxfill}. The reference crop is encoded by a frozen reference image encoder, while the target half is synthesized by the inpainting transformer. Because relative-scale correction is fundamentally a geometric edit, we add a depth ControlNet branch~\citep{zhang2023controlnet} and feed it the depth diptych $\mathcal{D}$. Its residuals are injected only into the target-half tokens, so that the reference side remains an appearance cue rather than becoming a second geometry target. This design encourages the model to use depth primarily to control the support and extent of the resized object in the target scene.

We also test an optional high-frequency branch for fine-detail preservation. Specifically, we extract an FFT-based high-frequency representation from the reference crop and form
\begin{equation}
\mathcal{H} = [\mathrm{HF}(R) \mid 0],
\end{equation}
which is projected and injected into the transformer hidden states through a lightweight branch. The motivation is that local resizing can easily smooth textures, edges, and small appearance cues; the high-frequency branch gives the model an explicit signal for recovering such details from the reference object.

A key motivation of this backend is to decouple \emph{editable context extent} from \emph{object extent}. In ordinary mask-conditioned insertion, the model can overfit to the mask boundary and treat it as the intended object boundary. That behavior is undesirable for scale correction, where the mask should reserve enough local context for shadows, contact regions, and boundary adaptation, but the resized object should not necessarily expand to fill the whole editable region. We therefore deliberately perturb and dilate the training masks, and rely on depth as the main cue for the intended object support. 

We train only lightweight adaptation modules: LoRA adapters~\citep{hu2021lora} on the DiT backbone, the depth ControlNet branch, and, when enabled, the high-frequency injection branch. The VAE, text encoder, and reference encoder remain frozen. The primary optimization objective is the latent flow-matching objective of the underlying DiT backbone,
\begin{equation}
\mathcal{L}_{\mathrm{fm}}
=
\left\|v_{\theta}(x_t,c) - (x_1-x_0)\right\|_2^2,
\end{equation}
where $x_t$ is the interpolated noisy latent, $x_0$ is the clean latent, $x_1$ is Gaussian noise, and $c$ denotes the full set of conditioning inputs. To better preserve local details after resizing, we optionally add a detail-aware reconstruction loss on high-frequency maps,
\begin{equation}
\mathcal{L}_{\mathrm{da}}
=
\left\|
\mathrm{HF}(\hat{I}) \odot M - \mathrm{HF}(I) \odot M
\right\|_2^2,
\end{equation}
and optimize the combined objective
\begin{equation}
\mathcal{L}
=
\mathcal{L}_{\mathrm{fm}}
+
\lambda_{\mathrm{da}} \mathcal{L}_{\mathrm{da}}.
\end{equation}

\paragraph{Training data construction.}
We construct correction-style training tuples from the same family of segmentation, video-object, saliency, fashion, and insertion datasets used by the insertion backbone, including SA-1B~\citep{kirillov2023segmentanything}, LVIS~\citep{Gupta19lvis}, saliency datasets, YouTube-VOS~\citep{xu2018youtubevos}, VIPSeg~\citep{miao2022vipseg}, MOSE~\citep{ding2023mose}, VITON-HD~\citep{choi2021vitonhd}, and AnyInsertion~\citep{song2025insertanything}. Each training sample is converted into a reference--target diptych by extracting a foreground reference crop, removing the original instance from the target side, and using the original image as supervision. We further apply appearance and geometric augmentations to the reference crop, synthetic partial occlusion to make the reference less idealized, and mask perturbation/dilation so that the model cannot trivially infer object extent from mask shape alone. {These choices tailor training to scale correction through task-specific priors and target-preserving augmentation, as used in other reconstruction settings~\citep{zhang2025radarode,zhang2025horcrux}: references may be incomplete, and the editable region includes both the resized object and its surrounding context.}

\subsection{Insertion Backend Substitution Study}
\label{app:backend_substitution}

\cref{tab:ablation_results} evaluates the final local generation step while holding the rest of \modelname{} fixed. For each input, we reuse the same agent plan, target box, segmentation mask, completed background, reference crop, and depth estimate, and replace only the insertion backend. This isolates whether differences come from the local synthesis model rather than from upstream diagnosis or preprocessing.

{\cref{tab:ablation_results} distinguishes the capabilities of interchangeable insertion backends. InsertAnything combines strong reference consistency with high no-reference quality, while the depth-aware backend provides the highest SSIM and SSIM-HF. CreatiLayout uses text and boxes; the reference-conditioned backends additionally receive the object's appearance. }

\begin{table}[t]
    \centering
    \caption{\textbf{Insertion backend substitution under fixed \modelname{} plans.}
    We keep agent planning, localization, segmentation, background completion, mask construction, and depth estimation fixed, and replace only the final local generation backend. Metrics are computed on Task~1 edited-object bounding boxes; higher is better.}
    \label{tab:ablation_results}
    \vspace{0.25em}

    \setlength{\tabcolsep}{6pt}
    \renewcommand{\arraystretch}{1.15}

    \resizebox{0.9\linewidth}{!}{
    \begin{tabular}{lcccccc}
        \toprule

        & \multicolumn{4}{c}{\textbf{Visual Consistency}} 
        & \multicolumn{2}{c}{\textbf{Generation Quality}} \\

        \cmidrule(lr){2-5}
        \cmidrule(lr){6-7}

        \textbf{Insertion backend} 
        & \textbf{CLIP-I (\%) $\uparrow$}
        & \textbf{DINO (\%) $\uparrow$}
        & \textbf{SSIM (\%) $\uparrow$}
        & \textbf{SSIM-HF (\%) $\uparrow$}
        & \textbf{LAION-Aes $\uparrow$}
        & \textbf{Q-Align-IQ $\uparrow$} \\

        \midrule

        Ours (depth-aware) & 86.5 & 70.8 & \textbf{50.2} & \textbf{75.6} & 3.68 & 3.58 \\
        InsertAnything~\citep{song2025insertanything} & \textbf{87.7} & \textbf{75.6} & 48.8 & 75.0 & \textbf{3.71} & \textbf{3.61} \\
        CreatiLayout~\citep{zhang2025creatilayout} & 75.9 & 38.3 & 31.4 & 70.7 & 3.30 & 2.46 \\
        Bifrost~\citep{li2024bifrost} & 82.5 & 66.2 & 40.3 & 73.5 & 3.65 & 3.44 \\

        \bottomrule
    \end{tabular}
    }
\vspace{-2ex}
\end{table}

\subsection{Correction-Model Training Ablation}
\label{app:training_ablation}

{\cref{tab:pretrain_ablation} identifies guidance and depth conditioning as the main contributors to the depth-aware backend. Guidance improves feature consistency and no-reference quality; depth further improves both identity and structural measurements. The detail-aware loss and high-frequency branch improve DINO, SSIM, and SSIM-HF, strengthening detail preservation. Together, the components support localized synthesis with explicit control of geometric support.}

\begin{table}[t]
    \centering
    \caption{\textbf{Training-design ablation for the depth-aware correction backend.}
    Each row incrementally adds a component to the same base diptych architecture: FLUX-style guidance conditioning, depth ControlNet conditioning, and the detail-aware loss plus high-frequency branch. Metrics are computed on Task~1 edited-object bounding boxes; higher is better.}
    \label{tab:pretrain_ablation}
    \vspace{0.25em}

    \setlength{\tabcolsep}{6pt}
    \renewcommand{\arraystretch}{1.15}

    \resizebox{0.9\linewidth}{!}{
    \begin{tabular}{lcccccc}
        \toprule

        & \multicolumn{4}{c}{\textbf{Visual Consistency}}
        & \multicolumn{2}{c}{\textbf{Generation Quality}} \\

        \cmidrule(lr){2-5}
        \cmidrule(lr){6-7}

        \textbf{Training configuration}
        & \textbf{CLIP-I (\%) $\uparrow$}
        & \textbf{DINO (\%) $\uparrow$}
        & \textbf{SSIM (\%) $\uparrow$}
        & \textbf{SSIM-HF (\%) $\uparrow$}
        & \textbf{LAION-Aes $\uparrow$}
        & \textbf{Q-Align-IQ $\uparrow$} \\

        \midrule

        Baseline & 74.2 & 30.6 & 44.1 & \textbf{75.8} & 3.19 & 1.91 \\
        Baseline + guidance & 77.6 & 44.8 & 40.8 & 74.2 & 3.40 & 2.13 \\
        Baseline + guidance + depth & \textbf{79.4} & 47.1 & 45.0 & 75.6 & \textbf{3.58} & \textbf{2.19} \\
        Baseline + guidance + depth + DAL + HF & \textbf{79.4} & \textbf{47.3} & \textbf{45.3} & \textbf{75.8} & 3.57 & \textbf{2.19} \\

        \bottomrule
    \end{tabular}
    }
\vspace{-2ex}
\end{table}

\section{Full Quantitative Results}
\label{app:full_results}

\providecommand{\appci}[3]{\makecell[c]{#1\\[-0.35ex]{\scriptsize [#2,#3]}}}

\subsection{Metrics and Statistical Protocol}
\label{app:full_results_metrics}
{Unless explicitly marked as cross-judge or human-panel results, quantitative results use the calibrated Gemini judge described in \cref{app:human_vlm_eval}.}  The atomic unit is an anchor--target pair rather than an image.  For each valid pair, the judge returns an ordinal scale score $s\in\{1,2,3,4,5\}$, where $s=3$ denotes a physically plausible target size relative to the anchor, $s<3$ denotes an undersized target, and $s>3$ denotes an oversized target.  We report scale error as $|s-3|$ averaged over scored pairs.  Plausible is the fraction of pairs with $s=3$; Too small and Too large are the fractions with $s\in\{1,2\}$ and $s\in\{4,5\}$, respectively; Severe is the fraction with $s\in\{1,5\}$.

{Scale metrics exclude pairs with absent, ambiguously duplicated, or unidentifiable objects and apply the scene-level prefilter.}  Valid img. / pairs denotes the number of images with at least one scored pair and the number of scored pairs after filtering.  {Bracketed ranges denote 95\% confidence intervals from 2,000 bootstrap resamples. Task~1 resamples images with their relations together; Task~3 groups S4/S5 entries by their shared source image.}  For one-pass generation or editing results, CIs are reported for pair-level mean scale error.  For correction experiments, matched-pair tables only include pairs that are scored both before and after editing; CIs are reported for before/after mean scale error and Gain.  Gain is the reduction in matched-pair scale error, so positive values indicate improvement.  B / W counts pairs whose absolute error $|s-3|$ becomes smaller or larger after editing.  Directional rate columns report empirical percentages.

\subsection{Task 1 Full Results}
\label{app:task1_full_results}
\cref{tab:app_task1_full_results} expands the main Task~1 table with valid coverage and the full error-direction breakdown.  MR denotes mean-regression error: cases where the physical target is smaller than the anchor but judged too large, or physically larger than the anchor but judged too small.

\begin{table*}[t]
\begingroup
\centering
\small
\caption{\textbf{Task 1 full results.} Results are split by S1, S2, and all Task~1 entries.  Error is mean $|s-3|$ with 95\% CI in brackets.  Plausible denotes score $3$; Too small denotes scores $1$--$2$; Too large denotes scores $4$--$5$; Severe denotes scores $1$ or $5$.  MR denotes mean-regression error.}
\label{tab:app_task1_full_results}
\renewcommand\arraystretch{1.05}
\setlength{\tabcolsep}{3.4pt}
\resizebox{\textwidth}{!}{
\begin{tabular}{@{}llccccccc@{}}
\toprule
Model & Split & Valid img. / pairs & Error $\downarrow$ & Plaus. (\%) $\uparrow$ & Too small (\%) $\downarrow$ & Too large (\%) $\downarrow$ & Severe (\%) $\downarrow$ & MR (\%) $\downarrow$ \\
\midrule
\multirow{3}{*}{Nano Banana 2} & S1 & 199 / 564 & \appci{0.505}{0.430}{0.581} & 65.1 & 17.0 & 17.9 & 15.6 & 15.8 \\
 & S2 & 196 / 557 & \appci{0.691}{0.614}{0.771} & 47.8 & 25.1 & 27.1 & 16.9 & 47.9 \\
 & All & 395 / 1121 & \appci{0.598}{0.542}{0.650} & 56.5 & 21.1 & 22.5 & 16.2 & 31.8 \\
\addlinespace[0.35ex]
\multirow{3}{*}{GPT-Image-2} & S1 & 199 / 564 & \appci{0.784}{0.691}{0.879} & 51.1 & 20.7 & 28.2 & 29.4 & 12.1 \\
 & S2 & 200 / 572 & \appci{0.629}{0.552}{0.707} & 52.8 & 22.7 & 24.5 & 15.7 & 35.8 \\
 & All & 399 / 1136 & \appci{0.706}{0.650}{0.765} & 51.9 & 21.7 & 26.3 & 22.5 & 24.0 \\
\addlinespace[0.35ex]
\multirow{3}{*}{Z-Image-Turbo} & S1 & 193 / 532 & \appci{0.643}{0.562}{0.723} & 53.4 & 22.6 & 24.1 & 17.7 & 35.7 \\
 & S2 & 181 / 470 & \appci{0.881}{0.803}{0.960} & 38.1 & 33.0 & 28.9 & 26.2 & 55.7 \\
 & All & 374 / 1002 & \appci{0.754}{0.695}{0.816} & 46.2 & 27.4 & 26.3 & 21.7 & 45.1 \\
\addlinespace[0.35ex]
\multirow{3}{*}{Grok Imagine} & S1 & 198 / 562 & \appci{0.600}{0.519}{0.685} & 58.7 & 19.9 & 21.4 & 18.7 & 26.9 \\
 & S2 & 200 / 573 & \appci{0.979}{0.906}{1.055} & 34.7 & 31.1 & 34.2 & 32.6 & 61.1 \\
 & All & 398 / 1135 & \appci{0.791}{0.733}{0.851} & 46.6 & 25.6 & 27.8 & 25.7 & 44.1 \\
\addlinespace[0.35ex]
\multirow{3}{*}{Qwen-Image 2512} & S1 & 196 / 539 & \appci{0.703}{0.623}{0.786} & 53.1 & 22.6 & 24.3 & 23.4 & 37.5 \\
 & S2 & 199 / 562 & \appci{1.105}{1.036}{1.179} & 27.0 & 35.8 & 37.2 & 37.5 & 67.6 \\
 & All & 395 / 1101 & \appci{0.908}{0.850}{0.968} & 39.8 & 29.3 & 30.9 & 30.6 & 52.9 \\
\addlinespace[0.35ex]
\multirow{3}{*}{FLUX.2} & S1 & 176 / 448 & \appci{0.828}{0.734}{0.920} & 45.5 & 26.6 & 27.9 & 28.3 & 40.4 \\
 & S2 & 178 / 490 & \appci{1.212}{1.141}{1.285} & 23.3 & 38.8 & 38.0 & 44.5 & 69.4 \\
 & All & 354 / 938 & \appci{1.029}{0.966}{1.088} & 33.9 & 32.9 & 33.2 & 36.8 & 55.5 \\
\addlinespace[0.35ex]
\multirow{3}{*}{SD3.5-Large} & S1 & 180 / 501 & \appci{0.956}{0.869}{1.039} & 36.7 & 32.7 & 30.5 & 32.3 & 46.7 \\
 & S2 & 166 / 444 & \appci{1.189}{1.111}{1.275} & 24.5 & 36.5 & 39.0 & 43.5 & 70.0 \\
 & All & 346 / 945 & \appci{1.066}{1.006}{1.127} & 31.0 & 34.5 & 34.5 & 37.6 & 57.7 \\
\addlinespace[0.35ex]
\bottomrule
\end{tabular}}
\endgroup
\end{table*}

\subsection{Task 2 Full Results}
\label{app:task2_full_results}
\cref{tab:app_task2_full_results} expands the main Task~2 table with the directional error rates.  Because Task~2 contains one human--product relation per image, image-level and pair-level coverage are nearly identical after filtering.

\begin{table*}[t]
\begingroup
\centering
\small
\caption{\textbf{Task 2 full results.} Each Task~2 image contains one product--human scale relation.  Error is mean $|s-3|$ with 95\% CI in brackets.  The remaining columns report the full score-direction breakdown for S3.}
\label{tab:app_task2_full_results}
\renewcommand\arraystretch{1.08}
\setlength{\tabcolsep}{5.0pt}
\resizebox{1\textwidth}{!}{
\begin{tabular}{@{}lcccccc@{}}
\toprule
Model & Valid img. / pairs & Error $\downarrow$ & Plaus. (\%) $\uparrow$ & Too small (\%) $\downarrow$ & Too large (\%) $\downarrow$ & Severe (\%) $\downarrow$ \\
\midrule
GPT-Image-2 & 294 / 294 & \appci{0.231}{0.180}{0.289} & 78.9 & 3.4 & 17.7 & 2.0 \\
Nano Banana 2 & 295 / 295 & \appci{0.268}{0.210}{0.325} & 75.6 & 4.4 & 20.0 & 2.4 \\
Seedream 4.5 & 295 / 295 & \appci{0.302}{0.237}{0.366} & 74.9 & 8.1 & 16.9 & 5.1 \\
\makecell[l]{Qwen-Image-Edit-2511} & 297 / 297 & \appci{0.327}{0.266}{0.391} & 71.0 & 4.4 & 24.6 & 3.7 \\
FLUX.1 Kontext-dev & 291 / 291 & \appci{0.423}{0.354}{0.491} & 63.2 & 15.1 & 21.6 & 5.5 \\
\makecell[l]{SD3.5-Large +\\IP-Adapter} & 269 / 269 & \appci{0.595}{0.513}{0.677} & 52.8 & 7.8 & 39.4 & 12.3 \\
\bottomrule
\end{tabular}}
\endgroup
\end{table*}

\subsection{Task 3 Full Results}
\label{app:task3_full_results}
\cref{tab:app_task3_full_results} reports the full Task~3 results for general-purpose editors.  S4 requires autonomous scale-error discovery, while S5 gives the target object, reference object, resize direction, and scale factor.  The before-edit row is the erroneous source image set used to construct Task~3.

\begin{table*}[t]
\begingroup
\centering
\small
\caption{\textbf{Task 3 full results for general-purpose editors.} Error and score-direction rates are computed on each editor's valid scored outputs.  Error and Gain include 95\% CIs in brackets.  Gain and B / W are computed on matched before--after pairs relative to the corresponding before-edit source image.  Valid coverage is reported separately from scale accuracy.}
\label{tab:app_task3_full_results}
\renewcommand\arraystretch{1.05}
\setlength{\tabcolsep}{3.2pt}
\resizebox{\textwidth}{!}{
\begin{tabular}{@{}llcccccccc@{}}
\toprule
Model & Split & Valid img. / pairs & Error $\downarrow$ & Plaus. (\%) $\uparrow$ & Too small (\%) $\downarrow$ & Too large (\%) $\downarrow$ & Severe (\%) $\downarrow$ & Gain $\uparrow$ & B / W \\
\midrule
\multirow{3}{*}{Before edit} & S4 & 98 / 98 & \appci{1.265}{1.112}{1.418} & 20.4 & 31.6 & 48.0 & 46.9 & -- & -- \\
 & S5 & 98 / 98 & \appci{1.265}{1.112}{1.418} & 20.4 & 31.6 & 48.0 & 46.9 & -- & -- \\
 & All & 196 / 196 & \appci{1.265}{1.112}{1.418} & 20.4 & 31.6 & 48.0 & 46.9 & -- & -- \\
\addlinespace[0.35ex]
\multirow{3}{*}{GPT-Image-2} & S4 & 97 / 97 & \appci{0.959}{0.784}{1.124} & 40.2 & 28.9 & 30.9 & 36.1 & \appci{+0.333}{0.198}{0.469} & 31 / 4 \\
 & S5 & 98 / 98 & \appci{0.408}{0.286}{0.531} & 66.3 & 16.3 & 17.3 & 7.1 & \appci{+0.866}{0.680}{1.052} & 60 / 5 \\
 & All & 195 / 195 & \appci{0.682}{0.565}{0.799} & 53.3 & 22.6 & 24.1 & 21.5 & \appci{+0.601}{0.474}{0.737} & 91 / 9 \\
\addlinespace[0.35ex]
\multirow{3}{*}{Nano Banana 2} & S4 & 97 / 97 & \appci{1.010}{0.855}{1.165} & 32.0 & 25.8 & 42.3 & 33.0 & \appci{+0.242}{0.105}{0.379} & 24 / 7 \\
 & S5 & 98 / 98 & \appci{0.929}{0.776}{1.082} & 33.7 & 25.5 & 40.8 & 26.5 & \appci{+0.340}{0.206}{0.474} & 31 / 4 \\
 & All & 195 / 195 & \appci{0.969}{0.839}{1.108} & 32.8 & 25.6 & 41.5 & 29.7 & \appci{+0.292}{0.181}{0.408} & 55 / 11 \\
\addlinespace[0.35ex]
\multirow{3}{*}{FLUX.1 Kontext-dev} & S4 & 99 / 99 & \appci{1.293}{1.141}{1.444} & 21.2 & 30.3 & 48.5 & 50.5 & \appci{-0.021}{-0.113}{0.072} & 8 / 9 \\
 & S5 & 94 / 94 & \appci{0.957}{0.787}{1.117} & 36.2 & 25.5 & 38.3 & 31.9 & \appci{+0.280}{0.140}{0.419} & 27 / 7 \\
 & All & 193 / 193 & \appci{1.130}{0.989}{1.270} & 28.5 & 28.0 & 43.5 & 41.5 & \appci{+0.126}{0.036}{0.215} & 35 / 16 \\
\addlinespace[0.35ex]
\multirow{3}{*}{Qwen-Image-Edit-2511} & S4 & 98 / 98 & \appci{1.245}{1.092}{1.398} & 23.5 & 28.6 & 48.0 & 48.0 & \appci{+0.021}{-0.082}{0.113} & 9 / 4 \\
 & S5 & 91 / 91 & \appci{1.000}{0.824}{1.165} & 34.1 & 25.3 & 40.7 & 34.1 & \appci{+0.253}{0.099}{0.407} & 23 / 5 \\
 & All & 189 / 189 & \appci{1.127}{0.984}{1.269} & 28.6 & 27.0 & 44.4 & 41.3 & \appci{+0.133}{0.038}{0.223} & 32 / 9 \\
\addlinespace[0.35ex]
\multirow{3}{*}{Seedream 4.5} & S4 & 100 / 100 & \appci{1.170}{1.010}{1.330} & 28.0 & 37.0 & 35.0 & 45.0 & \appci{+0.102}{-0.031}{0.235} & 18 / 8 \\
 & S5 & 96 / 96 & \appci{1.104}{0.938}{1.281} & 35.4 & 31.2 & 33.3 & 45.8 & \appci{+0.200}{0.042}{0.368} & 25 / 9 \\
 & All & 196 / 196 & \appci{1.138}{1.000}{1.283} & 31.6 & 34.2 & 34.2 & 45.4 & \appci{+0.150}{0.026}{0.267} & 43 / 17 \\
\addlinespace[0.35ex]
\multirow{3}{*}{\makecell[l]{SD3.5-Large +\\IP-Adapter}} & S4 & 14 / 14 & \appci{0.857}{0.429}{1.286} & 42.9 & 7.1 & 50.0 & 28.6 & \appci{-0.071}{-0.500}{0.429} & 3 / 5 \\
 & S5 & 59 / 59 & \appci{1.305}{1.085}{1.508} & 22.0 & 44.1 & 33.9 & 52.5 & \appci{-0.158}{-0.369}{0.053} & 7 / 13 \\
 & All & 73 / 73 & \appci{1.219}{1.000}{1.437} & 26.0 & 37.0 & 37.0 & 47.9 & \appci{-0.141}{-0.338}{0.056} & 10 / 18 \\
\bottomrule
\end{tabular}}
\endgroup
\end{table*}

\subsection{\modelname{} Correction on Task 1}
\label{app:rescale_task1_full_results}
\cref{tab:app_rescale_task1_full_results} reports matched-pair before--after results for applying \modelname{} to Task~1 generations from each source model.  The table uses only pairs scored on both the original and corrected image.

\begin{table*}[t]
\begingroup
\centering
\small
\caption{\textbf{\modelname{} correction on Task 1.} Metrics are computed on matched scorable pairs before and after correction.  Error before, Error after, and Gain include 95\% CIs in brackets.  Plaus. and Severe are reported as before / after percentages.}
\label{tab:app_rescale_task1_full_results}
\renewcommand\arraystretch{1.05}
\setlength{\tabcolsep}{3.3pt}
\resizebox{\textwidth}{!}{
\begin{tabular}{@{}llccccccc@{}}
\toprule
Source model & Protocol & Matched pairs & Error before $\downarrow$ & Error after $\downarrow$ & Gain $\uparrow$ & Plaus. before / after (\%) $\uparrow$ & Severe before / after (\%) $\downarrow$ & B / W \\
\midrule
\multirow{3}{*}{Nano Banana 2} & S1 & 534 & \appci{0.494}{0.417}{0.573} & \appci{0.390}{0.311}{0.467} & \appci{+0.105}{0.039}{0.174} & 65.7 / 73.8 & 15.2 / 12.7 & 106 / 61 \\
 & S2 & 544 & \appci{0.688}{0.610}{0.768} & \appci{0.379}{0.311}{0.443} & \appci{+0.309}{0.231}{0.386} & 48.2 / 69.3 & 16.9 / 7.2 & 188 / 58 \\
 & All & 1078 & \appci{0.592}{0.537}{0.650} & \appci{0.384}{0.334}{0.435} & \appci{+0.208}{0.155}{0.263} & 56.9 / 71.5 & 16.0 / 9.9 & 294 / 119 \\
\addlinespace[0.35ex]
\multirow{3}{*}{GPT-Image-2} & S1 & 501 & \appci{0.760}{0.665}{0.861} & \appci{0.471}{0.391}{0.557} & \appci{+0.289}{0.203}{0.380} & 53.1 / 68.7 & 29.1 / 15.8 & 139 / 44 \\
 & S2 & 556 & \appci{0.622}{0.543}{0.695} & \appci{0.417}{0.338}{0.492} & \appci{+0.205}{0.127}{0.287} & 53.4 / 66.9 & 15.6 / 8.6 & 158 / 70 \\
 & All & 1057 & \appci{0.688}{0.630}{0.750} & \appci{0.443}{0.387}{0.499} & \appci{+0.245}{0.192}{0.304} & 53.3 / 67.7 & 22.0 / 12.0 & 297 / 114 \\
\addlinespace[0.35ex]
\multirow{3}{*}{Z-Image-Turbo} & S1 & 496 & \appci{0.643}{0.564}{0.732} & \appci{0.371}{0.295}{0.452} & \appci{+0.272}{0.191}{0.359} & 53.4 / 71.8 & 17.7 / 8.9 & 160 / 54 \\
 & S2 & 441 & \appci{0.889}{0.808}{0.974} & \appci{0.499}{0.423}{0.571} & \appci{+0.390}{0.297}{0.495} & 37.9 / 61.0 & 26.8 / 10.9 & 184 / 59 \\
 & All & 937 & \appci{0.759}{0.702}{0.823} & \appci{0.431}{0.378}{0.485} & \appci{+0.328}{0.265}{0.394} & 46.1 / 66.7 & 22.0 / 9.8 & 344 / 113 \\
\addlinespace[0.35ex]
\multirow{3}{*}{Grok Imagine} & S1 & 527 & \appci{0.586}{0.505}{0.674} & \appci{0.334}{0.271}{0.402} & \appci{+0.252}{0.173}{0.341} & 59.8 / 75.7 & 18.4 / 9.1 & 147 / 54 \\
 & S2 & 566 & \appci{0.968}{0.900}{1.041} & \appci{0.553}{0.477}{0.635} & \appci{+0.415}{0.327}{0.501} & 35.2 / 58.1 & 32.0 / 13.4 & 241 / 66 \\
 & All & 1093 & \appci{0.784}{0.721}{0.844} & \appci{0.447}{0.393}{0.500} & \appci{+0.337}{0.275}{0.401} & 47.0 / 66.6 & 25.4 / 11.3 & 388 / 120 \\
\addlinespace[0.35ex]
\multirow{3}{*}{Qwen-Image 2512} & S1 & 487 & \appci{0.704}{0.620}{0.787} & \appci{0.460}{0.377}{0.544} & \appci{+0.244}{0.145}{0.349} & 53.2 / 68.6 & 23.6 / 14.6 & 151 / 68 \\
 & S2 & 542 & \appci{1.111}{1.039}{1.180} & \appci{0.387}{0.338}{0.446} & \appci{+0.723}{0.644}{0.797} & 26.8 / 66.6 & 37.8 / 5.4 & 315 / 39 \\
 & All & 1029 & \appci{0.918}{0.861}{0.977} & \appci{0.422}{0.370}{0.472} & \appci{+0.497}{0.428}{0.571} & 39.3 / 67.5 & 31.1 / 9.7 & 466 / 107 \\
\addlinespace[0.35ex]
\multirow{3}{*}{FLUX.2} & S1 & 403 & \appci{0.806}{0.712}{0.910} & \appci{0.422}{0.341}{0.514} & \appci{+0.385}{0.286}{0.488} & 47.1 / 70.0 & 27.8 / 12.2 & 148 / 41 \\
 & S2 & 439 & \appci{1.216}{1.136}{1.294} & \appci{0.617}{0.530}{0.704} & \appci{+0.599}{0.497}{0.703} & 23.2 / 54.4 & 44.9 / 16.2 & 224 / 40 \\
 & All & 842 & \appci{1.020}{0.955}{1.091} & \appci{0.524}{0.462}{0.591} & \appci{+0.496}{0.422}{0.568} & 34.7 / 61.9 & 36.7 / 14.3 & 372 / 81 \\
\addlinespace[0.35ex]
\multirow{3}{*}{SD3.5-Large} & S1 & 457 & \appci{0.943}{0.859}{1.033} & \appci{0.611}{0.514}{0.707} & \appci{+0.333}{0.249}{0.429} & 37.6 / 58.0 & 31.9 / 19.0 & 174 / 69 \\
 & S2 & 372 & \appci{1.169}{1.079}{1.262} & \appci{0.653}{0.552}{0.751} & \appci{+0.516}{0.415}{0.622} & 25.3 / 54.6 & 42.2 / 19.9 & 167 / 39 \\
 & All & 829 & \appci{1.045}{0.980}{1.109} & \appci{0.630}{0.560}{0.700} & \appci{+0.415}{0.348}{0.486} & 32.1 / 56.5 & 36.6 / 19.4 & 341 / 108 \\
\addlinespace[0.35ex]
\bottomrule
\end{tabular}}
\endgroup
\end{table*}

\subsection{\modelname{} Correction on Task 2}
\label{app:rescale_task2_full_results}
\cref{tab:app_rescale_task2_full_results} reports matched-pair \modelname{} correction results for Task~2.  Since each Task~2 image has one evaluated pair, matched pairs are equivalent to matched images after filtering.

\begin{table*}[t]
\centering
\small
\caption{\textbf{\modelname{} correction on Task 2.} Metrics are computed on matched scorable pairs before and after correction.  Error before, Error after, and Gain include 95\% CIs in brackets.  Plaus. and Severe are reported as before / after percentages.}
\label{tab:app_rescale_task2_full_results}
\renewcommand\arraystretch{1.08}
\setlength{\tabcolsep}{3.8pt}
\resizebox{1\textwidth}{!}{
\begin{tabular}{@{}lccccccc@{}}
\toprule
Source model & Matched pairs & Error before $\downarrow$ & Error after $\downarrow$ & Gain $\uparrow$ & Plaus. before / after (\%) $\uparrow$ & Severe before / after (\%) $\downarrow$ & B / W \\
\midrule
GPT-Image-2 & 290 & \appci{0.231}{0.176}{0.286} & \appci{0.090}{0.055}{0.128} & \appci{+0.141}{0.100}{0.186} & 79.0 / 92.1 & 2.1 / 1.0 & 40 / 1 \\
Nano Banana 2 & 291 & \appci{0.261}{0.210}{0.320} & \appci{0.086}{0.052}{0.120} & \appci{+0.175}{0.124}{0.227} & 76.3 / 92.4 & 2.4 / 1.0 & 54 / 7 \\
Seedream 4.5 & 285 & \appci{0.295}{0.228}{0.361} & \appci{0.168}{0.119}{0.221} & \appci{+0.126}{0.077}{0.179} & 75.8 / 86.3 & 5.3 / 3.2 & 40 / 9 \\
\makecell[l]{Qwen-Image-Edit-2511} & 284 & \appci{0.306}{0.246}{0.373} & \appci{0.144}{0.099}{0.194} & \appci{+0.162}{0.102}{0.222} & 73.2 / 87.7 & 3.9 / 2.1 & 51 / 12 \\
FLUX.1 Kontext-dev & 281 & \appci{0.406}{0.338}{0.477} & \appci{0.228}{0.171}{0.292} & \appci{+0.178}{0.121}{0.235} & 65.1 / 81.5 & 5.7 / 4.3 & 53 / 9 \\
\makecell[l]{SD3.5-Large +\\IP-Adapter} & 246 & \appci{0.565}{0.480}{0.655} & \appci{0.256}{0.191}{0.321} & \appci{+0.309}{0.240}{0.382} & 54.5 / 78.5 & 11.0 / 4.1 & 69 / 4 \\
\bottomrule
\end{tabular}}
\end{table*}

\subsection{\modelname{} Correction on Task 3}
\label{app:rescale_task3_full_results}
{\cref{tab:app_rescale_task3_full_results} reports feedback-guided \modelname{} once on the unique Task~3 source images. General-purpose editors receive the S4 or S5 prompts (\cref{tab:app_task3_full_results}); \modelname{} uses evaluator feedback as specified in~\cref{app:rescale_agent_plan}. The 93 matched source relations define the correction statistics and confidence intervals.}

\begin{table*}[t]
\begingroup
\centering
\small
\caption{\textbf{\modelname{} correction on Task 3.} Metrics are computed on matched scorable pairs before and after correction.  Error before, Error after, and Gain include 95\% CIs in brackets.  Plaus. and Severe are reported as before / after percentages.}
\label{tab:app_rescale_task3_full_results}
\renewcommand\arraystretch{1.08}
\setlength{\tabcolsep}{4.2pt}
\resizebox{0.95\textwidth}{!}{
\begin{tabular}{@{}lccccccc@{}}
\toprule
Protocol & Matched pairs & Error before $\downarrow$ & Error after $\downarrow$ & Gain $\uparrow$ & Plaus. before / after (\%) $\uparrow$ & Severe before / after (\%) $\downarrow$ & B / W \\
\midrule
Feedback-guided & 93 & \appci{1.258}{1.086}{1.419} & \appci{0.548}{0.409}{0.688} & \appci{+0.710}{0.484}{0.914} & 21.5 / 58.1 & 47.3 / 12.9 & 51 / 8 \\
\bottomrule
\end{tabular}}
\endgroup
\end{table*}

\subsection{Identity Preservation and Visual-Quality Metrics}
\label{app:identity_quality_full_results}
\cref{tab:app_rescale_quality_full} reports visual consistency and no-reference quality metrics for \modelname{} corrections.  {CLIP-I~\citep{Radford21CLIP} and DINO~\citep{Oquab23dinov2} measure reference consistency between the original and corrected images. SSIM~\citep{Wang04ssim} and SSIM-HF~\citep{Wang04ssim,Kanopoulos88sobel} measure image-level and high-frequency preservation. LAION-Aes~\citep{Schuhmann22LAIONAesthetics} and Q-Align-IQ~\citep{Wu24QAlign} measure no-reference image quality before and after correction.}

\begin{table*}[t]
\centering
\small
\caption{\textbf{Identity preservation and visual quality after \modelname{} correction.} Higher values indicate better preservation or quality.  {LAION-Aes and Q-Align-IQ percentages are $100(\mathrm{after}-\mathrm{before})/\mathrm{before}$, computed from the displayed means.}}
\label{tab:app_rescale_quality_full}
\renewcommand\arraystretch{1.05}
\setlength{\tabcolsep}{5.0pt}
\resizebox{0.95\textwidth}{!}{
\begin{tabular}{@{}lcccccc@{}}
\toprule
Setting & CLIP-I (\%) $\uparrow$ & DINO (\%) $\uparrow$ & SSIM (\%) $\uparrow$ & SSIM-HF (\%) $\uparrow$ & LAION-Aes before / after $\uparrow$ & Q-Align-IQ before / after $\uparrow$ \\
\midrule
Task 1 & 94.8 & 90.2 & 88.8 & 92.4 & 5.83 / 5.73 ($-1.7\%$) & 4.74 / 4.67 ($-1.5\%$) \\
Task 2 & 92.4 & 84.5 & 73.5 & 82.1 & 4.99 / 4.96 ({$-0.6\%$}) & 4.88 / 4.88 ($0.0\%$) \\
Task 3 & 95.6 & 88.9 & 89.3 & 92.9 & 5.83 / 5.73 ({$-1.7\%$}) & 4.76 / 4.67 ($-1.9\%$) \\
\bottomrule
\end{tabular}}
\end{table*}

\section{Additional Qualitative Results}
\label{app:qualitative}

This section provides additional qualitative examples for the three benchmark tasks and for \modelname{} correction.
{The examples illustrate task inputs, model outputs, and local scale corrections.}

\subsection{Benchmark Record Examples}
\label{app:qualitative_records}

To make the benchmark format explicit, we show one randomly sampled entry from each scenario in \cref{fig:qual_record_s1,fig:qual_record_s2,fig:qual_record_s3,fig:qual_record_s4,fig:qual_record_s5}.
For readability, absolute local path prefixes are shortened to dataset-relative or generated-image-relative paths, while the benchmark fields are otherwise preserved.

\begin{figure}[p]
\centering
\small
\renewcommand{\arraystretch}{1.12}
\begin{tabular}{@{}p{0.24\linewidth}p{0.71\linewidth}@{}}
\toprule
\textbf{Field} & \textbf{Value} \\
\midrule
\texttt{task\_id} & \texttt{T1\_0082} \\
\texttt{scenario} & \texttt{S1\_Natural\_Depth} \\
\texttt{num\_objects} & 2 \\
\texttt{objects\_included} & computer mouse; towel \\
\texttt{prompt} &
Strictly accurate real-world physical proportions, natural depth and perspective.
A tiny computer mouse rests slightly in front of a large, folded bath towel, emphasizing their size contrast.
They sit isolated on the smooth concrete floor of a vast, empty photography studio, casting natural shadows in clear depth.
Photorealistic. \\
\texttt{gt\_ratios} &
\texttt{computer mouse\_to\_towel}: target ratio 0.084; acceptable range [0.081, 0.087]. \\
\texttt{reference\_image\_path} & null \\
\bottomrule
\end{tabular}
\caption{\textbf{Example benchmark record for S1: Natural Depth.}
Task~1 S1 entries contain only a text prompt and pairwise physical-ratio metadata.
No metric object dimensions or reference image are provided to the generator.}
\label{fig:qual_record_s1}
\vspace{-1ex}
\end{figure}

\begin{figure}[p]
\centering
\small
\renewcommand{\arraystretch}{1.12}
\begin{tabular}{@{}p{0.24\linewidth}p{0.71\linewidth}@{}}
\toprule
\textbf{Field} & \textbf{Value} \\
\midrule
\texttt{task\_id} & \texttt{T1\_0238} \\
\texttt{scenario} & \texttt{S2\_Extreme\_Contrast} \\
\texttt{num\_objects} & 3 \\
\texttt{objects\_included} & violin; billiard ball; bowl \\
\texttt{prompt} &
A wooden violin, a polished billiard ball, and a ceramic bowl rest side-by-side on the endless grey floor of a massive, empty concrete warehouse.
The sharp, photorealistic lighting emphasizes the textural contrast between the wood, resin, and ceramic against the stark, neutral background.
Photorealistic, objects placed on the exact same depth plane, strictly accurate real-world physical proportions. \\
\texttt{gt\_ratios} &
\texttt{violin\_to\_billiard ball}: target ratio 10.526; acceptable range [10.120, 10.948].
\newline
\texttt{violin\_to\_bowl}: target ratio 4.000; acceptable range [3.843, 4.163].
\newline
\texttt{billiard ball\_to\_bowl}: target ratio 0.380; acceptable range [0.365, 0.395]. \\
\texttt{reference\_image\_path} & null \\
\bottomrule
\end{tabular}
\caption{\textbf{Example benchmark record for S2: Same Plane.}
S2 uses the same common-object metadata as S1, but prompts constrain objects to a shared depth plane so that image-space size differences more directly expose real-world scale errors.}
\label{fig:qual_record_s2}
\vspace{-1ex}
\end{figure}

\begin{figure}[p]
\centering
\small
\renewcommand{\arraystretch}{1.08}
\begin{tabular}{@{}p{0.24\linewidth}p{0.71\linewidth}@{}}
\toprule
\textbf{Field} & \textbf{Value} \\
\midrule
\texttt{task\_id} & \texttt{T2\_0202} \\
\texttt{scenario} & \texttt{S3\_Human\_Product\_Anchor} \\
\texttt{num\_objects} & 2 \\
\texttt{objects\_included} & trash can; human foot/leg \\
\texttt{prompt} &
A photorealistic, premium catalog shot of a rectangular steel slim trash can.
The bin stands about 64.5 centimeters tall, reaching just below the knee of a person standing next to it, with a length of 46.5 centimeters and a width of 29.5 centimeters.
A human foot wearing a casual sneaker is naturally pressing down on the step pedal at the base of the bin to open the lid, while the lower leg is visible beside it to demonstrate the scale.
The camera captures the scene from a mild three-quarter perspective with bright, even studio lighting.
The sleek metallic silhouette of the trash can remains entirely unobstructed to clearly display its real-world bulk and aspect ratio.
Strict real-world physical proportions between the human foot, leg, and the product. \\
\texttt{gt\_ratios} &
\texttt{trash can\_to\_human foot/leg}: target ratio 2.434; acceptable range [2.271, 2.613]. \\
\texttt{reference\_image\_path} & \texttt{ABO/images/original/6b/6bca188f.jpg} \\
\texttt{raw\_listing\_title} & Amazon Basics Rectangular Soft Close Steel Slim Trash Can 40 \\
\texttt{refine\_meta} & refined with Gemini; model \texttt{gemini-3.1-pro-preview}. \\
\texttt{product\_scale} &
source: ABO Dataset; category id: \texttt{B07PCXZ14V}; length 46.51 cm; width 29.49 cm; height 64.49 cm; typical length 64.49 cm; acceptable range [63.20, 65.78] cm. \\
\bottomrule
\end{tabular}
\caption{\textbf{Example benchmark record for S3: Human--Product.}
Task~2 entries include a product reference image, product-level metric dimensions, a human anchor, and a single product--human scale relation.}
\label{fig:qual_record_s3}
\vspace{-1ex}
\end{figure}

\begin{figure}[p]
\centering
\small
\renewcommand{\arraystretch}{1.10}
\begin{tabular}{@{}p{0.24\linewidth}p{0.71\linewidth}@{}}
\toprule
\textbf{Field} & \textbf{Value} \\
\midrule
\texttt{task\_id} & \texttt{T3\_0166} \\
\texttt{scenario} & \texttt{S4\_Hard\_Auto\_Discovery} \\
\texttt{source\_task\_id} & \texttt{T2\_0140} \\
\texttt{source\_task\_type} & \texttt{T2} \\
\texttt{prompt\_type} & \texttt{hard\_auto\_discovery} \\
\texttt{image\_path} & \texttt{generated/task2/Qwen\_Image\_Edit\_2511\_1024/T2\_0140.png} \\
\texttt{objects\_included} & wine bottle; human head/face \\
\texttt{gt\_ratios} &
\texttt{wine bottle\_to\_human head/face}: target ratio 1.270; acceptable range [1.185, 1.364]. \\
\texttt{prompt} &
Check whether object size proportions in this image are unrealistic.
Reference lengths --- ``wine bottle'': approximately 30.48 cm characteristic, LxWxH 8.3 x 30.5 x 30.5 cm; human part: ``human head/face'' approximately 24.0 cm.
If wrong relative to these references, rescale ``wine bottle'' only; if already plausible, keep the image unchanged.
Preserve object identity, pose, viewpoint, composition, background, and lighting. \\
\texttt{reference\_image\_path} & \texttt{ABO/images/original/52/52d1c3b3.jpg} \\
\texttt{source\_size\_score} & 4 \\
\bottomrule
\end{tabular}
\caption{\textbf{Example benchmark record for S4: Hard Auto-Discovery.}
S4 provides the source image and scale references without prescribing a resize direction or factor. For human--product cases, the prompt identifies the product as editable and keeps the human anchor fixed.}
\label{fig:qual_record_s4}
\vspace{-1ex}
\end{figure}

\begin{figure}[p]
\centering
\small
\renewcommand{\arraystretch}{1.10}
\begin{tabular}{@{}p{0.24\linewidth}p{0.71\linewidth}@{}}
\toprule
\textbf{Field} & \textbf{Value} \\
\midrule
\texttt{task\_id} & \texttt{T3\_0013} \\
\texttt{scenario} & \texttt{S5\_Precise\_Scale\_Instruction} \\
\texttt{source\_task\_id} & \texttt{T1\_0321} \\
\texttt{source\_task\_type} & \texttt{T1} \\
\texttt{prompt\_type} & \texttt{precise\_scale\_instruction} \\
\texttt{image\_path} & \texttt{generated/task1/FLUX\_2\_Fal/T1\_0321.png} \\
\texttt{objects\_included} & car; tennis racket \\
\texttt{gt\_ratios} &
\texttt{car\_to\_tennis racket}: target ratio 6.560; acceptable range [6.303, 6.827]. \\
\texttt{prompt} &
Shrink the tennis racket by an exact scale factor of 0.790000 while keeping the car completely unchanged.
Strictly preserve the original identity, background, lighting, and composition, making absolutely no extra edits beyond this size correction. \\
\texttt{edit\_plan} &
edit target: tennis racket; reference object: car; scale factor: 0.790; direction: shrink. \\
\texttt{pair\_key} & \texttt{car\_to\_tennis racket} \\
\texttt{generated\_ratio} & 5.200 \\
\texttt{target\_ratio} & 6.560 \\
\texttt{reference\_image\_path} & null \\
\texttt{source\_size\_score} & 4 \\
\bottomrule
\end{tabular}
\caption{\textbf{Example benchmark record for S5: Precise Scale Instruction.}
S5 uses the same type of erroneous source image as S4, but directly specifies the target object, reference object, resize direction, and numeric scale factor.}
\label{fig:qual_record_s5}
\vspace{-1ex}
\end{figure}

\subsection{Task 1: Common-Object Generation Examples}
\label{app:qualitative_task1}

\begin{figure}[p]
    \centering
    \includegraphics[width=\linewidth]{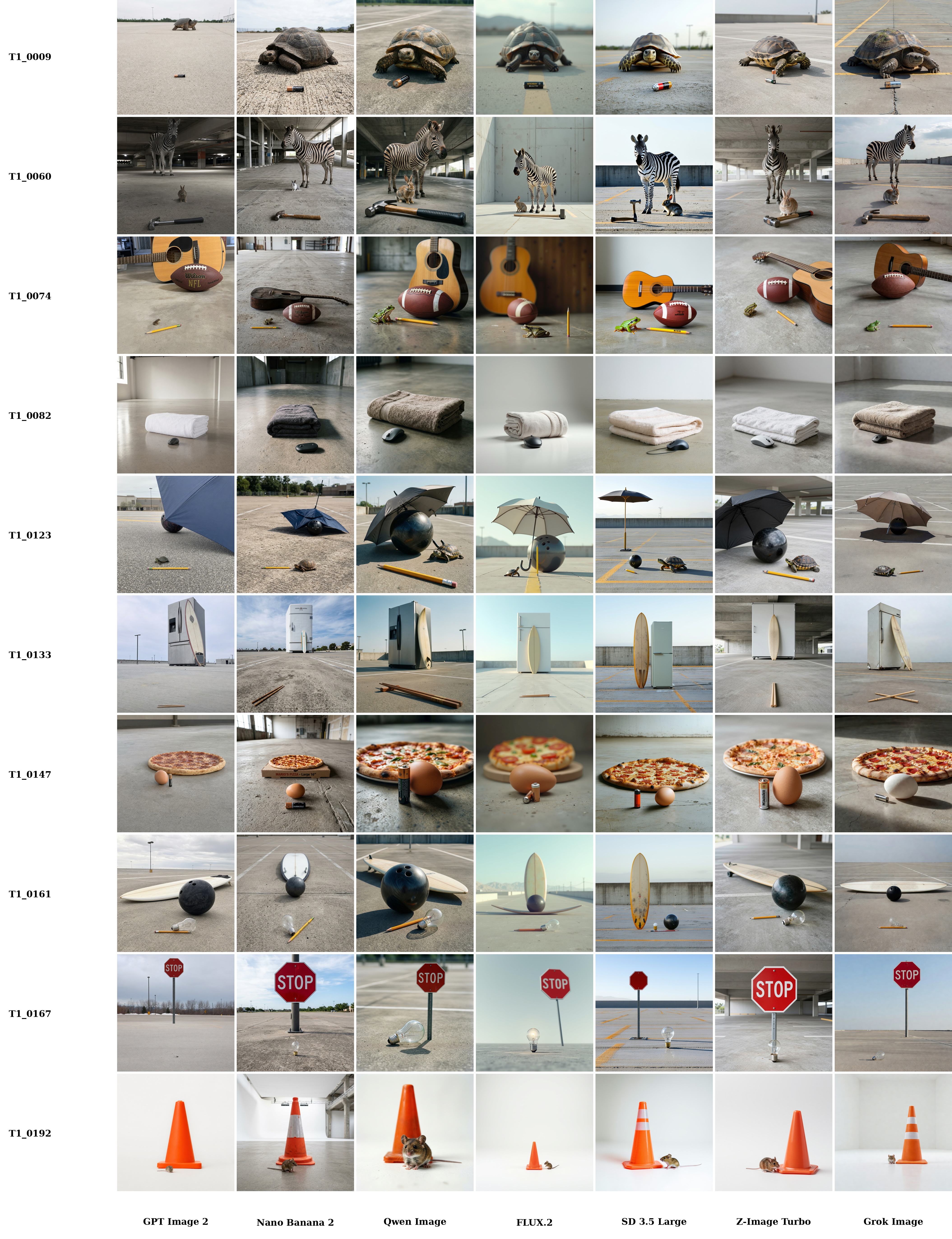}
    \vspace{-0.5ex}
    \caption{\textbf{Task~1 S1 Natural Depth examples.}
    Each row corresponds to one common-object prompt and each column corresponds to an evaluated generator.
    S1 allows perspective and depth ordering, so generated images may use near--far placement while still being evaluated for real-world relative scale.}
    \label{fig:qual_task1_s1}
    \vspace{-1ex}
\end{figure}

\begin{figure}[p]
    \centering
    \includegraphics[width=\linewidth]{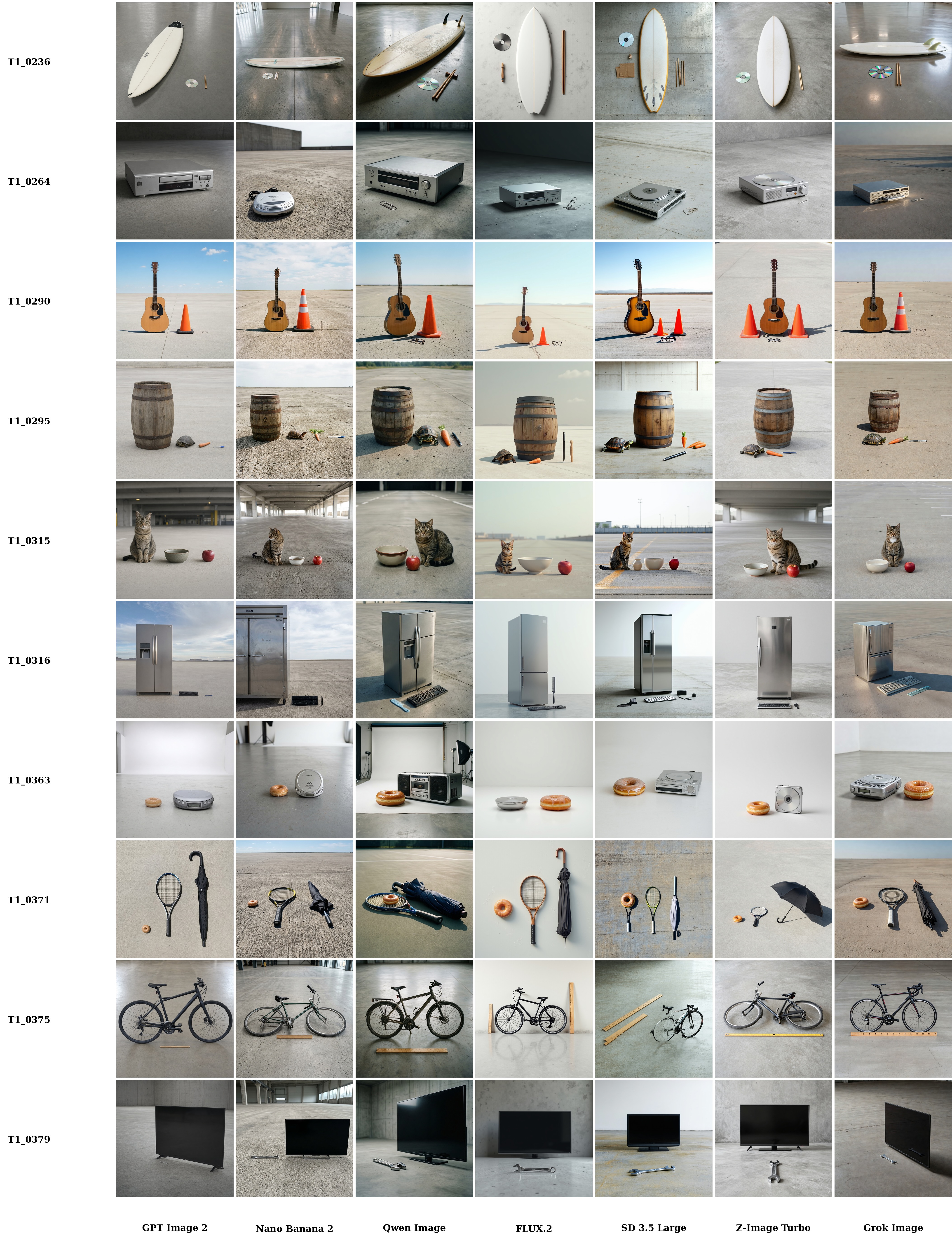}
    \vspace{-0.5ex}
    \caption{\textbf{Task~1 S2 Same-Plane examples.}
    Objects are prompted to lie on the same depth plane, reducing perspective ambiguity and making relative scale errors more directly visible in image space.}
    \label{fig:qual_task1_s2}
    \vspace{-1ex}
\end{figure}

\cref{fig:qual_task1_s1,fig:qual_task1_s2} show additional Task~1 generations across the evaluated text-to-image models.
S1 permits natural depth variation, while S2 constrains objects to approximately the same plane, making relative-size errors more visually exposed.

\subsection{Task 2: Human-Product Generation Examples}
\label{app:qualitative_task2}

\begin{figure}[p]
    \centering
    \includegraphics[width=\linewidth]{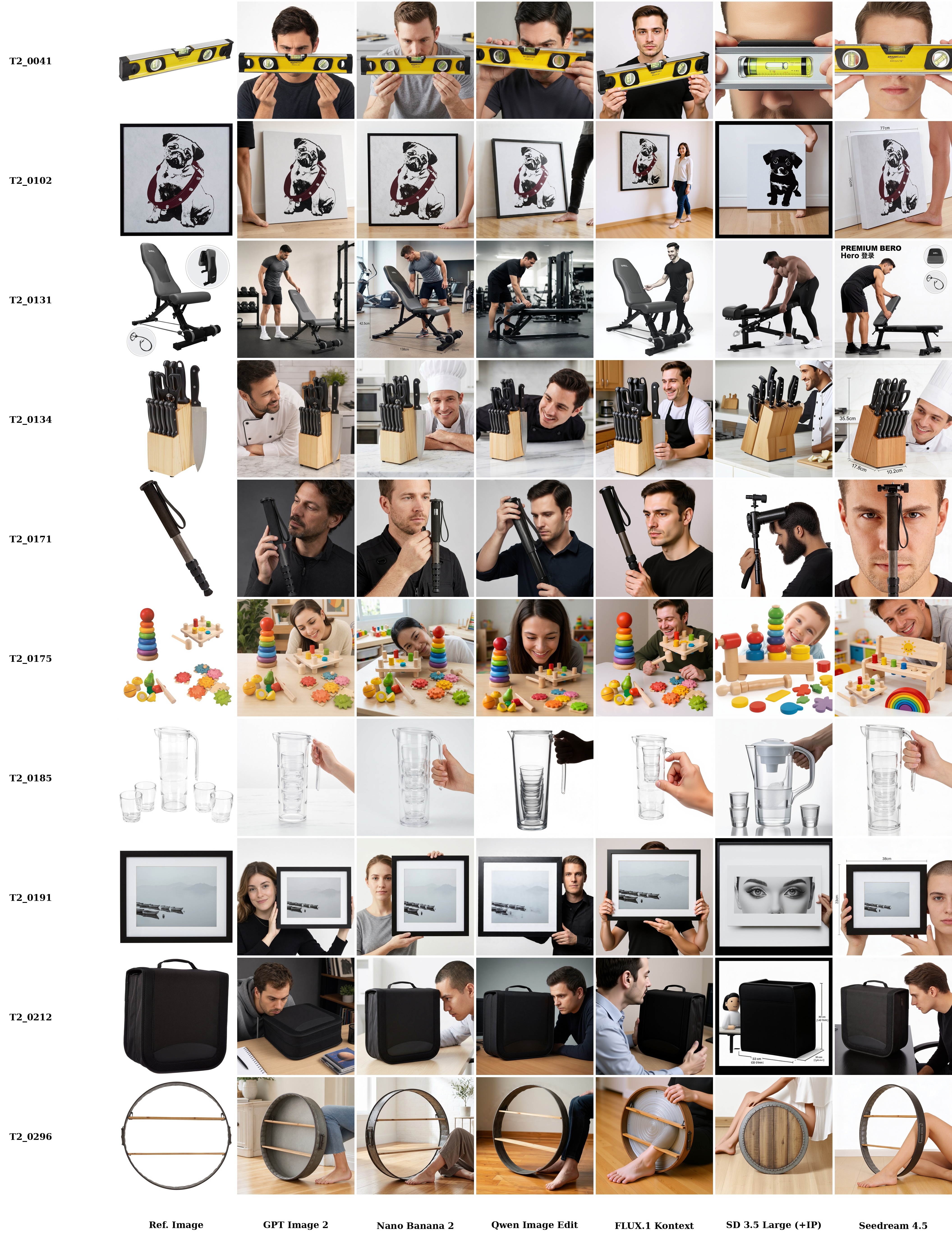}
    \vspace{-0.5ex}
    \caption{\textbf{Task~2 human-product generation examples.}
    Each row contains one product reference image and the corresponding model outputs.
    Prompts provide product dimensions and require a human anchor, so the task tests whether the model can translate explicit metric information into plausible human--product proportions.}
    \label{fig:qual_task2}
    \vspace{-1ex}
\end{figure}

\cref{fig:qual_task2} shows additional Task~2 examples.
The leftmost column shows the product reference image, and the remaining columns show image-conditioned generations from different models.
These examples illustrate that preserving product identity and realizing metric scale relative to a human anchor are distinct requirements.

\subsection{Task 3: General-Purpose Editor Correction Examples}
\label{app:qualitative_task3}

\begin{figure}[p]
    \centering
    \includegraphics[width=\linewidth]{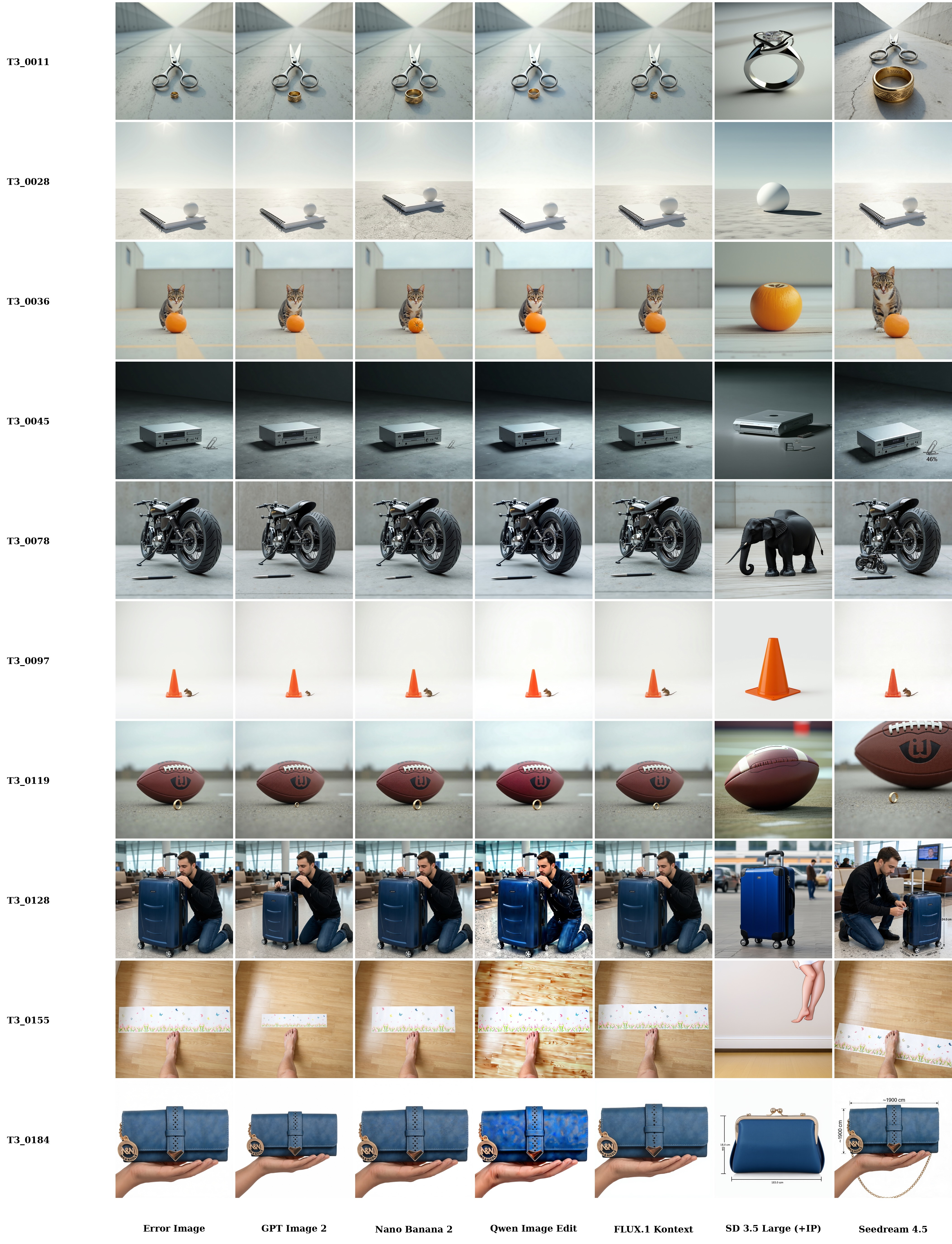}
    \vspace{-0.5ex}
    \caption{\textbf{Task~3 general-purpose editor correction examples.}
    The source image contains a known scale error from Task~1 or Task~2.
    General-purpose editors are asked to correct the error, either through auto-discovery or through a precise resize instruction depending on the scenario.}
    \label{fig:qual_task3_editor}
    \vspace{-1ex}
\end{figure}

\cref{fig:qual_task3_editor} shows qualitative results for general-purpose editors on Task~3.
The first column is the erroneous source image, and the remaining columns are editor outputs under the corresponding correction prompt.
The examples illustrate the diagnosis--execution gap: editors often preserve visual realism but may under-correct, over-correct, or regenerate content beyond the intended localized scale edit.

\subsection{\modelname{} Correction Examples}
\label{app:qualitative_rescale}

\begin{figure}[p]
    \centering
    \includegraphics[width=\linewidth]{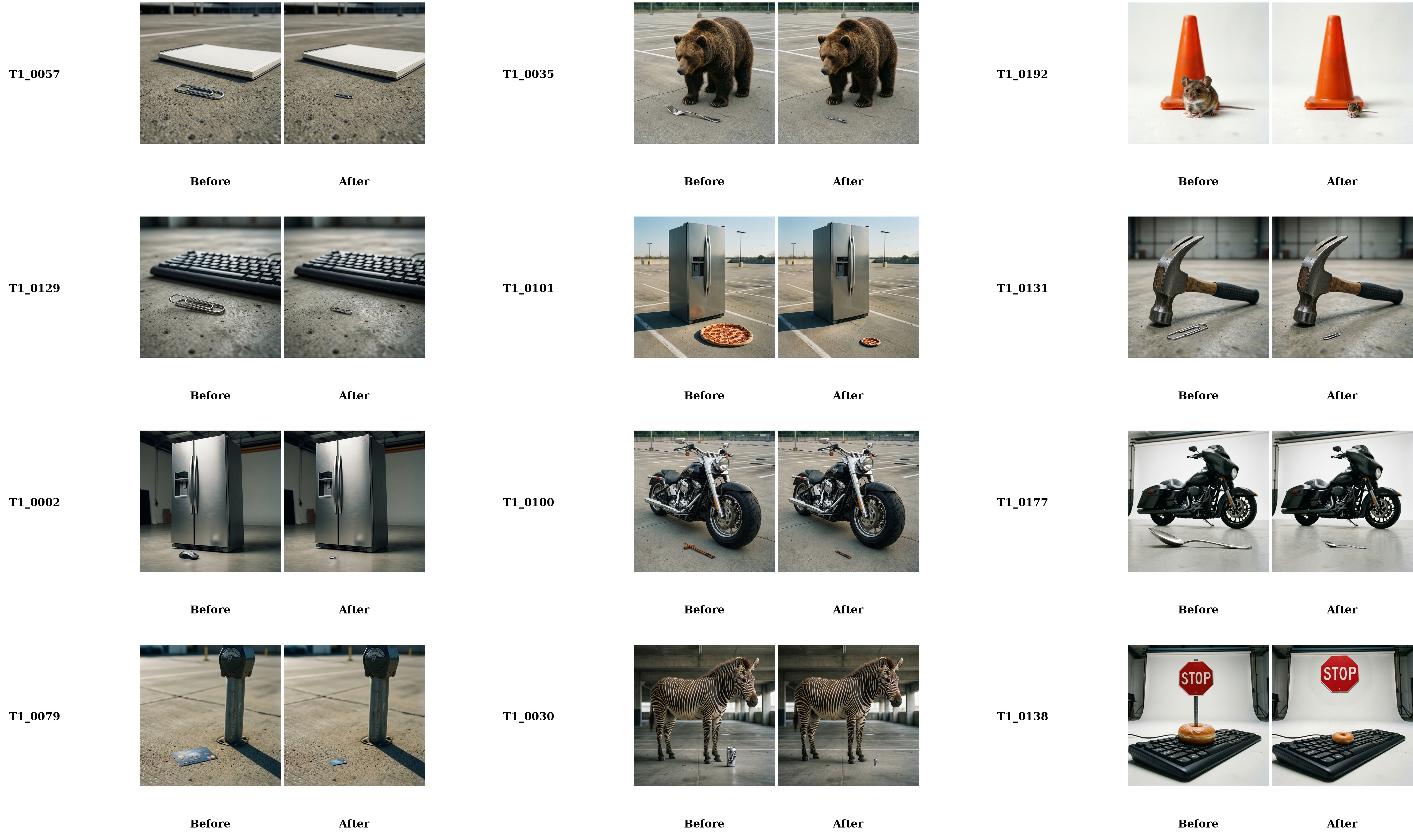}
    \vspace{-0.5ex}
    \caption{\textbf{\modelname{} corrections on S1 Natural Depth examples.}
    Each example shows the original generated image and the corrected output.
    \modelname{} uses the diagnosed scale relation to resize the local target while preserving the natural-depth composition.}
    \label{fig:qual_rescale_s1}
    \vspace{-1ex}
\end{figure}

\begin{figure}[p]
    \centering
    \includegraphics[width=\linewidth]{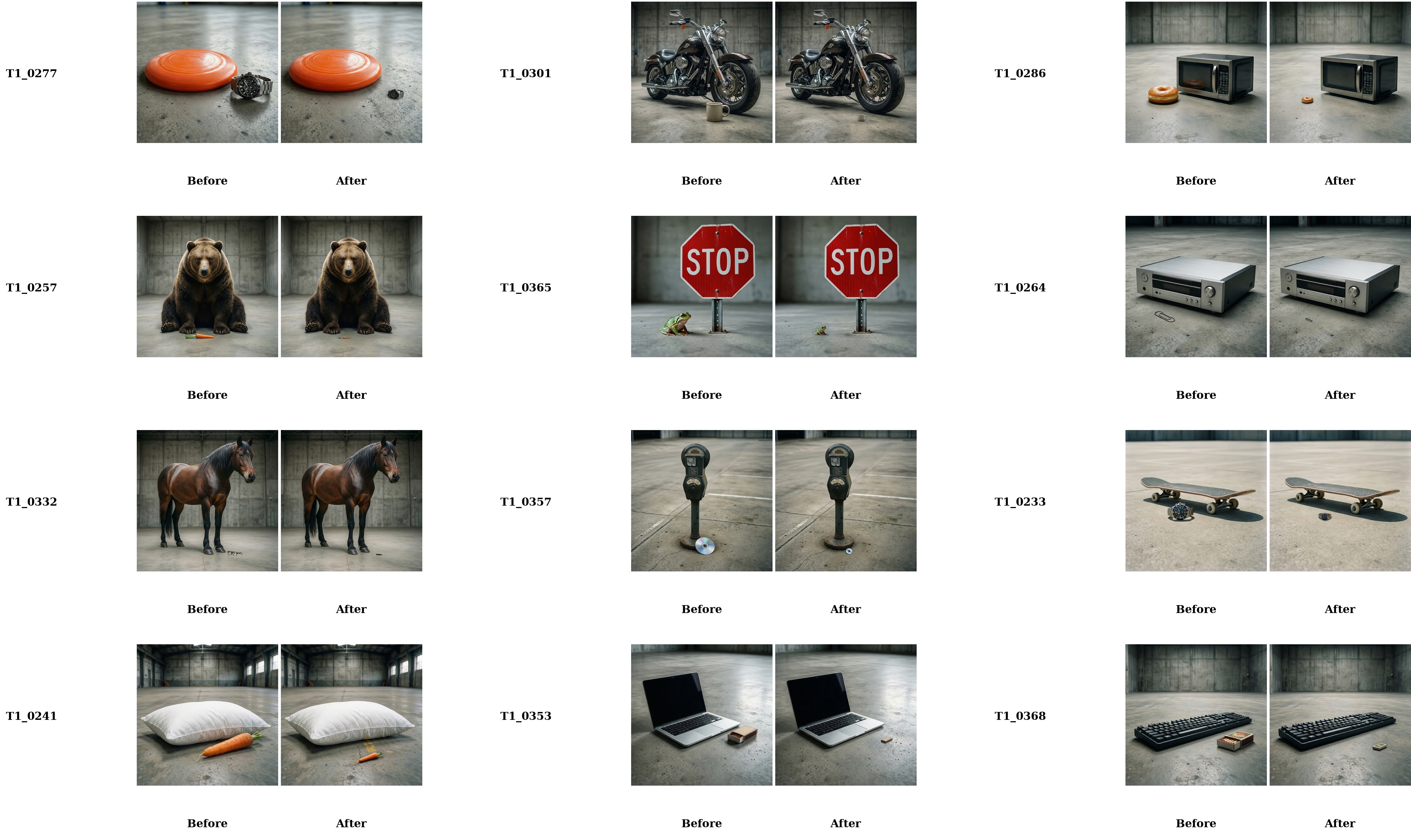}
    \vspace{-0.5ex}
    \caption{\textbf{\modelname{} corrections on S2 Same-Plane examples.}
    Because S2 constrains objects to the same depth plane, scale errors are visually direct; the correction mainly requires accurate local resizing and reinsertion.}
    \label{fig:qual_rescale_s2}
    \vspace{-1ex}
\end{figure}

\begin{figure}[p]
    \centering
    \includegraphics[width=\linewidth]{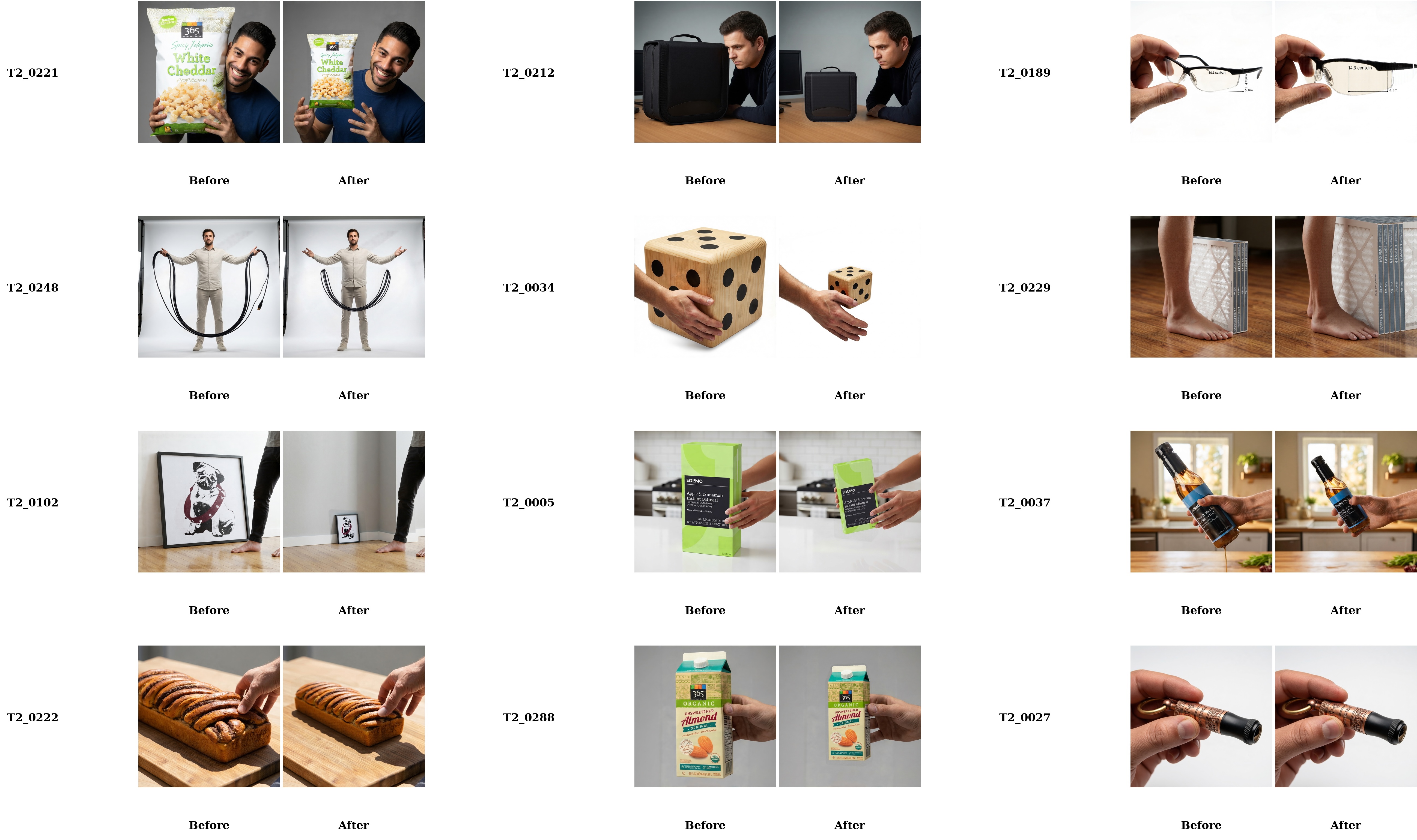}
    \vspace{-0.5ex}
    \caption{\textbf{\modelname{} corrections on S3 Human--Product examples.}
    These examples show corrections where the product is resized relative to a human anchor while preserving the product identity and the surrounding interaction context.}
    \label{fig:qual_rescale_s3}
    \vspace{-1ex}
\end{figure}

\begin{figure}[p]
    \centering
    \includegraphics[width=\linewidth]{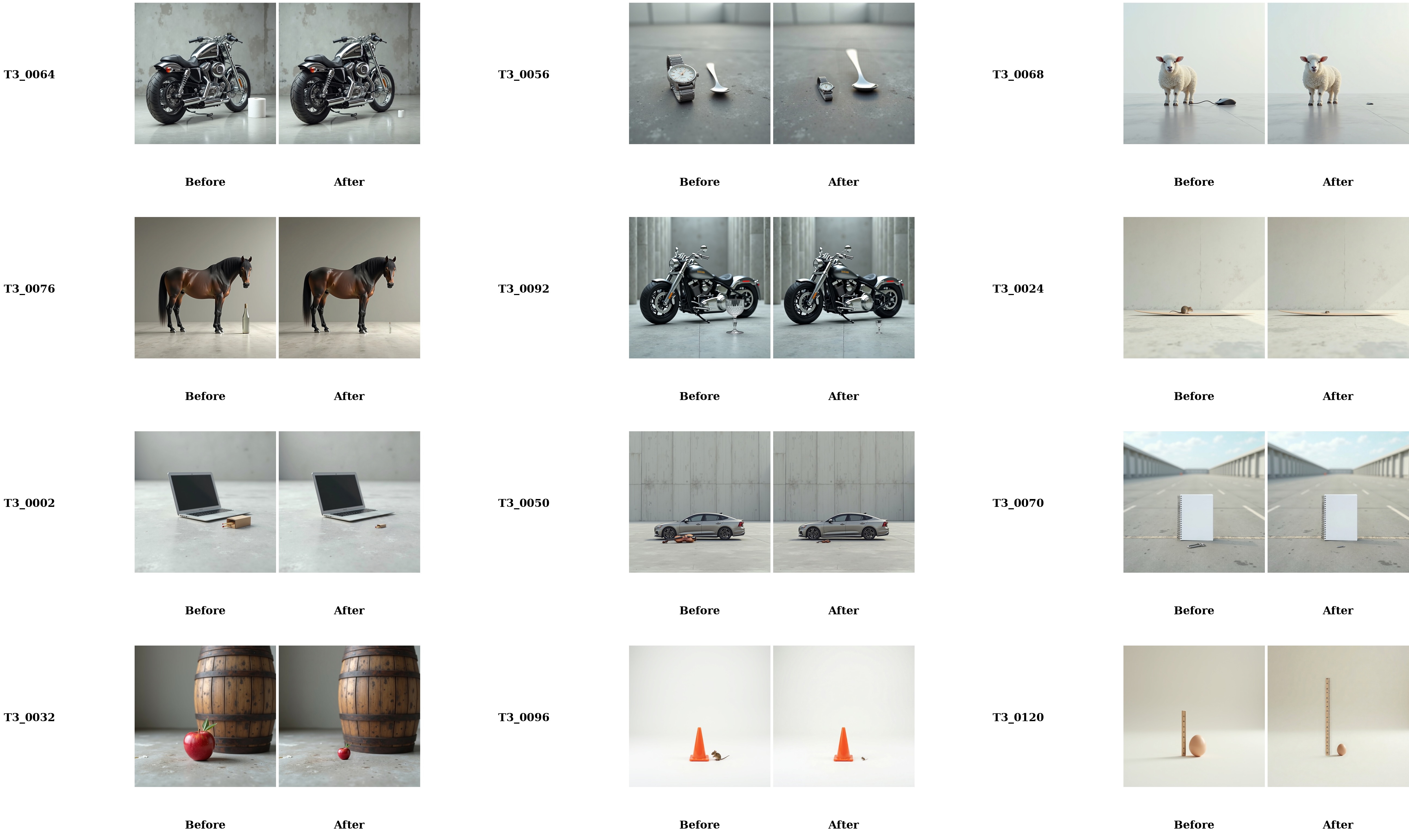}
    \vspace{-0.5ex}
    \caption{\textbf{\modelname{} corrections on S4 Hard Auto-Discovery examples.}
    {These source images belong to S4; the shown \modelname{} outputs use the feedback-guided correction protocol described in~\cref{app:rescale_agent_plan}.}}
    \label{fig:qual_rescale_s4}
    \vspace{-1ex}
\end{figure}

\begin{figure}[p]
    \centering
    \includegraphics[width=\linewidth]{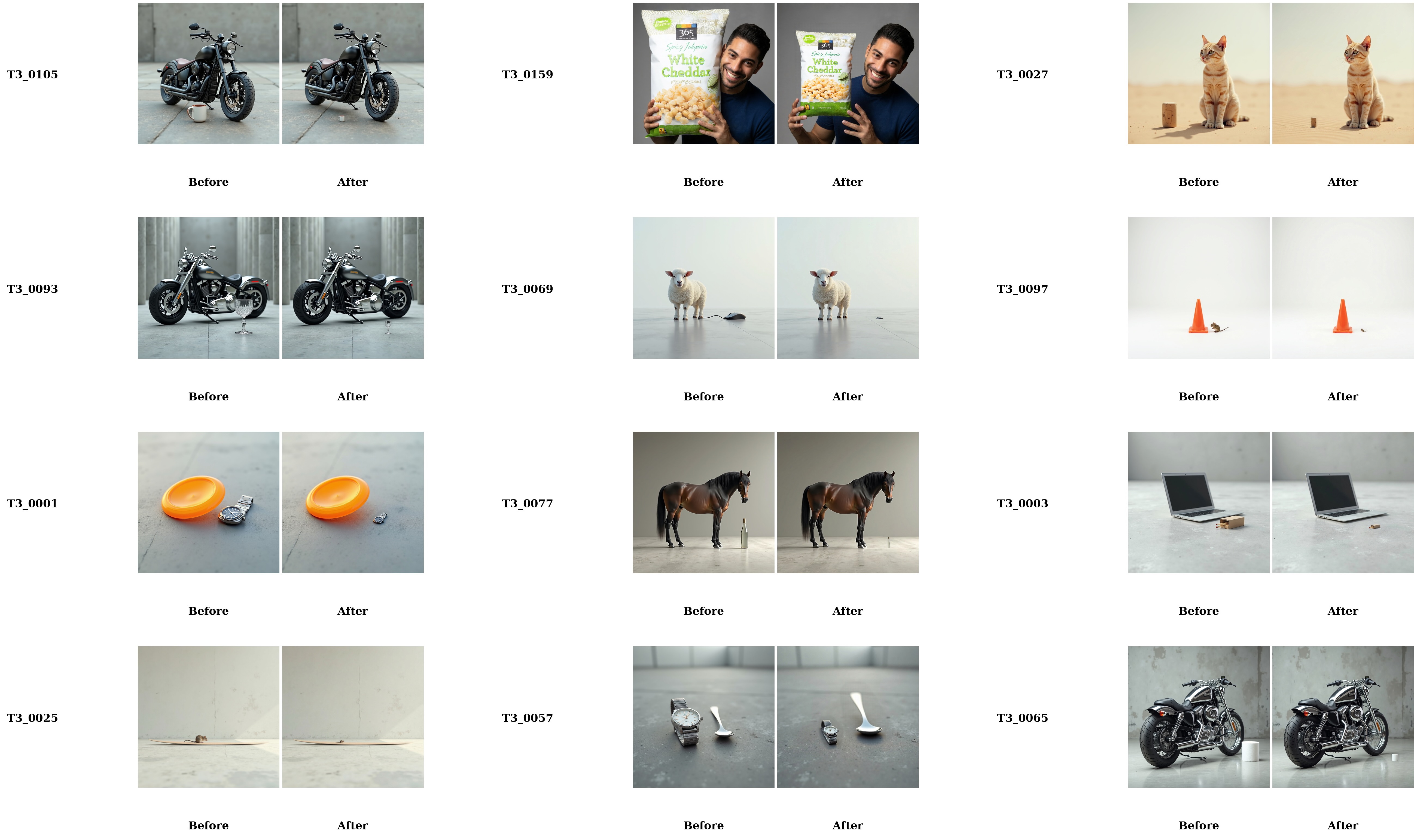}
    \vspace{-0.5ex}
    \caption{\textbf{\modelname{} corrections on S5 Precise Scale Instruction examples.}
    {These source images also define S5 entries. \modelname{} uses the same feedback-guided correction outputs for both prompt variants.}}
    \label{fig:qual_rescale_s5}
    \vspace{-1ex}
\end{figure}

\cref{fig:qual_rescale_s1,fig:qual_rescale_s2,fig:qual_rescale_s3,fig:qual_rescale_s4,fig:qual_rescale_s5} show additional before--after examples from \modelname{}.
{The before--after views show target resizing and reinsertion within the surrounding scene.}

\subsection[Correction Decisions and Local Synthesis]{{Correction Decisions and Local Synthesis}}
\label{app:qualitative_failures}

\begin{figure}[p]
    \centering
    \includegraphics[width=\linewidth]{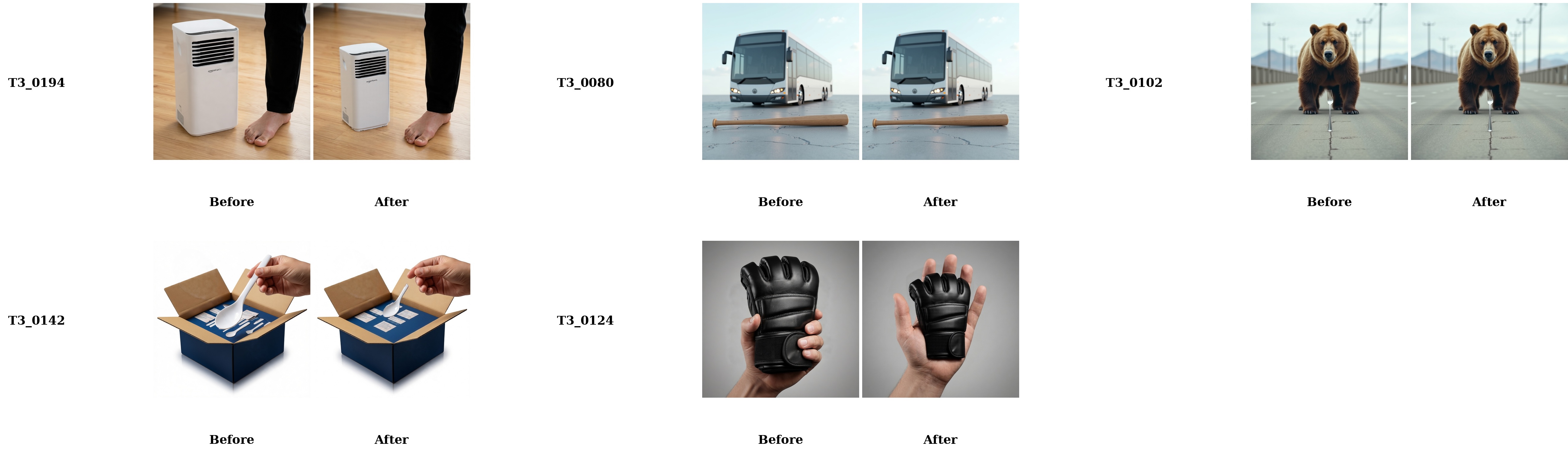}
    \vspace{-0.5ex}
    \caption{{\textbf{Diagnosis and execution in S4 Hard Auto-Discovery.}
    Uncorrected scale errors, inaccurate factors, and insufficient resizing distinguish the decisions needed before and during local correction.}}
    \label{fig:qual_fail_s4}
    \vspace{-1ex}
\end{figure}

\begin{figure}[p]
    \centering
    \includegraphics[width=\linewidth]{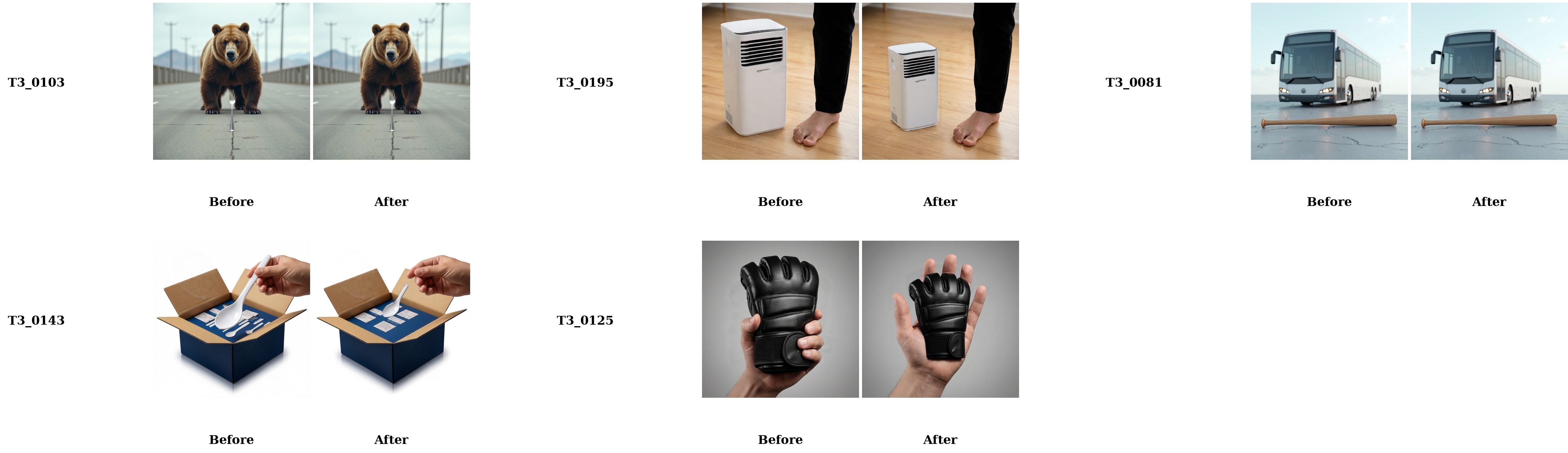}
    \vspace{-0.5ex}
    \caption{{\textbf{Local synthesis in S5 Precise Scale Instruction.}
    With the target and scale factor specified, artifacts, identity changes, and support-region deviations isolate the visual-preservation demands of insertion.}}
    \label{fig:qual_fail_s5}
    \vspace{-1ex}
\end{figure}

{\cref{fig:qual_fail_s4,fig:qual_fail_s5} separate correction decisions from local image synthesis. S4 combines object selection, scale estimation, and edit execution; its examples show how target or factor errors propagate to the corrected image. S5 specifies the edit plan and isolates boundary blending, shape preservation, and local content consistency. Together, these cases motivate a backend that decouples the edit mask from the final object size and shape while preserving object identity.}

\section{Comparison to GenSpace}
\label{app:genspace_comparison}

GenSpace is the closest prior benchmark to GenScale because it includes a relative-size test within a broader spatial-awareness suite. Its Relative Size criterion treats an object pair as correct when the prompt-specified larger object has at least 1.2 times the predicted volume of the smaller object. GenScale instead uses object-pair-specific physical size ratios and tolerance intervals, which enables a finer-grained evaluation of whether the rendered scale matches real-world proportions.

To quantify this difference, we apply the GenSpace Relative Size test to Task~1 object pairs and compare its binary judgments with the GenScale pairwise scale labels. We exclude pairs for which GenSpace fails to detect one or more named objects (14.5\% of pairs) and omit Task~2 because human--product metric scale with product references is out of distribution for GenSpace. As shown in~\cref{tab:genspace_comparison}, many pairs that pass GenSpace are still judged incorrect by GenScale, indicating that GenScale detects scale errors that are invisible to a coarse larger-versus-smaller criterion.
\begin{table}[h]          
\caption{\textbf{Comparison between GenScale and GenSpace.} Percentages are computed over 5,357 Task~1 object pairs.}
  \centering
  \label{tab:genspace_comparison}         
  \begin{tabular}{ll cc}                                                                  
  \toprule
   & & \multicolumn{2}{c}{GenSpace} \\                                                    
  \cmidrule(lr){3-4}                                                                      
   & & Correct& Incorrect  \\                                 
  \midrule                                                                                
  \multirow{2}{*}{Ours} & Correct   & 25.2 &   4.1 \\   
                        & Incorrect  & 60.1 &   10.6\\  
  \bottomrule                                                                             
  \end{tabular}                                                                                                            
  \end{table}

\clearpage\begingroup
\section{Evaluation Reliability and Correction Analysis}
\label{app:new_results}
These analyses establish benchmark coverage, evaluator agreement, and the effectiveness of localized correction. Scale error is mean $|s-3|$, where $s=3$ denotes plausible scale; gain is the paired reduction in error.

\subsection{Category Coverage and Stable Model Comparisons}
\label{app:new_coverage}
Task~1 contains 400 prompts, 1,143 relations, and 97 common-object categories, with 5/23/44 relations per category (minimum/median/maximum; Gini coefficient 0.196). Scale errors extend across the category pool: 96.9\% of categories have plausible rates below 60\%. Prompt-level bootstrap resolves 18 of 21 comparisons among seven generators at 95\% confidence, and the leading model ranks first in 99.85\% of resamples. This supports broad category coverage and stable model-level conclusions.

Task~2 covers 276 product types and four human-anchor regimes. \cref{tab:new_anchors} shows that hand-scale interaction is the most demanding regime; its approximately 51\% share emphasizes a practically relevant scale-control challenge.

\begin{table}[htbp]\begingroup\centering\small
\caption{\textbf{Task~2 results by human-anchor regime.} Pooled scored outputs across models.}
\label{tab:new_anchors}
\begin{tabular}{@{}lccc@{}}\toprule
Anchor & Relations & Scale error$\downarrow$ & Plausible (\%)$\uparrow$\\\midrule
Hand &886&0.376&64.8\\
Head/face &441&0.351&71.0\\
Full body &239&0.318&78.7\\
Foot/leg &175&0.297&78.9\\\bottomrule
\end{tabular}\endgroup\end{table}

\paragraph{Physical-instance interpretation.}
Scale references describe ordinary full-size instances. Intentional toys, miniatures, or replicas are flagged separately from scale errors. 

\subsection{Human Agreement and Cross-vendor Evaluation}
\label{app:new_judges}
Nine annotators provide 4,523 ratings on 285 images and 527 relations. Ordinal Krippendorff's $\alpha=0.617$ (95\% CI [0.580,0.651]), mean pairwise QWK is 0.613, and split-half consensus correlation is 0.805. Per-rater leave-one-out agreement appears in~\cref{tab:human_gemini_alignment}; pooled comparisons provide the human reference in~\cref{tab:new_judges}.

\cref{tab:new_judges} demonstrates transfer of the scale rubric to Claude Opus 5, GPT-5.6 Sol, and Gemini 3.6 Flash~\citep{anthropic2026opus5,openai2026gpt56sol,google2026gemini36flash} without model-specific tuning. Inter-judge ordinal agreement is $\alpha=0.726$. The three alternative judges also rank feedback-guided \modelname{} first on the correction source images, supporting the correction result across vendors. Judge comparisons use a common 1024-pixel resolution. 

On metric human--product scenes, the default judge achieves MAE 0.270 against human consensus, compared with 0.397 for the human leave-one-out reference.

\subsection{Ratio Elicitation for Open-weight Evaluation}
\label{app:new_open}
GenScale-Open estimates the visible target/anchor ratio and converts it into an ordinal scale score. Ratio elicitation improves human alignment across Qwen3-VL and InternVL3~\citep{bai2025qwen3vl,zhu2025internvl3} (\cref{tab:new_open_shared}), making an explicit physical comparison a useful interface for local scale evaluation.

\begin{table}[htbp]\begingroup\centering\small
\caption{\textbf{Ratio elicitation for open-weight VLMs.} Pearson correlation with human consensus for direct rubric scoring and ratio-based scoring. GPU configurations specify inference resources.}
\label{tab:new_open_shared}
\begin{tabular}{@{}lccc@{}}\toprule
Model & Direct rubric $r\uparrow$ & Ratio elicitation $r\uparrow$ & GPUs\\\midrule
Qwen3-VL-32B &.211&.524&$1\times$B200\\
Qwen3-VL-235B &.234&.528&$4\times$H100\\
InternVL3-78B &.227&.520&$2\times$B200\\\bottomrule
\end{tabular}\endgroup\end{table}

\paragraph{Human-calibrated elicitation.}
Using \texttt{Qwen/Qwen3-VL-32B-Instruct-FP8}, we compare seven predefined prompt variants on 525 relations from 283 images with valid predictions for every variant. Each of 1,000 image-disjoint splits assigns 170 images to development and 113 to testing; the development side selects the prompt by correlation with human consensus. Asking for depicted lengths in centimeters is selected in every split and reaches mean held-out $r=0.664$, improving on the unmodified ratio prompt by 0.182. The central 95\% of split correlations is [0.594,0.727]. Inference uses vLLM~\citep{kwon2023vllm} on one B200, temperature 0.35, a 16,384-token context, and a 2,500-token output limit.

\subsection{Efficient Judge Sampling}
\label{app:new_cost}
\cref{tab:new_cost} compares estimated API costs for a reference workload of approximately 1,618 relations per model. \cref{tab:new_sampling} tests aggregation at different sample budgets using subsets of the same five judgments.

\begin{table}[htbp]\begingroup\centering\small
\caption{\textbf{Estimated evaluation cost per new model (USD).} Common workload of approximately 1,618 relations, with API rates used in the experiments.}
\label{tab:new_cost}
\begin{tabular}{@{}ccccc@{}}\toprule
Samples/pair & \shortstack{Gemini 3.1 Pro\\Preview} & Gemini 3.6 Flash & Claude Opus 5 & GPT-5.6 Sol\\\midrule
1&7.11&5.22&17.40&17.77\\
3&21.32&15.66&52.20&53.30\\
5&35.53&26.10&87.01&88.83\\\bottomrule
\end{tabular}\endgroup\end{table}

\begin{table}[htbp]\begingroup\centering\small
\caption{\textbf{Effect of judge sample count.} Correlation with human consensus, maximum model-level error change relative to five samples, and model ordering. Subsets of the same five judgments isolate the aggregation effect.}
\label{tab:new_sampling}
\begin{tabular}{@{}cccc@{}}\toprule
Samples/pair & $r_{\mathrm{human}}\uparrow$ & Maximum $|\Delta\mathrm{error}|\downarrow$ & Seven-model ordering\\\midrule
1&.812&.02&Identical\\
3&.826&--&Identical\\
5&.832&.00&Default\\\bottomrule
\end{tabular}\endgroup\end{table}

A single judgment preserves the seven-model ordering at approximately one-fifth of the evaluation cost. Repeated judgments strengthen human alignment, providing a practical choice between economical screening and the five-sample protocol used for the main comparisons.

\subsection{Human Evaluation of Correction}
\label{app:new_humans}
Four raters, including one author, evaluate scale before and after correction. The study assigns 120 source/edited pairs: 85 sampled across tasks and source models, plus 35 challenging pairs covering aggressive rescaling, initially plausible relations, and difficult scene conditions. Of these, 112 pairs have valid consensus in both images, giving the 77 representative and 35 challenge pairs in~\cref{tab:new_humans}. Image order and method labels are hidden. Raters use object identities, physical lengths, and the five-level rubric in~\cref{app:human_vlm_eval}, marking unidentifiable pairs unscorable. Consensus uses the modal score with median-based tie breaking; gain intervals resample image pairs. Compensation follows the same \$30 gift-card standard as the original annotation study.

\cref{tab:new_humans} confirms substantial scale gains in both groups. In the representative group, 62 of 68 initially erroneous relations reach plausible scale. This identifies correction of genuine scale errors as the primary source of improvement. The default evaluator also follows human leave-one-out agreement on the source and edited images (\cref{tab:new_panel_alignment}).

\begin{table}[htbp]\begingroup\centering\small
\caption{\textbf{Evaluator agreement on correction pairs.} Pearson correlation on source and edited images. Evaluators use full panel consensus; human leave-one-out uses the remaining raters.}
\label{tab:new_panel_alignment}
\begin{tabular}{@{}lcc@{}}\toprule
Images & $r_{\mathrm{default,human}}\uparrow$ & $r_{\mathrm{human,LOO}}\uparrow$\\\midrule
Source &.733&.738\\
Edited &.498&.488\\\bottomrule
\end{tabular}\endgroup\end{table}

\subsection{Correction Gains and Edit Selection}
\label{app:new_overshoot}
\modelname{} raises the plausible rate by 21.0 percentage points across 6,865 matched Task~1 relations. The 2,020 repaired relations outnumber the 580 newly introduced scale errors by 3.5:1, yielding a net increase of 1,440 plausible relations.

The frisbee--watch example provides a concrete physical reference: 27.4\,cm versus 4.0\,cm, or 6.85:1. Feedback-guided \modelname{} attains score~3, illustrating the benchmark's ability to assess physical scale independently of visual attractiveness. Multi-object examples in~\cref{fig:qual_rescale_s1,fig:qual_rescale_s2} further illustrate how a local edit affects relations sharing the resized object.

\subsection{Structural Reliability and the Role of Feedback}
\label{app:new_ablation}
The component study evaluates 100 erroneous cases. A structural defect is a previously scorable relation becoming unscorable through a missing, duplicated, merged, or degraded object. \cref{tab:new_structure} isolates the contribution of depth conditioning and the localization interface.

\begin{table}[htbp]\begingroup\centering\small
\caption{\textbf{Structural preservation under component changes.} Defect rate and 95\% interval. Feedback-guided configurations evaluate 349 relations; autonomous operation evaluates 289 relations (6 defects).}
\label{tab:new_full_structure}\label{tab:new_structure}
\begin{tabular}{@{}lcc@{}}\toprule
Configuration & Defect rate (\%)$\downarrow$ & 95\% CI\\\midrule
Full pipeline&1.15&[.45,2.91]\\
No depth&9.46&[6.81,12.98]\\
Detector box instead of SAM 2 mask&1.72&[.79,3.70]\\
No global review&2.29&[1.17,4.46]\\
Fully autonomous&2.08&[.95,4.45]\\\bottomrule
\end{tabular}\endgroup\end{table}

Depth conditioning reduces structural defects by approximately eightfold compared with removing depth, supporting its role in preserving scene geometry. Detector-box substitution also retains low defect rates, supporting a modular localization interface.

\begin{table}[htbp]\begingroup\centering\small
\caption{\textbf{Contribution of evaluator feedback.} Both configurations receive object identities and physical-size metadata. Feedback supplies relation scores and explanations; edit coverage counts cases receiving a correction among the same 100 inputs.}
\label{tab:new_autonomy}
\begin{tabular}{@{}lc@{}}\toprule
Input & Edited cases\\\midrule
With evaluator feedback&67/100\\
Without evaluator feedback&31/100\\\bottomrule
\end{tabular}\endgroup\end{table}

Evaluator feedback increases correction coverage (\cref{tab:new_autonomy}), connecting benchmark diagnosis to localized editing. The main experiments use this feedback-guided configuration.

\subsection{Correction Across Scene Conditions}
\label{app:new_hard}
Condition-wise analysis uses 7,017 initially scorable Task~1 relations with both source and edited records passing the scene prefilter. Of these, 6,865 remain scorable after editing; 152/7,017 (2.17\%, 95\% Wilson interval [1.85,2.53]) become unscorable. Scale gains use the matched relations within each condition. Aspect-ratio and contrast thresholds define the upper third of their respective distributions; material tags identify transparent or specular objects.

\begin{table}[htbp]\begingroup\centering\small
\caption{\textbf{Correction across scene conditions.} Gain is the paired reduction in scale error on matched relations. Structural defects count previously scorable relations becoming unscorable, with the corresponding initial denominator. Tags describe object count, aspect ratio, image contrast, and material.}
\label{tab:new_hard}
\begin{tabular}{@{}lccc@{}}\toprule
Condition & Matched relations & Gain$\uparrow$ & Defects $k/n$ (\%)$\downarrow$\\\midrule
Two objects&1,132&0.532&4/1,136 (0.35)\\
Three objects&2,380&0.373&29/2,409 (1.20)\\
Four or more objects&3,353&0.281&119/3,472 (3.43)\\
Compact/mid objects&4,688&0.374&113/4,801 (2.35)\\
Elongated objects&2,177&0.312&39/2,216 (1.76)\\
Normal contrast&4,600&0.369&71/4,671 (1.52)\\
High contrast&2,265&0.323&81/2,346 (3.45)\\
Opaque objects&6,113&0.365&138/6,251 (2.21)\\
Transparent/specular objects&752&0.269&14/766 (1.83)\\
\bottomrule\end{tabular}\endgroup\end{table}

Positive gains across all conditions show that correction extends across clutter, lighting contrast, elongated shapes, and transparent or specular materials. 

\subsection{Evaluation Resources}
\label{app:compute}\label{app:llm_vlm_usage}
The default evaluator uses repeated VLM API judgments, with estimated costs and sampling choices in~\cref{app:new_cost}. Open-weight evaluation uses the GPU configurations in~\cref{tab:new_open_shared}. The correction implementation configures one H100 per worker; concurrency scales through multiple workers. 
\endgroup

\clearpage\begingroup
\section{Release, Licenses, Safeguards, and Societal Considerations}
\label{app:release_ethics}

\subsection{Benchmark and Code Release}

The \href{https://huggingface.co/datasets/lingxiao2049/GenScale}{\benchmarkname{} dataset}, \href{https://github.com/lingxiao-li/GenScale}{source code}, and \href{https://huggingface.co/lingxiao2049/Rescale}{\modelname{} checkpoint} are publicly available through the \href{https://lingxiao-li.github.io/genscale/}{project page}.
The dataset includes prompts, scale metadata, selected product references, and generated Task~3 inputs, as described in~\cref{app:release_format}. It also provides \href{https://huggingface.co/datasets/lingxiao2049/GenScale/blob/main/croissant.json}{Croissant metadata} with Responsible AI fields.
The code repository provides evaluation and score aggregation scripts, \modelname{} inference code, and instructions for setup and use.

\subsection{Existing Assets and Licenses}
\label{app:licenses}

The code is released under the MIT License, the benchmark dataset under CC BY 4.0, and the \modelname{} weights under the FLUX.1 [dev] Non-Commercial License.
The dataset preserves attribution for ABO product metadata and reference images~\citep{collins2022abo}, with source identifiers for tracing the original records.
Third-party assets and model dependencies retain their respective license terms, as documented in the release notices.

\subsection{Responsible Release Safeguards}
\label{app:safe_guards}

The benchmark is designed to evaluate physical-scale consistency rather than identity, demographic attributes, or private personal information.
Prompts avoid requesting named individuals, sensitive personal attributes, or targeted depictions of real people.
Human figures in Task~2 are used as generic scale anchors for product-size evaluation, and the evaluation focuses only on the relative size between the product and the human anchor.

Before release, we remove or withhold samples that contain unsafe, private, or personally identifying content, and we retain the invalid-pair filtering used in the evaluation pipeline.
We also avoid releasing private API credentials, hidden model outputs, or any data that would enable unauthorized use of proprietary systems.

Users of the released code and benchmark will be asked to follow the licenses of the underlying datasets and model providers and to disclose synthetic-image editing when outputs are used outside research evaluation.

\subsection{Human Annotation Compensation and Participant Risk}
\label{app:human_annotation}

Human annotation was used to calibrate the ordinal VLM judge and to validate the evaluation protocol.
Annotators were asked to inspect generated images and assign pairwise relative-scale labels according to the provided rubric.
The task involved ordinary visual judgments over generated or product-reference images and did not require annotators to disclose demographic attributes, private personal information, political views, health information, or other sensitive data.

Each annotation session was designed to take approximately one to two hours.
Annotators received a fixed \$30 gift card, corresponding to an effective hourly compensation of \$15--\$30 depending on completion time.

Annotators were compensated for completing the task rather than for matching any model prediction or preferred answer, reducing incentive pressure to provide a particular label.
The task posed minimal participant risk: the images were non-sensitive visual scenes, the instructions did not ask annotators to make personal disclosures, and the collected annotations were used only in aggregate for evaluator calibration.

\subsection{Broader Impacts}
\label{app:borader_impacts}

Reliable relative-scale evaluation can improve the physical realism of image generation and editing systems.
This is useful for scientific benchmarking, product visualization, assistive design tools, embodied-AI simulation, and other settings where object size should be physically plausible.
By making scale errors measurable, \benchmarkname{} can also help reveal systematic weaknesses in current generative models that are not captured by generic image-quality or text-alignment metrics.

At the same time, improved scale correction may make synthetic images harder to recognize as edited, especially in contexts where unrealistic object size would otherwise reveal manipulation.

The public release documents intended research uses and preserves source and edit provenance where available. We encourage downstream users to disclose synthetic edits in public-facing applications.

\endgroup

\clearpage\section*{NeurIPS Paper Checklist}
\begin{enumerate}

\item {\bf Claims}
    \item[] Question: Do the main claims made in the abstract and introduction accurately reflect the paper's contributions and scope?
    \item[] Answer: \answerYes{}
    \item[] Justification: The abstract and introduction state the paper's main contributions: GenScale, a human-calibrated relative-scale evaluation protocol, the Rescale correction pipeline, and benchmark results on contemporary generators and editors. These claims are supported by the benchmark construction in~\cref{sec:genscale}, the evaluator calibration in~\cref{sec:genscale}, the correction method in~\cref{sec:rescale}, and the empirical results in~\cref{sec:evaluation_results}.

\item {\bf Limitations}
    \item[] Question: Does the paper discuss the limitations of the work performed by the authors?
    \item[] Answer: \answerYes{}
    \item[] Justification: Sec.~\ref{sec:conclusion} discusses the scope and limitations of GenScale and Rescale, including the focus on relative object scale, the use of visually identifiable categories with stable physical dimensions, and assumptions about visible and locally editable objects.

\item {\bf Theory assumptions and proofs}
    \item[] Question: For each theoretical result, does the paper provide the full set of assumptions and a complete (and correct) proof?
    \item[] Answer: \answerNA{}
    \item[] Justification: The paper does not present formal theoretical results, theorems, or proofs. Its contributions are empirical, methodological, and benchmark-oriented.

    \item {\bf Experimental result reproducibility}
    \item[] Question: Does the paper fully disclose all the information needed to reproduce the main experimental results of the paper to the extent that it affects the main claims and/or conclusions of the paper (regardless of whether the code and data are provided or not)?
    \item[] Answer: \answerYes{}
    \item[] Justification: The paper describes the benchmark construction, physical-size metadata, task inputs, filtering criteria, ordinal evaluation protocol, judge calibration, repeated-query aggregation, evaluated models, and reported metrics in Secs.~\ref{sec:genscale} -- \ref{sec:evaluation} and the appendix. We have released the benchmark metadata, prompts, evaluation scripts, judge prompts, and Rescale implementation details to support reproduction of the main results.

\item {\bf Open access to data and code}
    \item[] Question: Does the paper provide open access to the data and code, with sufficient instructions to faithfully reproduce the main experimental results, as described in supplemental material?
    \item[] Answer: \answerYes{}
    \item[] Justification: The benchmark and code are publicly available through \url{https://lingxiao-li.github.io/genscale/}. The releases include benchmark metadata, prompts, evaluation scripts, and the \modelname{} implementation, together with usage instructions and licensing information. The benchmark schema and release details are described in~\cref{app:release_format,app:release_ethics}.

\item {\bf Experimental setting/details}
    \item[] Question: Does the paper specify all the training and test details (e.g., data splits, hyperparameters, how they were chosen, type of optimizer) necessary to understand the results?
    \item[] Answer: \answerYes{}
    \item[] Justification: The paper specifies the benchmark splits, task/scenario definitions, model-facing inputs, filtering rules, pairwise evaluation protocol, calibrated judge settings, repeated scoring and aggregation procedure, and evaluation metrics in Secs.~\ref{sec:genscale} -- \ref{sec:evaluation} and the appendix. {The default Rescale pipeline uses fixed external modules; the optional trained insertion-backend exploration is described separately in~\cref{app:rescale_details}.}

\pagebreak
\item {\bf Experiment statistical significance}
    \item[] Question: Does the paper report error bars suitably and correctly defined or other appropriate information about the statistical significance of the experiments?
    \item[] Answer: \answerYes{}
    \item[] Justification: The appendix reports uncertainty estimates for the main quantitative results and evaluator calibration. In particular, \cref{app:bootstrap_ci} describes image-level bootstrap confidence intervals for Gemini--human alignment, and~\cref{app:full_results} reports 95\% confidence intervals from 2{,}000 bootstrap resamples for the main benchmark and correction results.

\item {\bf Experiments compute resources}
    \item[] Question: For each experiment, does the paper provide sufficient information on the computer resources (type of compute workers, memory, time of execution) needed to reproduce the experiments?
    \item[] Answer: {\answerNo{}}
    \item[] Justification: {\cref{app:compute} and \cref{app:new_cost} report the supplied open-judge GPU configurations and estimated scoring costs. Aggregate execution time, memory usage, and total API usage were not recorded.}

\item {\bf Code of ethics}
    \item[] Question: Does the research conducted in the paper conform, in every respect, with the NeurIPS Code of Ethics \url{https://neurips.cc/public/EthicsGuidelines}?
    \item[] Answer: \answerYes{}
    \item[] Justification: The research is an evaluation benchmark and correction method for generated-image scale realism, uses externally documented object-size metadata and product references, and reports human annotation procedures. We have reviewed the NeurIPS Code of Ethics and are not aware of any deviation from it.

\item {\bf Broader impacts}
    \item[] Question: Does the paper discuss both potential positive societal impacts and negative societal impacts of the work performed?
    \item[] Answer: \answerYes{}
    \item[] Justification: \cref{app:borader_impacts} discusses both positive impacts, such as improved scale-aware evaluation and product visualization, and potential negative impacts, such as making synthetic edits harder to recognize. It also describes mitigation-oriented release choices, including research-oriented release, provenance preservation when possible, and encouraging disclosure of edited synthetic images.

\item {\bf Safeguards}
    \item[] Question: Does the paper describe safeguards that have been put in place for responsible release of data or models that have a high risk for misuse (e.g., pre-trained language models, image generators, or scraped datasets)?
    \item[] Answer: \answerYes{}
    \item[] Justification: \cref{app:safe_guards} describes responsible release safeguards: prompts avoid named individuals and sensitive personal attributes, unsafe or personally identifying samples are removed or withheld, private credentials and hidden model outputs are not released, and proprietary assets are redistributed only when permitted. The release is framed as benchmark code and metadata for research-oriented scale-error diagnosis rather than as a general-purpose deceptive editing tool.

\item {\bf Licenses for existing assets}
    \item[] Question: Are the creators or original owners of assets (e.g., code, data, models), used in the paper, properly credited and are the license and terms of use explicitly mentioned and properly respected?
    \item[] Answer: \answerYes{}
    \item[] Justification: \cref{app:licenses} describes attribution and licensing for existing assets, including ABO product images and metadata, public physical-size sources, pretrained perception and insertion models, and commercial model outputs. The release preserves upstream license requirements, source identifiers, citation lists, and provider terms, and avoids redistributing third-party assets when redistribution is not permitted.

\item {\bf New assets}
    \item[] Question: Are new assets introduced in the paper well documented and is the documentation provided alongside the assets?
    \item[] Answer: \answerYes{}
    \item[] Justification: The paper introduces GenScale as a new benchmark asset and documents its task taxonomy, metadata fields, physical-size grounding, prompt construction, filtering rules, evaluation protocol, and release format in Sec.~\ref{sec:genscale} and the appendix. The released asset includes documentation for metadata schemas, prompts, pairwise records, evaluation scripts, intended use, and limitations.

\item {\bf Crowdsourcing and research with human subjects}
    \item[] Question: For crowdsourcing experiments and research with human subjects, does the paper include the full text of instructions given to participants and screenshots, if applicable, as well as details about compensation (if any)? 
    \item[] Answer: \answerYes{}
    \item[] Justification: \cref{app:human_vlm_eval} provides the human annotation interface, participant instructions, ordinal rubric, invalid-pair criteria, and consensus construction, while~\cref{app:human_annotation} reports compensation and participant-risk details. Annotators completed approximately one to two hours of visual scale-labeling work and received a fixed \$30 gift card, corresponding to \$15--\$30 per hour. {The four-rater correction study uses the same compensation standard; its protocol is described in~\cref{app:new_humans}.}

\item {\bf Institutional review board (IRB) approvals or equivalent for research with human subjects}
    \item[] Question: Does the paper describe potential risks incurred by study participants, whether such risks were disclosed to the subjects, and whether Institutional Review Board (IRB) approvals (or an equivalent approval/review based on the requirements of your country or institution) were obtained?
    \item[] Answer: \answerNo{}
    \item[] Justification: The paper describes the annotation task, compensation, and participant risks in~\cref{app:human_annotation}, and the task involved only minimal-risk visual judgments over non-sensitive generated or product-reference images without collecting sensitive personal information. {No formal IRB approval or exemption determination was obtained.}

\item {\bf Declaration of LLM usage}
    \item[] Question: Does the paper describe the usage of LLMs if it is an important, original, or non-standard component of the core methods in this research? Note that if the LLM is used only for writing, editing, or formatting purposes and does \emph{not} impact the core methodology, scientific rigor, or originality of the research, declaration is not required.
    \item[] Answer: \answerYes{} 
    \item[] Justification: The paper uses LLM/VLM-based components as part of the core methodology rather than only for writing or formatting. Specifically, \cref{sec:genscale} describes the Gemini-based human-calibrated pairwise scale judge, including its inputs, repeated sampling, and aggregation protocol, and~\cref{sec:rescale} describes the multimodal agent used by \modelname{} to diagnose scale errors and plan localized corrections. {Full default-judge prompts appear in~\cref{app:gemini_judge_prompt}; supplemental evaluator analyses appear in~\cref{app:new_results}.}

\end{enumerate}




\end{document}